\RequirePackage{fix-cm}
\documentclass[twocolumn]{svjour3}          
\smartqed  
\usepackage{graphicx}
\usepackage{multirow}
\usepackage[misc]{ifsym}
\usepackage{booktabs}
\usepackage{subfigure}
\usepackage{natbib}
\usepackage[breaklinks,colorlinks,linkcolor=blue,anchorcolor=blue,citecolor=blue,urlcolor=blue, backref=true]{hyperref}
\usepackage[ruled]{algorithm2e}		
\usepackage{amsmath,amsfonts,amssymb,bm}

\graphicspath{{figure/}}   

\begin{document}
	
\title{A Deep Neural Network Surrogate Modeling Benchmark for Temperature Field Prediction of Heat Source Layout}
%


\author{Xianqi Chen \textsuperscript{1,2*}   \and
        Xiaoyu Zhao \textsuperscript{2*} \and 
        Zhiqiang Gong \textsuperscript{2*} \and
        Jun Zhang \textsuperscript{2} \and
        Weien Zhou \textsuperscript{2} \and
        Xiaoqian Chen \textsuperscript{2} \and
        Wen Yao \textsuperscript{2 \Letter}
}


\institute{
	\Letter{\space Corresponding Author} \\
	\email{wendy0782@126.com} \\
	\at
	\textsuperscript{1} College of Aerospace Science and Engineering, National University of Defense Technology, Changsha 410073, China \\
	\textsuperscript{2} National Innovation Institute of Defense Technology, Chinese Academy of Military Science, Beijing 100071, China \\
	\textsuperscript{*} These authors contributed equally to this work and should be considered co-first authors.  \\
}

\date{Received: date / Accepted: date}

\maketitle

\begin{abstract}

Thermal issue is of great importance during layout design of heat source components in systems engineering, especially for high functional-density products. 
Thermal analysis generally needs complex simulation, which leads to an unaffordable computational burden to layout optimization as it iteratively evaluates different schemes. 
Surrogate modeling is an effective way to alleviate computation complexity. 
However, temperature field prediction (TFP) with complex heat source layout (HSL) input is an ultra-high dimensional nonlinear regression problem, which brings great difficulty to traditional regression models. 
The Deep neural network (DNN) regression method is a feasible way for its good approximation performance. 
However, it faces great challenges in both data preparation for sample diversity and uniformity in the layout space with physical constraints, and proper DNN model selection and training for good generality, which necessitates efforts of both layout designer and DNN experts. 
To advance this cross-domain research, this paper proposes a DNN based HSL-TFP surrogate modeling task benchmark. With consideration for engineering applicability, sample generation, dataset evaluation, DNN model, and surrogate performance metrics, are thoroughly studied. 
Experiments are conducted with ten representative state-of-the-art DNN models. 
Detailed discussion on baseline results is provided and future prospects are analyzed for DNN based HSL-TFP tasks.

\end{abstract}

\section{Introduction}
\label{intro}

Recently, heat management \citep{forrester2009, emam2019}, especially over electronic devices with smaller size and higher power density has played a more and more important role in industrial designs, such as the layout of microelectronics within the micro-/nano-satellite \citep{chen2018} and the design of printed circuit boards (PCB) \citep{geczy2017}. 
Over-high temperature usually damages the electronics and compromises the performance of the equipment \citep{cheng2009, arshad2020}, even leading to serious accidents, especially for high functional-density products.
Therefore, the heat management attempts to lower the temperature and improve the temperature uniformity of electronics to avoid the poor performance and short life of the components \citep{chen2016b, martinez2019}.
Generally, optimizing the layout of the electronic components is a simple and cheap approach for effective passive heat management, raising a hot research topic of heat source layout optimization. 

The heat source layout optimization can be regarded as one kind of layout optimization driven by thermal performance, where the precise thermal evaluation plays an extremely important role in guiding the optimization search and finding the final optimal layout design.
Traditionally, most of prior works \citep{song2011, chen2016a, chen2016b, aslan2018} utilize the numerical simulation technique to calculate the temperature response of different layout schemes, such as finite element method (FEM) \citep{hughes2015} and finite difference method (FDM) \citep{gu2017}.
Although these numerical methods can provide the right evaluation for optimization, they are usually time-consuming especially with an extremely fine mesh for high precision.
In addition, the iterative search process in layout optimization requires to repeatedly call this numerical tool to calculate the temperature field of layout schemes one by one for massive design evaluation, which is computationally expensive and even unaffordable in practical engineering \citep{cuco2015, chen2018}.

To overcome this problem, using the surrogate model instead of physics simulation to achieve an end-to-end and nearly real-time temperature field prediction tends to be a feasible option for handling this kind of expensive optimization problem \citep{monier2017}.
Therefore, to motivate the surrogate modeling research, we refine the Temperature Field Prediction of Heat Source Layout (HSL-TFP) task as a base task for realizing further surrogate-assisted layout optimization.
In this task, the heat source layout is regarded as the design variable and the whole temperature distribution over the layout domain caused by the layout is the response variable, both of which are treated as images.
Hence, this task can be constructed as an image-to-image regression problem.
If the input and output images are both discretized as 200$\times$200 matrices, an accurate 40000-dimensional surrogate is demanded.
From this point, the ultra-high dimensional nonlinear characteristics of the HSL-TFP task makes it a challenging problem.

However, most of the researches on surrogate modeling focus on traditional regression models, such as polynomial regression \citep{murcia2018}, Gaussian process \citep{lackey2019}, support vector machine (SVM) regression \citep{ciccazzo2014}, radial basis function (RBF) network \citep{zhang2020}, and artificial neural network \citep{white2019, xu2019}.
These machine learning-based surrogate methods can adaptively learn the inherent laws behind the data following the given criterion (e.g., mean absolute error (MAE) \citep{goel2007}, mean square error (MSE) \citep{zhang2012}) and then obtain the predictions with the learned model.
Nevertheless, due to the limited representation ability of these shallow models, they cannot be applied to our ultra-high dimensional HSL-TFP task and provide sufficiently accurate evaluation for further layout optimization.
Therefore, exploring surrogate models with better representation ability tends to be necessary for effective and efficient predictions over the HSL-TFP task.


%

Deep neural networks (DNNs) which have demonstrated excellent performance in many tasks, such as the computer vision tasks \citep{yang2017}, provide another way to realize real-time predictions. 
DNNs with multi-layers, which are represented by convolutional neural networks (CNNs), can capture both local and global information \citep{gong2019} and find subtle differences from different input layout images and thereby provide better predictive effects on the temperature field. 
For the specific HSL-TFP task, our previous work \citep{chen2020} has illustrated the feasibility and effectiveness of using deep learning surrogate modeling to implement the ultra-high dimensional mapping task.
Therein the DNN surrogate was developed based on the feature pyramid network (FPN) framework and impressive prediction performance was demonstrated and verified.
However, the layout case considered in this preliminary research only involved simple identical heat source components.
To facilitate the HSL-TFP research in real-world applications, we construct a complex layout task where components are of different sizes, shapes and intensities in this paper.
Note that heat sources can only be placed in cells under the 10$\times$10 grid system in our prior work.
By contrast, components are allowed to move cell by cell under the 200$\times$200 grid system that represents the layout domain, thus making it an approximately continuous layout task.
Therefore, all these improvements in our HSL-TFP task guarantee the generality of problem definition, making it more practical and simultaneously intractable.

To compensate for the shortage of insufficient development of DNN surrogates, we propose to establish a new HSL-TFP based ultra-high dimensional regression surrogate modeling benchmark.
Three major issues that hinder the establishment of this task benchmark are discussed as follows.
\begin{itemize}
\item \textit{Lacking a proper benchmark dataset for research of DNN surrogate models over HSL-TFP task:} 
One basic geometry constraint that should be strictly satisfied in the layout is the non-overlapping constraint between components or between components and the layout domain.
Only the layout scheme without overlap can be regarded as one valid layout sample.
Hence, this is a constrained sampling problem.
An intuitive approach to solve this problem is the acceptance-rejection sampling method. 
But experiments show that there is an extremely low success rate of obtaining valid samples, which greatly hampers access to a benchmark dataset.
Besides, the requirement on the uniformity and diversity of samples greatly increases the difficulty of constructing such a benchmark dataset.

\item \textit{Lacking a set of baselines over deep surrogate models from available deep regression methods:} 
Although some prior surrogate models have been presented to advance the HSL-TFP task, most of them mainly focus on general machine learning-based surrogate models. 
Only a few works involve the DNN surrogate modeling for the HSL-TFP task, e.g. \citet{chen2020}. 
Despite that, faced with the requirement of high prediction performance of our new difficult HSL-TFP task, it is urgent to construct a set of DNN surrogate models as baselines for deep research.
\item \textit{Lacking effective evaluation metrics for prediction performance:} Prior works mainly use MAE as the metric to measure the performance of different surrogate models.
However, for this specific task, the MAE cannot comprehensively describe the global or local subtle difference of the prediction for different layout images.
In order to provide a multi-view investigation on DNN surrogates, more effective metrics, which can describe both the local and global difference, are required to assess the predictive capability of different methods.
\end{itemize}

By carefully tackle the above-mentioned issues, in this paper, we establish a new benchmark dataset, that is, heat source layout dataset (HSLD), as well as a set of DNN surrogate models and three types of evaluation metrics, with the purpose of fully advance the research on the HSL-TFP task.
HSLD provides the research community a benchmark resource for the development of the state-of-the-art deep surrogate methods in the HSL-TFP task and the baseline results of deep surrogate models based on available DNN regression models.
In addition, our experiments on HSLD demonstrate that it is quite helpful to reflect the shortcomings of current DNN surrogate models in the HSL-TFP task.
In summary, the following contributions are made in this paper.
\begin{enumerate}
  \item We refine the Temperature Field Prediction of Heat Source Layout (HSL-TFP) task from layout optimization by giving both the flowchart description and formula expression. The HSL-TFP task is first systematically analyzed in this paper as the base task for deep research.
  \item We construct a new large-scale, diversified dataset HSLD for the HSL-TFP task. The dataset is, to the best of our knowledge, the first one which aims for the training and evaluation of DNN regression surrogates. 
  The samples are obtained with high diversity by the proposed two random layout sampling methods and several special layout sampling strategies. 
  These sampling techniques guarantee the accessibility of the large-scale and high-quality dataset.
  The HSLD can provide the research community a better data resource to evaluate and advance the state-of-the-art algorithms in the HSL-TFP task.
  \item We develop a set of representative DNN surrogate models by combining different backbone with available deep regression frameworks for the HSL-TFP task.
  \item Besides, three types of metrics are proposed to evaluate the performance of DNN surrogate models from different aspects, thereby providing the statistical results of these surrogates as baselines to advance other state-of-the-art methods for the HSL-TFP task.
  \item The source codes of our implementation for all the baseline algorithms are released and will be general tools for other researchers.
\end{enumerate}

The rest of this paper is organized as follows. 
The temperature field prediction of heat source layout (HSL-TFP) task is first defined in Section \ref{sec:definition} with three typical heat conduction optimization cases. 
In Section \ref{sec:data}, to handle the difficulty of constrained sampling, we propose two random layout sampling methods, namely the sequence layout sampling and Gibbs layout sampling method, as well as some special layout sampling strategies.
In addition to this, the construction of the HSLD is presented with a comprehensive analysis followed in the same section.
Subsequently, we illustrate the development of ten representative DNN surrogate models for the HSL-TFP task in Section \ref{sec:baseline} and then propose three levels of metrics to evaluate different methods for HSL-TFP task in Section \ref{sec:metric}. 
In Section \ref{sec:experiments}, the evaluation and the comparison of baseline algorithms on HSLD under different experimental settings are given. Finally, we summarize this paper with some discussions and future prospects in Section \ref{sec:conclusions}.


\section{Temperature Field Prediction of Heat Source Layout (HSL-TFP)}
\label{sec:definition}

Consider arranging several electronic components in a specific layout container, and these components generate heat when working. The layout container can be modeled as a two-dimensional layout domain and each component on the container can be simplified as a heat source. The optimization of heat source layout can help to ensure the performance and lifetime of electronic components.

Generally, the mathematical formulation for layout optimization task can be written as
\begin{equation}
	{X}^*=\arg \min\limits_{X} \left( \max(T_{X}) \right)
\end{equation}
where ${X}$ is the layout scheme of the components and $T_{X}$ is the corresponding temperature field under the layout ${X}$.
The discrete description of the temperature field under a grid system is defined as the temperature matrix denoted by $Y$ below.
Therefore, the key process of the layout optimization task is to obtain the corresponding temperature field of different layouts. 

Actually, the steady temperature field $T$ over the domain for a two-dimensional heat conduction problem satisfies Poisson's equation, which is formulated as
\begin{equation}
\begin{split}
& \frac{\partial}{\partial x}\left(k \frac{\partial T}{\partial x}\right)+\frac{\partial}{\partial y}\left(k \frac{\partial T}{\partial y}\right)+\phi(x, y)=0 \\
& {\text{Boundary: }} {\quad T=T_{0} \quad \text{or} \quad 
	k \frac{\partial T}{\partial \mathbf{n}}=0 \quad} \\
& \qquad \qquad \qquad {\text{or} \quad k \frac{\partial T}{\partial \mathbf{n}}=h\left(T-T_{0}\right)}
\label{eq1}
\end{split}
\end{equation}
where $\phi(x,y)$ denotes the intensity distribution function affected by the heat source layout $X$, and $k$ is the thermal conductivity of the domain which is set to 1 in this study.
Below the governing equation, the accompanying equations represent three boundary conditions for unique solutions, including Dirichlet boundary conditions (Dirichlet BCs) where $T_0$ is the isothermal boundary temperature value, Neumann boundary conditions (Neumann BCs) where zero heat flux is exchanged, and Robin boundary conditions (Robin BCs) where $T_0$ denotes the surrounded fluid temperature value and $h$ defines the convective heat transfer coefficient.

\begin{figure}[!htbp]
	\centering	
	\includegraphics[width=1\linewidth]{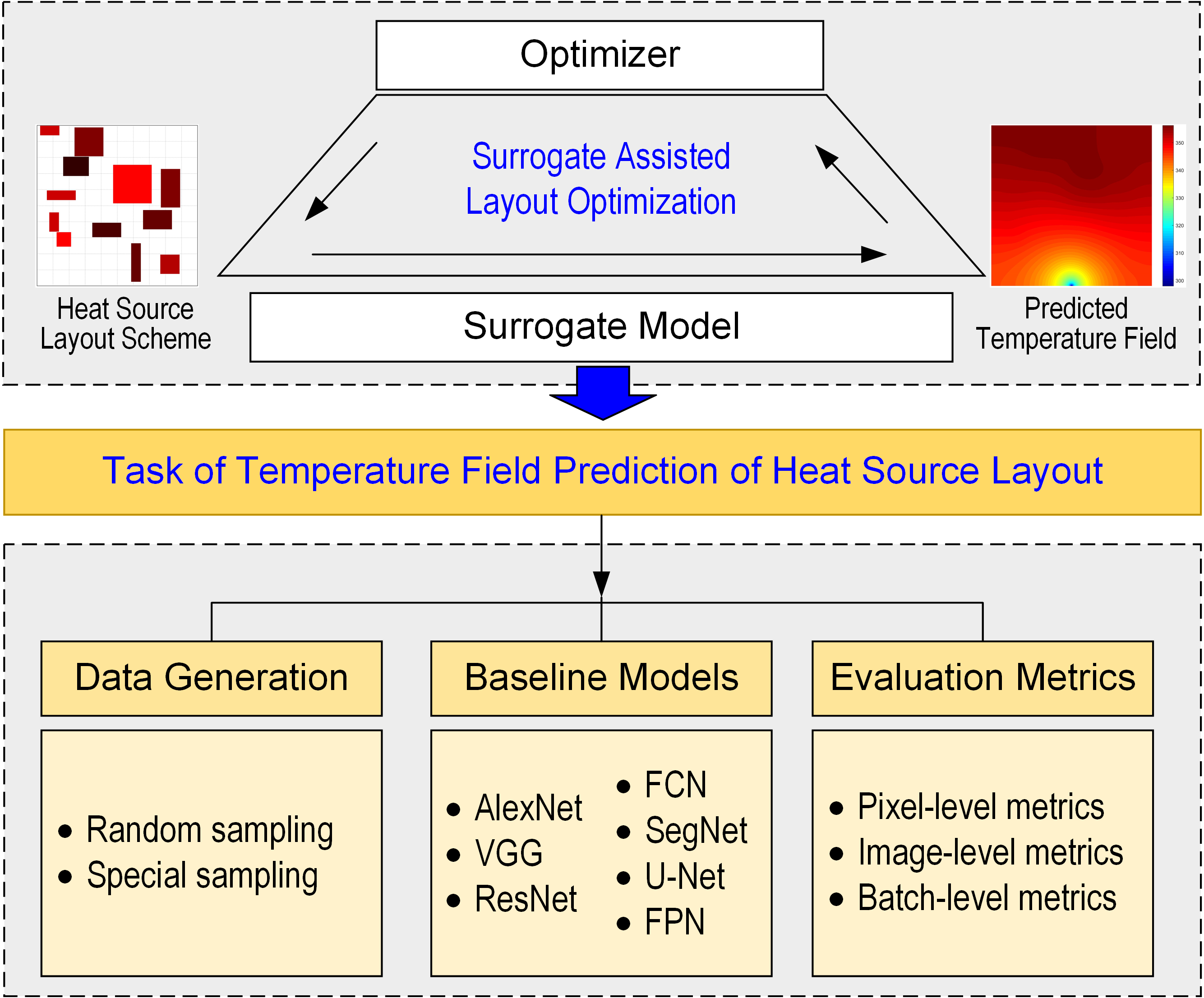}
	\caption{The overview of the HSL-TFP task.}
	\label{fig:hsl_tfp}
\end{figure}

Due to the important role of predicting the temperature field of a given heat source layout in layout optimization task, we refine the temperature field prediction of heat source layout (HSL-TFP) task as the base task for deep research in this work (see details in Fig. \ref{fig:hsl_tfp}).
As the figure shows, the aim of the HSL-TFP task is to explore effective and efficient prediction methods for further analysis in layout optimization tasks, in order to replace the time-consuming numerical simulation process and achieve nearly real-time prediction.

For further research on the HSL-TFP task, 
two commonly used heat conduction problems are considered, namely the volume-to-boundary (VB) problem and the
volume-to-point (VP) problem. The VP heat conduction optimization is a fundamental problem for electronic cooling, which considers a finite heat-generating volume cooled by a small patch of heat
sink on the boundary. For the VB problem, the condition of each boundary can be isothermal,
adiabatic, or convective.

In this work, the layout container is modeled as a square domain with the size of $0.1m \times 0.1m$.
Then, we conduct research on three specific cases, namely the VB problem with identical condition boundaries, the VB problem with different condition boundaries, and the VP problem.

\begin{figure}[!htbp]
	\centering	
	\subfigure[Case 1]{
		\includegraphics[width=0.3\linewidth]{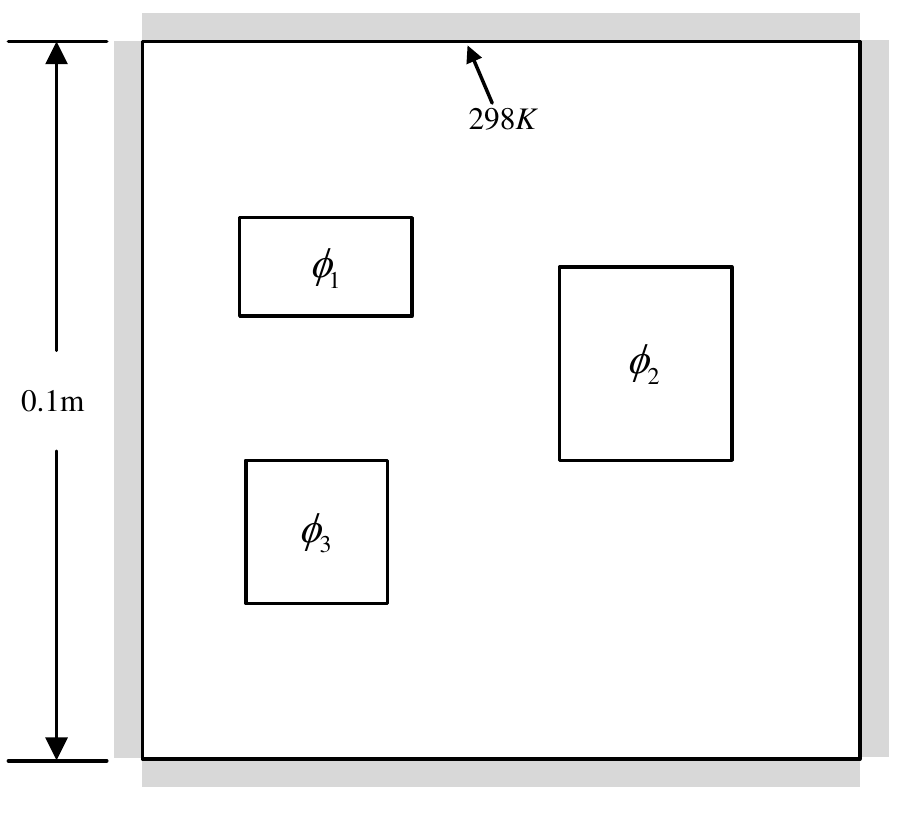}
		\label{fig:case1}
	}
	\subfigure[Case 2]{
		\includegraphics[width=0.3\linewidth]{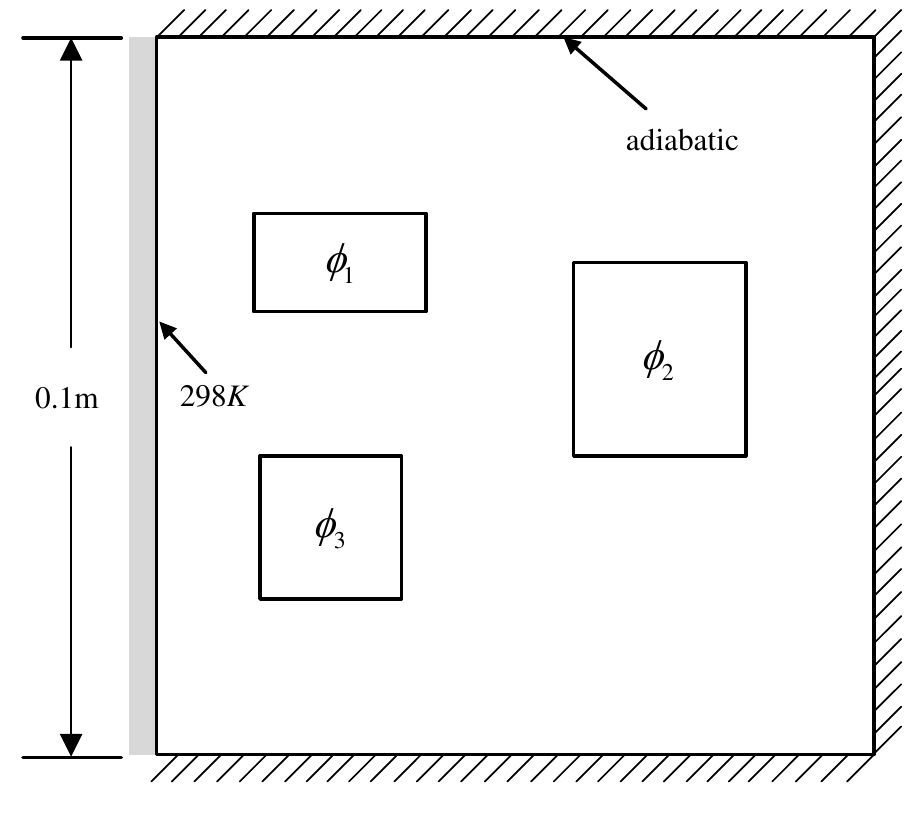}
		\label{fig:case2}
	}
	\subfigure[Case 3]{
		\includegraphics[width=0.3\linewidth]{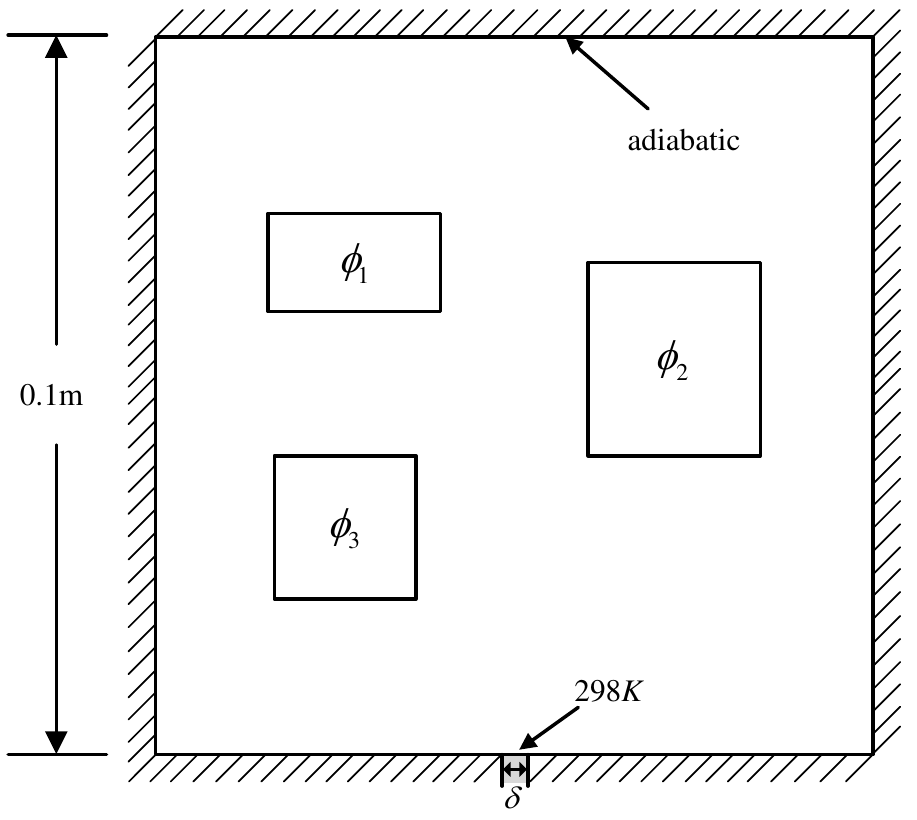}
		\label{fig:case3}
	}
	\caption{Schematic of the computational domain for the three specific problems included by the HSLD.}
	\label{fig:case}
\end{figure}

{\bf Case 1:} For the VB problem with identical boundary conditions, all the boundaries are isothermal with constant temperature valued 298K (see Fig. \ref{fig:case1} for better understanding).

{\bf Case 2:} For the VB problem with different boundary conditions, one boundary is set to constant temperature valued 298K (Dirichlet BC) and the others are adiabatic (Neumann BC). Fig. \ref{fig:case2} shows this type of VB problem.

{\bf Case 3:} For the VP problem, the width of the heat sink $\delta$ is set to $0.001m$, with a constant temperature valued 298K (Dirichlet BC). The rest boundaries are subject to Neumann BC. Fig. \ref{fig:case3} describes the VP problem of this type.

Based on these cases, this work constructs a new diversity, large-scale dataset which is named as heat source layout dataset (HSLD) to advance the state-of-the-art methods for the HSL-TFP task. Following we will introduce the detailed construction of the HSLD and further develop a set of baseline deep surrogate methods for evaluation.

\section{Generating HSLD: A New Data Set for HSL-TFP}
\label{sec:data}

In this section, one heat source layout problem in the two-dimensional layout plane is defined to construct the data set with the specific information of 12 layout components (heat sources) being provided in Table \ref{table:component}.
All the components are of rectangular shape, as well as the layout container.
As the former section defines, the size of the layout container is $0.1\text{m} \times 0.1\text{m}$ with its side length being denoted by $L = 0.1\text{m}$.
To simplify the problem but without loss of generality, the placement angles of components keep fixed and only the translation of them is allowed in our data set.

The layout domain is meshed to a $200 \times 200$ grid and the components are also discretized in this grid system. 
The minimum distance that the component can move by is $1/200$ of $L$, which is the side length of one cell.
Hence, even though the component can only translate cell by cell, it can be regarded as a continuous layout problem.
Following the formulation of HSL-TFP, the layout scheme as the input is represented as a $200 \times 200 $ image matrix, where the value of cells that are occupied by components is set as their intensity.

\begin{table}
  \centering
    \caption{The characteristics of layout components in our data set: HSLD}
    \label{table:component}
  \begin{tabular}{ c c c c}
    \hline
  Component & Length(m) & Width(m) & Intensity(W/m$^2$) \\
	\hline
     1 & 0.016 & 0.012 & 4000 \\
     2 & 0.012 & 0.006 & 16000 \\
     3 & 0.018 & 0.009 & 6000 \\
     4 & 0.018 & 0.012 & 8000 \\
     5 & 0.018 & 0.018 & 10000 \\
     6 & 0.012 & 0.012 & 14000 \\
     7 & 0.018 & 0.006 & 16000 \\
     8 & 0.009 & 0.009 & 20000 \\
     9 & 0.006 & 0.024 & 8000  \\
    10 & 0.006 & 0.012 & 16000 \\
    11 & 0.012 & 0.024 & 10000 \\
    12 & 0.024 & 0.024 & 20000 \\
    \hline
  \end{tabular}
\end{table}

%

\subsection{Two Random Layout Sampling Methods}

In the layout problem, one basic geometry constraint which should be strictly satisfied is the non-overlapping constraint.
This constraint requires no overlap between different components and also no protrusion between components and the layout container.
Only the layout scheme that meets the non-overlapping constraint can be defined as the feasible one.
It means that the layout sample that is geometrically feasible can be regarded as the valid sample.
Therefore, this layout sampling problem is a constrained sampling problem. 

There are mainly two types of approaches to deal with the constraint in realizing the random sampling method.
One is to passively handle the non-overlapping constraint. 
Specifically, the acceptance-rejection sampling method can be applied to this problem.
Firstly, we can randomly generate the layout sample without considering the non-overlapping constraint. 
Then make a judgment whether the generated layout sample satisfies the constraint.
If it does, this sample is valid and can be saved. Otherwise, it will be discarded.
However, taking our case study as an example, experimental results show that only a few valid samples can be obtained after repeating the unconstrained layout sampling process 30000 times.
It means there is a much higher probability to sample invalid layout schemes than valid ones.
It illustrates that the acceptance-rejection sampling method maintains a very low efficiency, especially when the spatial density of components in the container is high.

Hence, to efficiently collect massive valid samples, we must resort to the other approach, which is to actively incorporate the non-overlapping constraint with the layout sampling process. 
Simultaneously, the developed sampling method should satisfy the randomness of sampling within the feasible layout space in order to guarantee that any one feasible layout scheme can be acquired in this way.
Based on the above two considerations, two random layout sampling methods are proposed to efficiently and randomly generate valid layout samples: \textit{sequence layout sampling} (SeqLS) method and \textit{Gibbs layout sampling} (GibLS) method.

\subsubsection{Sequence layout sampling (SeqLS) method}

The main idea of the SeqLS method is to add components in the layout domain one by one and place each component randomly within its current feasible layout region.
The feasible layout region for one component is defined as the union of all available layout positions where there exists no overlap with the other components and the layout container after placing this component.
It should be noticed that the feasible layout region for one component relies on not only the layout container but also the components that have been placed before this one.
Hence, the feasible layout region differs every time.
Besides, the feasible layout region should be the union of multiple discrete pieces of feasible layout space which results from the satisfaction of the non-overlapping constraint with the other components.
Thus it is very difficult to directly describe the varying feasible layout region in a unified approach, not to mention realize the random layout sampling.

To alleviate this difficulty, we propose to firstly discretize the layout domain to a grid system with the purpose of approximately representing the complicated two-dimensional feasible layout region with a finite number of position points.
This approximation is reasonable in part because the layout scheme is processed as an image matrix when finally being fed into the prediction model. 
The other reason is that the grid with a sufficient density can guarantee the description precision, maintaining the same idea as implemented in the finite-element method.
Every grid coordinate can be regarded as one available location for placing the component.
Then we propose to construct the virtual energy matrix (VEM) of one component to indicate where the component occupies, which is denoted by VEM$_i$. 
The elements of VEM correspond to grid nodes and thus VEM maintains the same size as the number of grid nodes. 
The value of the element in VEM is set as 1 when the corresponding grid node is covered by the component, otherwise 0.
The VEM of the layout container, which is denoted by VEM$_0$, is defined as the zero-value matrix.
Besides, the extended VEM (eVEM) is also defined to describe the relative position relationship between two components, which is denoted by eVEM$_{ij}$.
On the basis of the VEM of one component $i$ that has been placed in the layout domain, the value of the elements in eVEM will be set as 1 if there is overlap when the corresponding grid points are occupied by the other component $j$. 
By this definition, the positions where their values are 0 represent the feasible layout region where the non-overlapping constraint between component $i$ and component $j$ can be satisfied.

Assume that after $k$ components ($i = 1,2,..., k$) have been placed in the layout domain, component $j$ will be randomly placed in the feasible layout region. 
Following the definitions above, the feasible layout region of component $j$ can be equivalently described by the zero-value positions of the integrated eVEM (IeVEM), which is calculated as
\begin{equation}
	\text{IeVEM}_j = \sum_{i=0}^{k} \text{eVEM}_{ij}
\end{equation}
where component $i=0$ stands for the layout container. Therefore, the random placement of component $j$ can be easily realized by randomly choose one zero-value position in IeVEM$_j$ for component $j$. 
By this way, the two-dimensional feasible layout region can be approximated equivalent to be finite feasible layout position candidates, which makes the random layout sampling easier and more achievable.

\begin{figure}[!htbp]
	\centering	
	\includegraphics[width=0.7\linewidth]{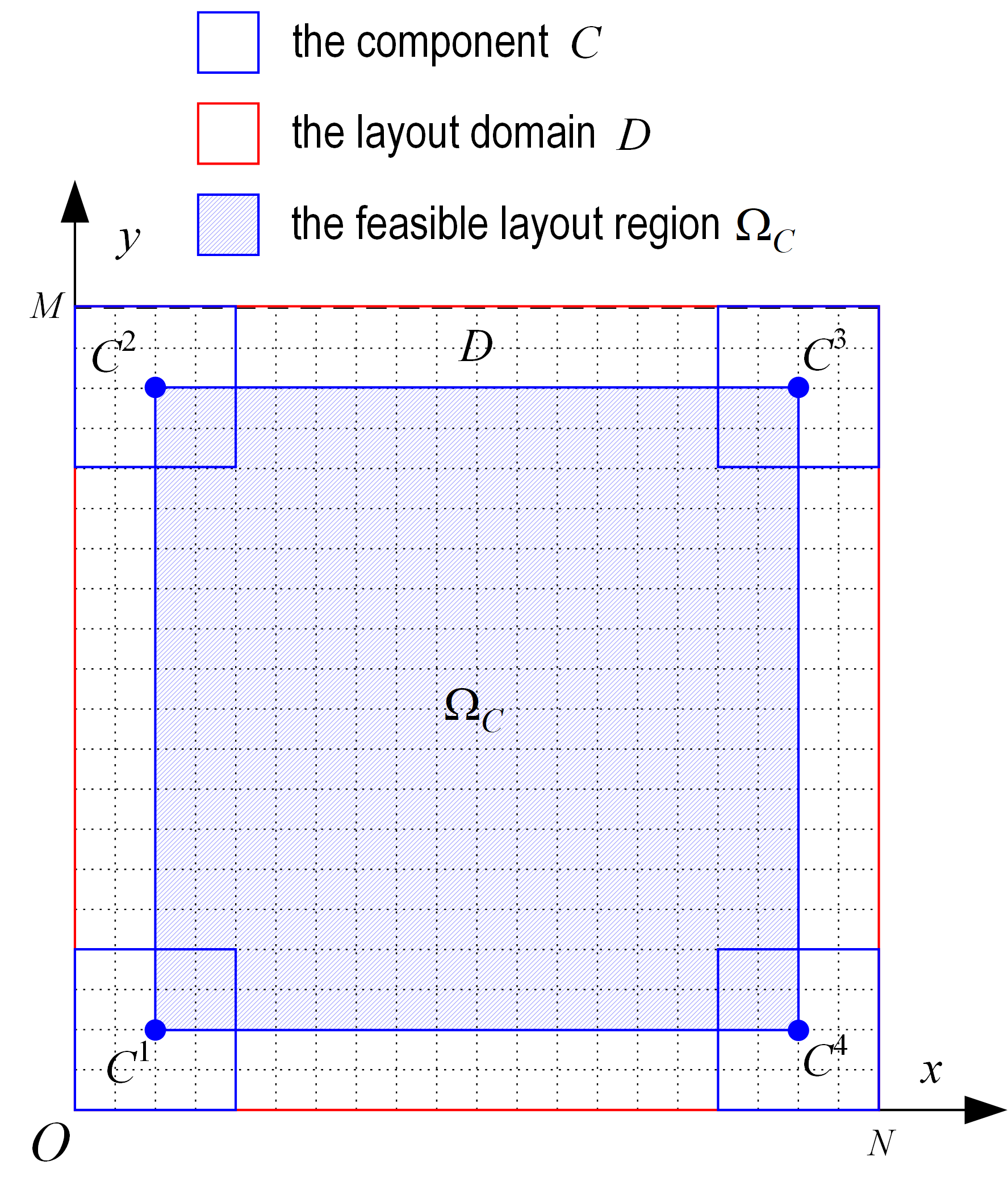}
	\caption{The illustration of the feasible layout region $\Omega_C$ of component $C$ in the layout container.}
	\label{fig:SeqLS_1}
\end{figure}

\begin{figure}[!htbp]
	\centering	
	\includegraphics[width=0.7\linewidth]{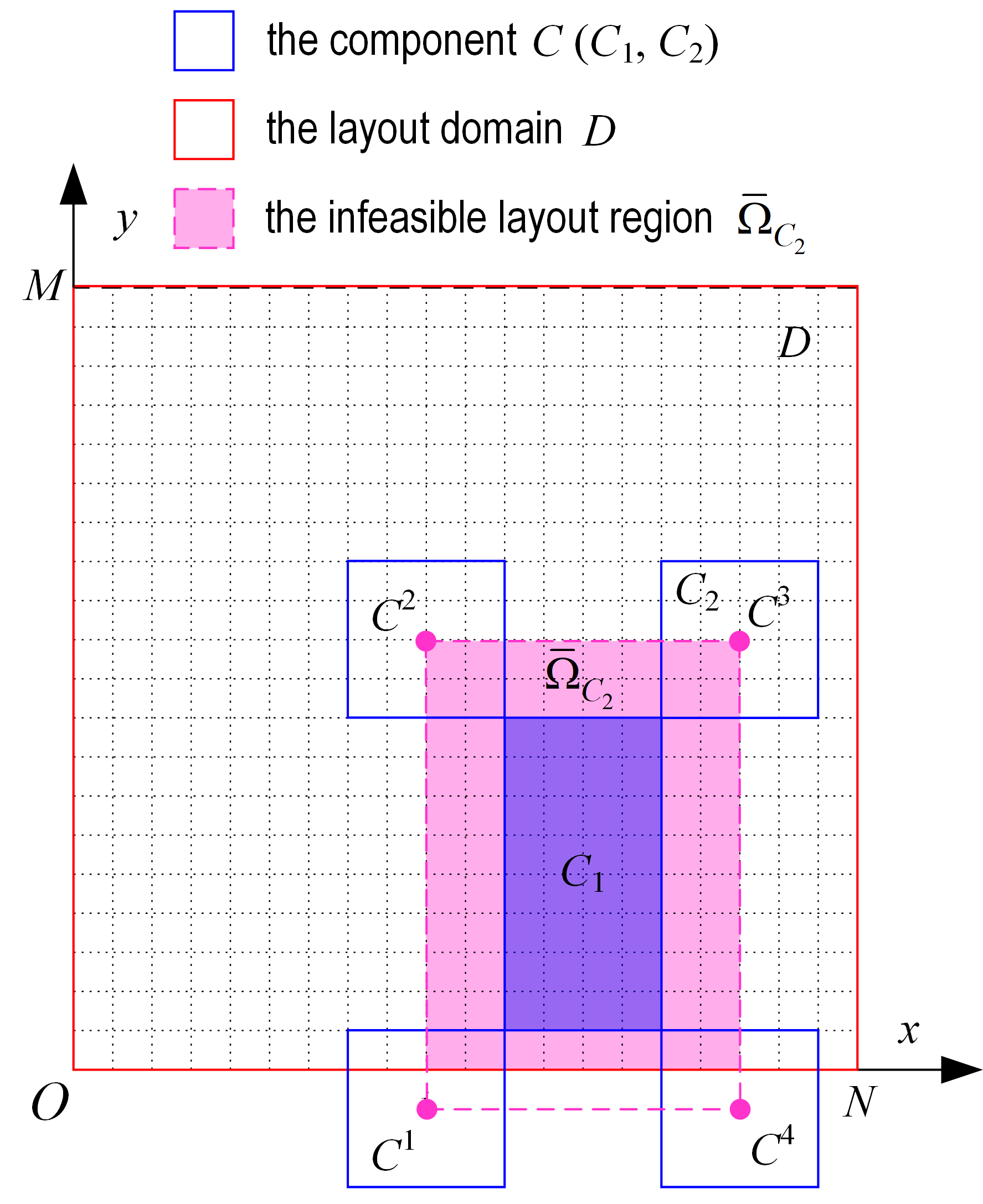}
	\caption{The illustration of the feasible layout region $\Omega_{C_2} = D - \overline{\Omega}_{C_2} $ of component $C_2$ in the layout domain containing component $C_1$}
	\label{fig:SeqLS_2}
\end{figure}

The main framework of the SeqLS method is described in Algorithm \ref{alg:SeqLS}. 
The layout sequence $S$ can be randomly generated also. 
It is notable that this sequence will not harm the randomness of the final generated layout sample as every component is totally randomly placed within its feasible layout region.
However, it will influence the sampling success rate as if the smaller components are placed first, there is a higher probability that there might be no available space for the bigger ones. 
Consequently, the layout sequence is usually determined as the descending order of the area of components to guarantee a higher success rate of random sampling.

Taking the layout problem with $N_s$ rectangular components as an example, the reference point to determine the position of one component is chosen as the center of its shape.
The whole layout domain $D$ is discretized into a $N \times M$ grid system and the grid size of components can be calculated easily.
Thus the size of the VEM is $(N+1) \times (M+1)$, as well as the eVEM. 
The identification of the feasible layout region $\Omega_C$ of the component $C$ to be placed in the layout container is illustrated in Fig. \ref{fig:SeqLS_1}.
As long as the central point of the rectangular component is placed in the region $\Omega_C$ (including the boundary), the non-overlapping constraint between the component and the container can be guaranteed to be satisfied.
According to the definition of eVEM, the value of elements of eVEM$_{0C}$ in the region $\Omega_C$ should be set as 0 and the other should be set as 1.
As illustrated in Fig. \ref{fig:SeqLS_2}, it is easier to describe the infeasible layout region $\overline{\Omega}_{C_2}$ of component $C_2$ in the layout domain containing component $C_1$ only.
Hence, the value of eVEM$_{C_1C_2}$ in this region should be set as 1 and the other should be set as 0.

\begin{algorithm*}[!htbp]
	\caption{The sequence layout sampling (SeqLS) method}
	\label{alg:SeqLS}
	\LinesNumbered
	\KwIn{The size of $N_s$ components and layout domain, the mesh size $N \times M$}
	\KwOut{One randomly generated layout sample}
	
	Initialize the layout sequence $S$ \\
	Mesh the layout domain as a $N \times M$ grid system \\
	Mesh the components in the $N \times M$ grid system and use grid coordinates to describe their positions\\ 
	Initialize index $i=1$ and the first component to be placed is component $i$ \\
	Initialize the VEM of layout container: VEM$_0$ = zeros($N+1$, $M+1$) \\
	Initialize the VEM of all components: VEM$_i$ = zeros($N+1$, $M+1$) $(i=1,2,...,N_s)$ \\
	\tcp{Randomly place components one by one}
	\For{$i=1:N_s$}{
		Identify the feasible layout region of component $i$ in the layout container \\
		Set up the value of eVEM$_{0i}$ \\
		\While{$i \geq 2$}{
			\For{$ j = 1:(i-1) $}{
				Initialize eVEM$_{ji}$ = VEM$_j$ \\
				Identify the feasible layout region of component $i$ in the layout domain containing component $j$ only\\
				Set the value of eVEM$_{ji}$ \\
			}
		}
		Calculate the integrated eVEM of component $i$: $\text{IeVEM}_i = \sum_{j=0}^{i-1} \text{eVEM}_{ji}$ \\
		Find all the zero-value elements and calculate the number of them as $N_0$ \\
		\eIf{$N_0 > 0$}
		{
			Randomly choose one zero-value position in IeVEM$_i$ as the location of the center of component $i$ \\
			Set the value of VEM$_i$ according to the location of component $i$ \\
			Transform the grid coordinates to real coordinates \\
		}
		{
			\tcp{This indicates an unsuccessful layout sampling.}
			Restart this sampling process \\
		}
	}
	Record the coordinates of components and return the sampled layout scheme.
\end{algorithm*}

\subsubsection{Gibbs layout sampling (GibLS) method}


As stated in the previous part, the main difficulty of performing the random layout sampling is the description of the two-dimensional feasible layout region of components under the non-overlapping constraint so that we cannot directly draw samples from the complicated layout space.
Hence, we propose to use a Markov chain Monte Carlo (MCMC) sampling technique to solve this constrained sampling problem.
To be specific, we follow the idea from Gibbs sampler \citep{2006Pattern} and convert the high-dimensional constrained layout sampling into a sequence of one-dimensional constrained conditional sampling processes.
For this reason, we refer to the proposed method as \textit{Gibbs layout sampling} (GibLS).
More precisely, when performing conditional sampling, only one coordinate variable of one component is considered as varying while the other components and the other coordinate of this component are kept fixed.
Therefore, the feasible layout region for sampling consists of finite pieces of one-dimensional segments under the non-overlapping constraint, which can be readily described and sampled in a continuous way.
Apart from this, it should also be noticed that any layout scheme can be sampled by the conditional distribution given another layout as long as these two layouts can be converted to each other by translating components within the layout domain and without leaving this plane. 

\begin{figure}[!htbp]
	\centering	
	\includegraphics[width=0.7\linewidth]{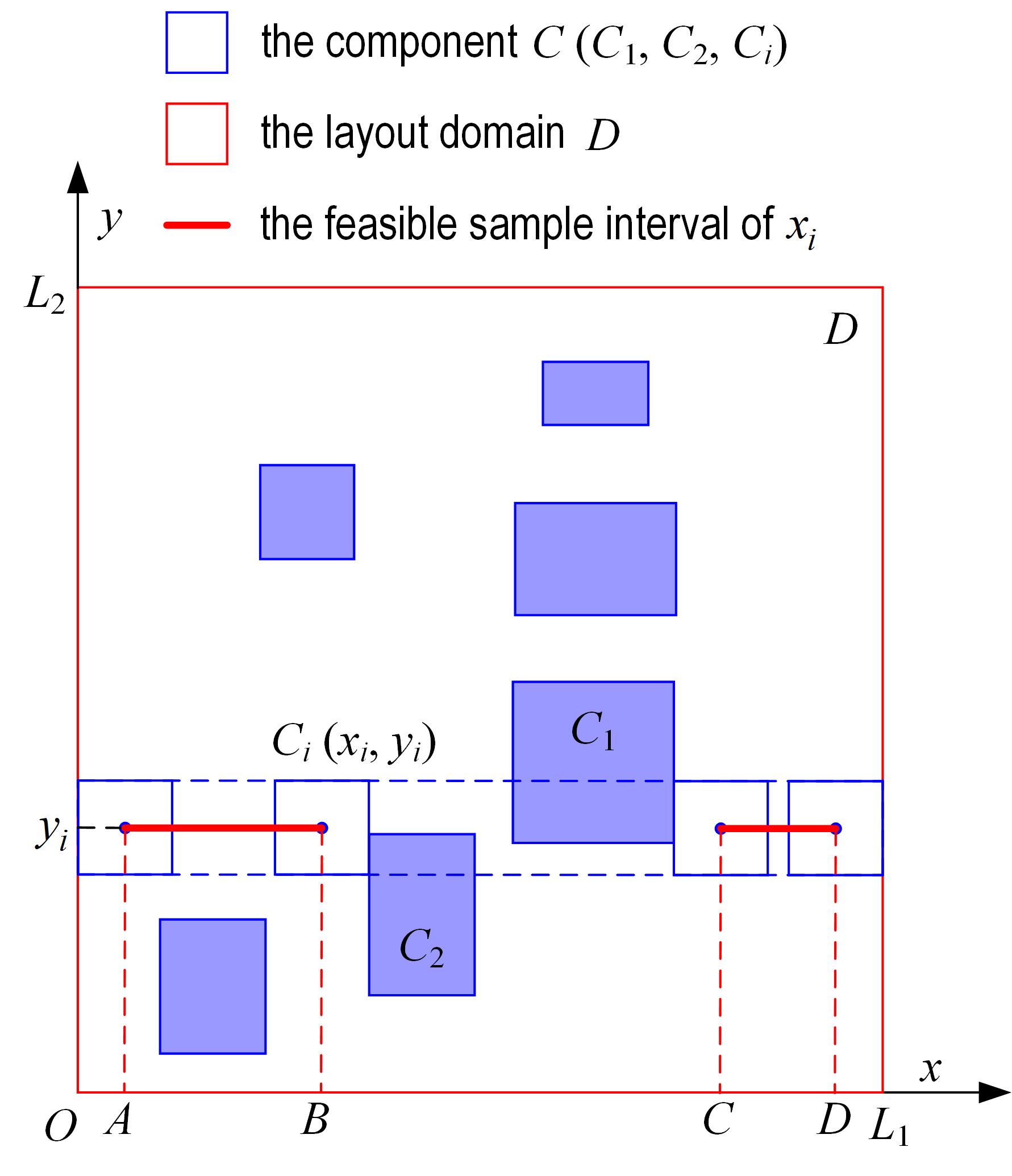}
	\caption{The illustration of the feasible sample interval of component $C_i$ during one-dimensional conditional sampling of variable $x_i$}
	\label{fig:GibLS}
\end{figure}

In a two-dimensional layout domain, the coordinates of the reference points of layout components are used to determine their positions, and then $2N_s$ variables are required to represent the layout scheme with $N_s$ components.
The layout scheme $X$ can be written as
\begin{equation}
\begin{aligned}
	X &= \left\lbrace {x_1, y_1, x_2, y_2, ..., x_{N_s}, y_{N_s}} \right\rbrace \\
	&\stackrel{\triangle}{=} \left\lbrace  {x_1, x_2, ..., x_{2N_s-1}, x_{2N_s}} \right\rbrace 
\end{aligned}
\end{equation}
where $(x_i, y_i)(i=1,2,...,N_s)$ represent the coordinates of component $i$.

\begin{algorithm*}[!htbp]
	\caption{The Gibbs layout sampling (GibLS) method}
	\label{alg:GibLS}
	\LinesNumbered
	\KwIn{The size of $N_s$ components and layout domain, the initial layout $X_0$, the burn-in period $B$, the interval period $I$ }
	\KwOut{A collection of $n$ randomly generated layout samples:  $\bm{X} = \left\lbrace { X_i | i=1,2, ..., n } \right\rbrace $}
	
	Initialize the starting point as $X_0 = \left\lbrace {x_1^0, x_2^0, ..., x_{2N_s-1}^0, x_{2N_s}^0} \right\rbrace $ \\
	Initialize the iteration index: $k = 0$ \\
	Initialize the number of the generated samples: $n_s = 0$ \\
	\While{$n_s < n$}
	{
		\tcp{Gibbs updates}
		\For{$ i=1:2N_s $}{
			Identify the feasible layout region of variable $x_i^k$ under the non-overlapping constraint \\
			Sample $x_i^{k+1} \sim p(x_i | x_1^{k+1}, ..., x_{i-1}^{k+1}, x_{i+1}^k, ..., x_{2Ns}^k)$
		}
		Obtain one new layout sample: $X_{k+1} = \left\lbrace {x_1^{k+1}, x_2^{k+1}, ..., x_{2N_s-1}^{k+1}, x_{2N_s}^{k+1}} \right\rbrace $ \\
		$ k = k + 1 $ \\
		\eIf{$ k \leq B $}
		{
			\tcp{Discard the first $B$ samples}
			continue the loop
		}
		{
			\tcp{Save one sample every $I$ iterations}
			\If{$mod(k-B-1,I) = 0$ }{
				$n_s = n_s + 1 $  \\
				Save the new layout sample $X_k$ as $X_{n_s}$
			}
		}
	}
	Return the collection of $n$ layout samples $\bm{X}$.
\end{algorithm*}

The algorithm flowchart of the GibLS method is described in Algorithm \ref{alg:GibLS}. 
One iteration is defined where each variable has been sampled once and one new layout sample is produced, which is referred to as \textit{Gibbs updates}.
In our case, $2N_s$ samplings are required.
Following the instruction of Gibbs sampling, a burn-in period $B$, which includes $B$ iterations, is used to reduce the influence of choosing different starting points and reach a detailed balance, guaranteeing the performance of the GibLS algorithm. 
In other words, the first $B$ samples should be discarded.
During the one-dimensional conditional sampling, the feasible layout region of one single variable should be identified first according to the non-overlapping constraint. 
As illustrated in Fig. \ref{fig:GibLS}, the feasible sample interval of $x$-axis coordinate $x_i$ of component $C_i$ is the union of two pieces of segments: $A B$ and $C D$. 
In this way, the conditional sampling of variable $x_i$ can be easily implemented.

Besides, it can be easily known that the layout sample $X_k$ depends on $X_{k-1}$. 
When the feasible sample intervals for most variables are small, the values of these variables are very close, causing that two layout samples generated subsequently will seem very similar.
Thus, to enhance the diversity of sampling, another parameter, the interval period $I$, is introduced to control the sample distance.
That is, the new sample is saved as an effective one once every $I$ iterations, as described in Algorithm \ref{alg:GibLS}.

\subsubsection{A comparative discussion on the SeqLS and GibLS method}

In this part, a comparative discussion on the two above sampling methods is provided with the purpose of illustrating their respective advantages and disadvantages.

The proposed two layout sampling techniques, SeqLS and GibLS, are both proven to be effective and efficient in our study. 
Despite the fact that the involved components are of rectangular shapes in this layout problem, it is worth noticing that our random sampling methods are also feasible and applicable to those cases involving components with complex shapes.
One challenging step embedded in this process is to determine the non-overlapping region between two complex components when actively incorporating the constraint, as done in Fig. \ref{fig:SeqLS_2} and Fig. \ref{fig:GibLS}.
However, this problem can be handled by using some advanced geometry modeling methods, such as the non-fit polygon method \citep{Bennell2000a} and the phi-function method \citep{Chernov2010, Chen2021}.

\begin{figure}[!htbp]
	\centering
	\includegraphics[width=\linewidth]{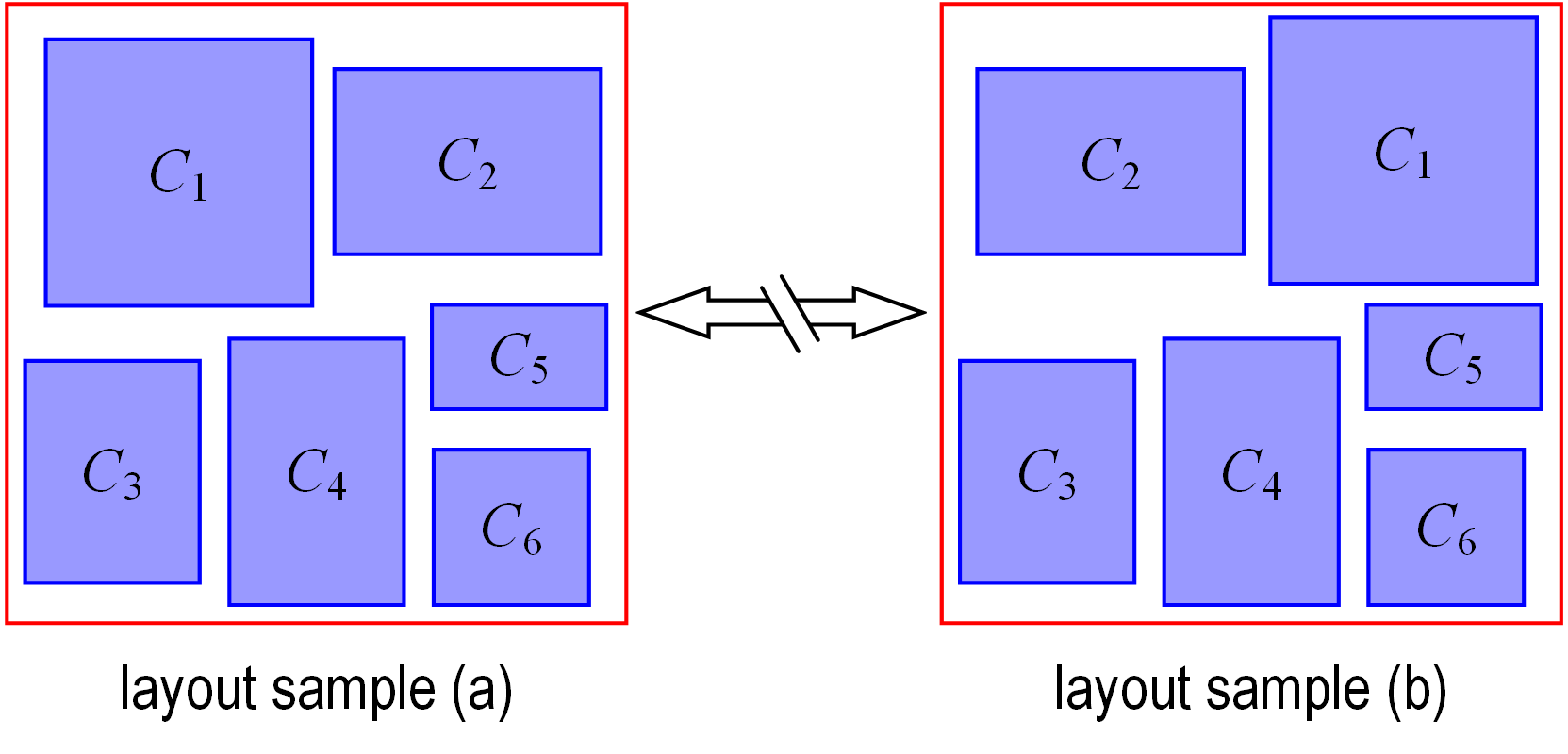}
	\caption{The limitation of sampling two compact layout samples using the GibLS method}
	\label{fig:GibLS_Compact}
\end{figure}

Although there is no restriction on the shapes of layout components theoretically, the efficacy of random layout sampling methods partly depends on the spatial layout density. 
Notably, when the spatial layout density is high, which means a relatively compact layout problem, the success rate of the SeqLS method will become low as feasible layout samples are much difficult to be randomly acquired.
Regardless of this aspect, it is likely to sample different valid layout schemes using the SeqLS method as long as they exist.
However, on the contrary, it is much less likely to obtain the diversified layout samples in such a compact layout domain by using the GibLS method.
For example, two compact layout samples ((a) and (b)) are displayed in Fig. \ref{fig:GibLS_Compact} and the only difference is that component $C_1$ and $C_2$ exchange their positions with each other.
If we take sample (a) as the starting point and follow the GibLS procedure, component $C_1$ only can move in its small surrounding local area during sampling and there is no possibility to draw layout sample (b) from the constrained conditional distribution.
This means that when two samples are not connected, they cannot be obtained by the state transition process.
Thus, the proposed GibLS method can hardly generate diversified layout samples in this situation and the SeqLS would be more suitable.

In addition, the efficacy of the GibLS method is also affected by the sample interval period $I$.
If we want to reduce the similarity of two saved samples, the interval period $I$ should be set as a larger number, in turn lowering the sampling efficiency and increasing the consumption time. 
Due to the fact that the SeqLS method simulates the manual sequential placement, each sampling is totally independent, ensuring the sampling diversity.

Furthermore, by contrast with the continuous sampling in each step of the GibLS method, it is also notable that the SeqLS is realized on the basis of the discretization description of the layout domain and components, causing a two-fold drawback.
One is that some layout samples might be missed due to the existence of the approximation error.
The other is that, compared to the GibLS method, it will cost more memory owing to the storage of VEMs and eVEMs.
This might enable different applications in other scenarios.

\subsection{Special Layout Sampling Strategies}
\label{sec: special samples}

In the previous part, two random layout sampling techniques are introduced.
The SeqLS and GibLS method tend to generate layout schemes where components are dispersed within the whole layout container. 
Although, theoretically, any feasible layout scheme can be sampled by these two methods, there is still a low probability to generate enough samples with specific patterns that designers might concern in a limited number of samplings, such as the compact layout scheme where all components closely gather together.
If the training and testing are both based on random layout samples, the assessment on the surrogate will be one-sided and the results cannot reflect the real prediction performance on the entire layout space.
This statement has been verified in our previous research \citep{chen2020}. 
In order to evaluate different surrogates thoroughly, some special samples are intentionally proposed to be generated following several special heuristic layout rules, serving as the diversified test data.

Three types of special layout samples are summarized here: \textit{corner samples}, \textit{group samples} and \textit{part-space samples}.
Corner samples are observed with more worse prediction performance and thus specially generated as test data.
Group samples are intentionally designed to demonstrate the performance of the neural network-based surrogate predicting the composed object which happens due to several touching components with the same intensity.
Part-space samples are defined to represent compact layout schemes that are usually encountered in the surrogate-assisted layout optimization.

\medskip
\noindent
\textbf{{Corner samples}}

\begin{figure}[!htbp]
	\centering	
	\includegraphics[width=1\linewidth]{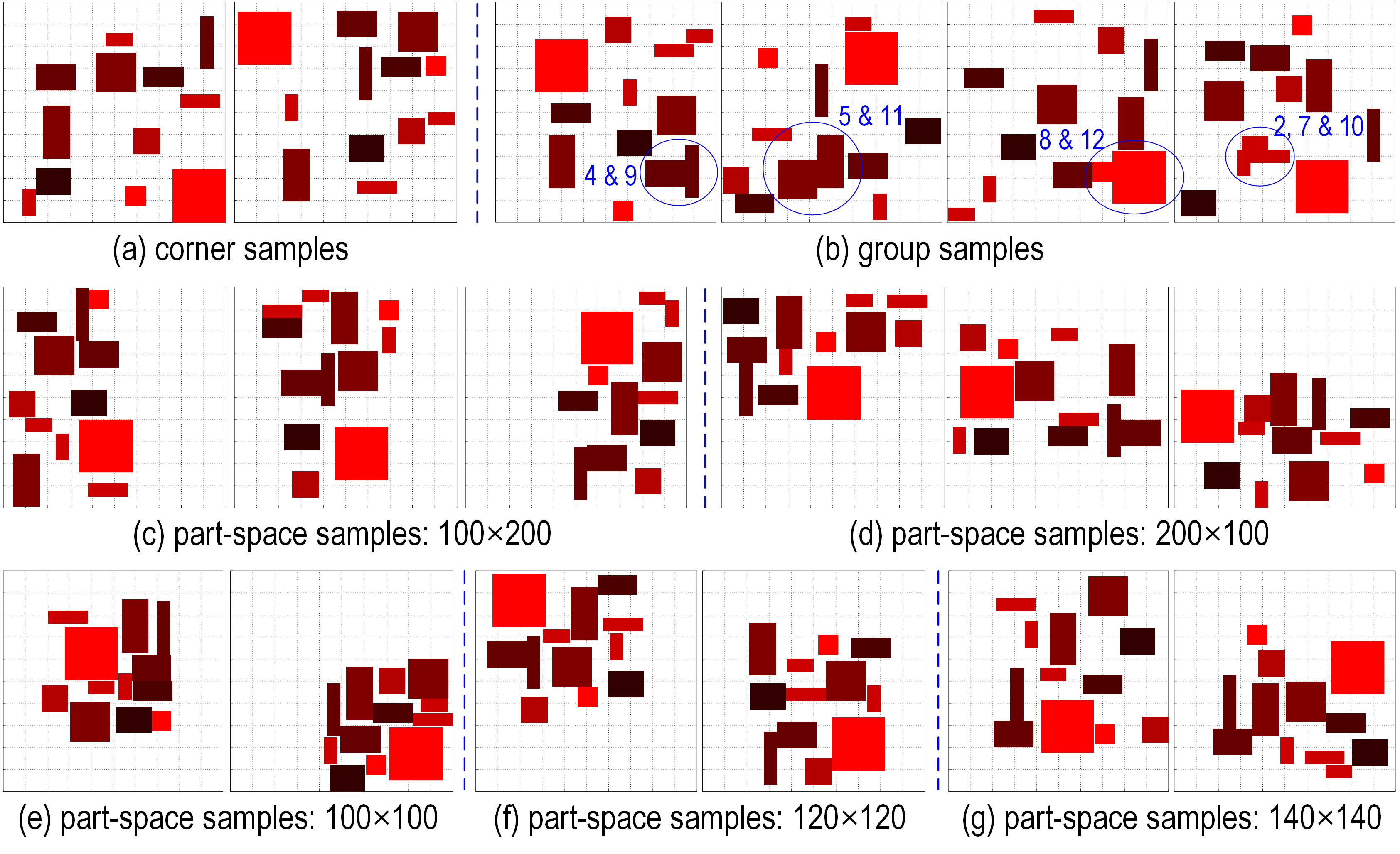}
	\caption{Three types of special samples: corner samples (a), group samples (b), part-space samples (c-g)}
	\label{fig:specialsamples}
\end{figure}

The corner sample is defined that the component with the maximum area and intensity, which is component 12 in our case, is required to be placed around four corners of the layout domain while the other components can be placed randomly.
Two examples of this kind are shown in Fig. \ref{fig:specialsamples}(a).
Empirically, the temperature field predictions on these corner samples are less accurate than those on other random samples.

\medskip
\noindent
\textbf{{Group samples}}

The group sample requires that the components with the same intensity are placed closely next to each other so that they are just touching. 
The procedure to generate this kind of sample is to first randomly place one component and then randomly place the next one surrounding the boundary of the previous one.
When there are three components that have the same intensity, the third one is randomly placed along the union boundary of the first two components. 

This special sample is defined based on the consideration that the combination of these touching components in the input layout image will make some new complex shapes, which might be challenging to be recognized by neural networks.
In other words, these group samples might bring a higher requirement on the neural network surrogate modeling. 

In our problem, there are four groups of components that meet the requirement. 
Three two-component groups include components 4 and 9, 5 and 11, and 8 and 12. 
One three-component group includes components 2, 7 and 10.
Some examples are provided in Fig. \ref{fig:specialsamples}(b). 

\medskip
\noindent
\textbf{{Part-space samples}}

The part-space sample is proposed to represent the compact layout scheme where the components are restricted to be placed in a reduced layout space that is some part of the layout domain.

We firstly define a reduced layout space as one-half of the layout domain, which covers $100 \times 200$ or $200 \times 100$ cells.
The reduced layout space varies from the left to the right, or from the top to the bottom every 20 cells.
Six samples for each reduced layout space are presented in Fig. \ref{fig:specialsamples}(c) and (d).
The central points of rectangular components are required to lie in this space and their specific positions are randomly sampled.
In addition, we define another several reduced layout spaces as $100 \times 100$, $120 \times 120$ and $140 \times 140$. 
The locations of these reduced spaces are also randomly sampled from the $200 \times 200$ layout domain.
Two examples in each sub-space are provided in Fig. \ref{fig:specialsamples}(e-g).

It should be emphasized that the generation of these special samples is realized by incorporating these heuristic layout sampling rules with the SeqLS method and it is easier to implement this modification than using the GibLS method.
Moreover, with the reduction of the available layout space, that is, the increase of the spatial layout density, it is much harder to sample a valid layout scheme, resulting in a lower success rate of sampling.
When the layout space is restricted in a $100 \times 100$ grid, the SeqLS process needs to be performed 2042 times for generating 1000 valid layout schemes in our trial, causing a success rate of 48.9$\%$.
This number becomes around 88.3$\%$ and 96.1$\%$ for $120 \times 120$ and $140 \times 140$, respectively.
In this situation, it would be much more inefficient or even impossible to use the GibLS method to generate diversified samples based on the conditional sampling.

\subsection{Temperature Field Calculation}

\begin{figure}[!htbp]
	\centering	
	\subfigure[layout]{\includegraphics[width=0.45\linewidth]{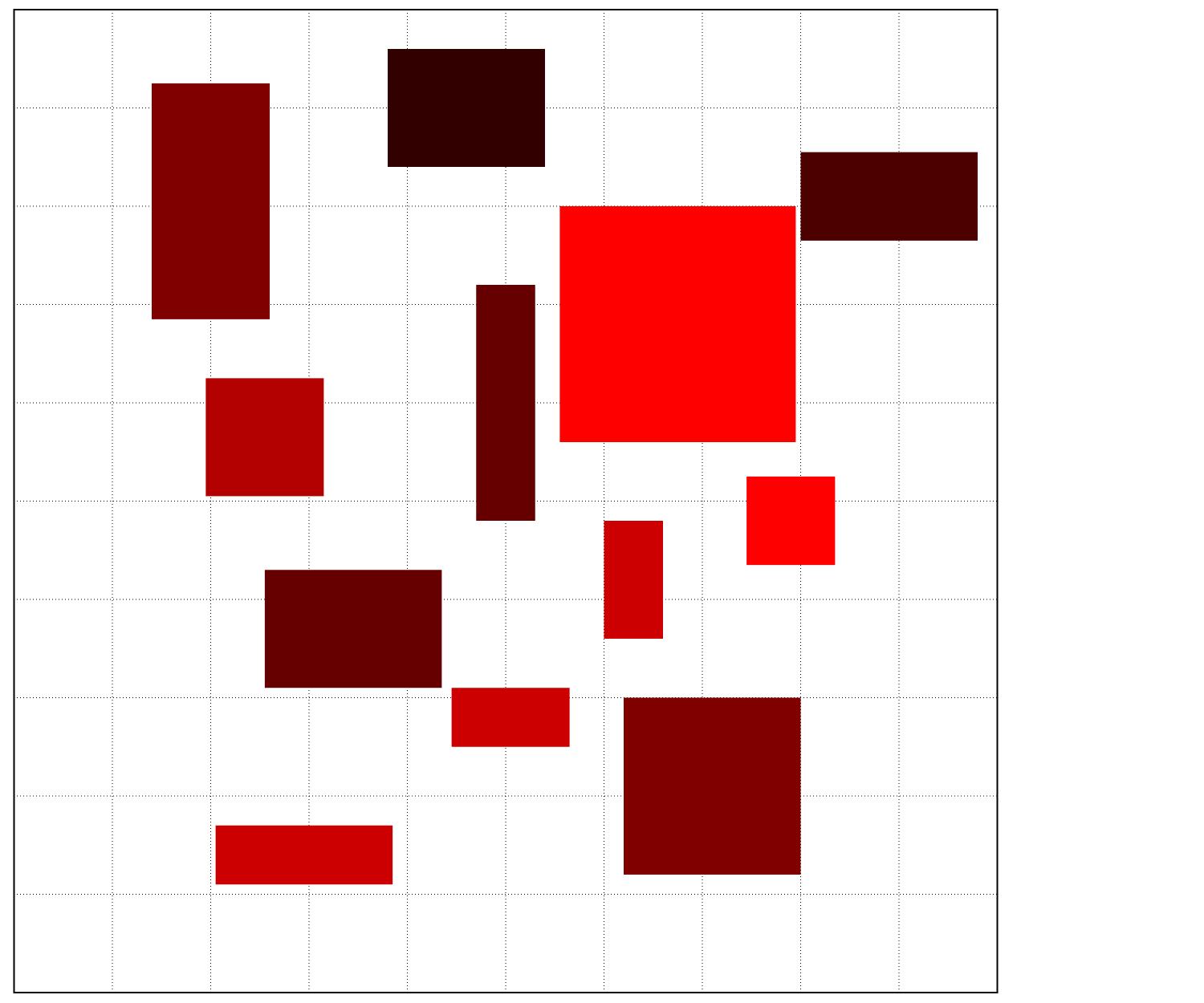}}
	\subfigure[Case 1]{\includegraphics[width=0.45\linewidth]{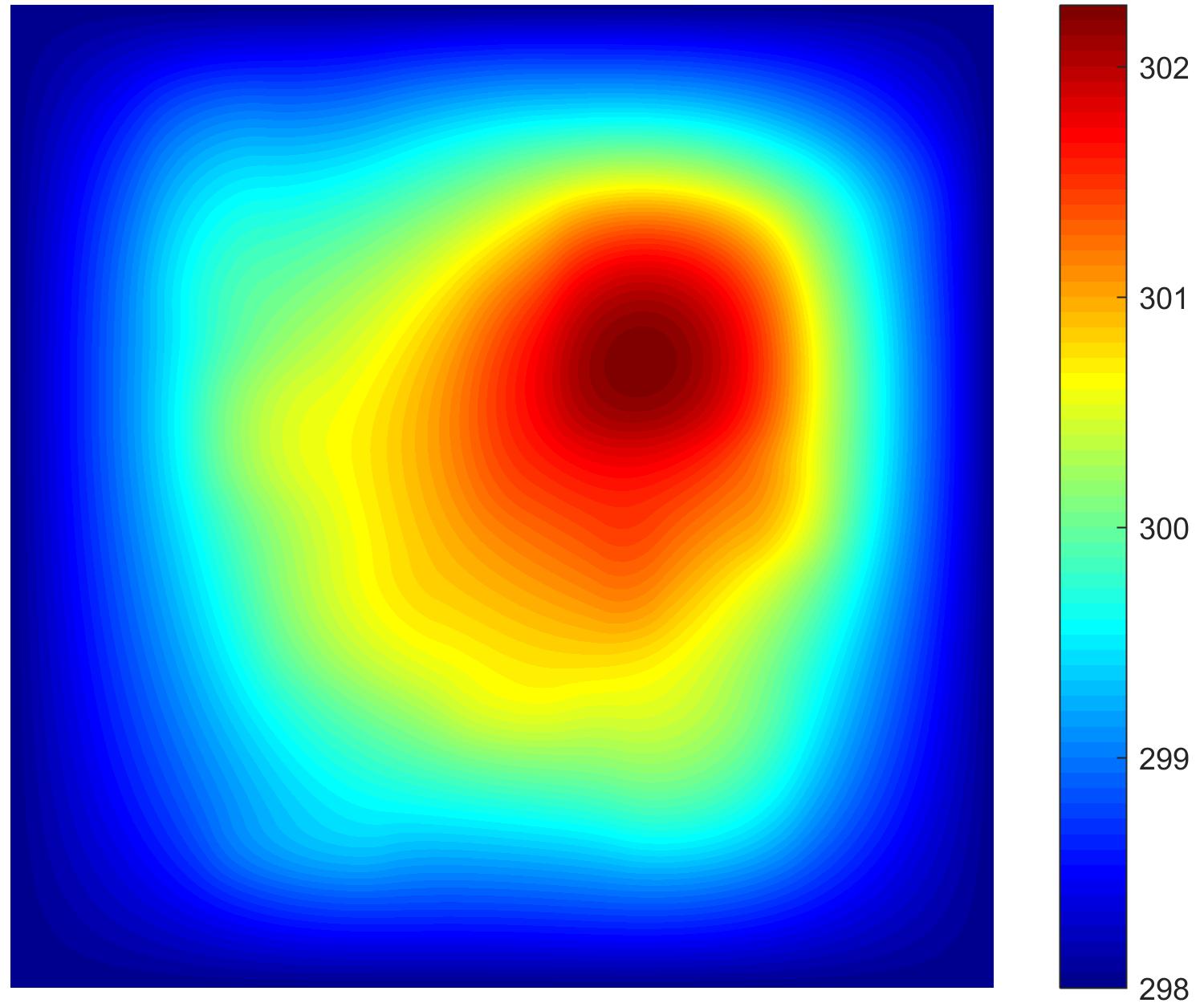}}
	\\
	\subfigure[Case 2]{\includegraphics[width=0.45\linewidth]{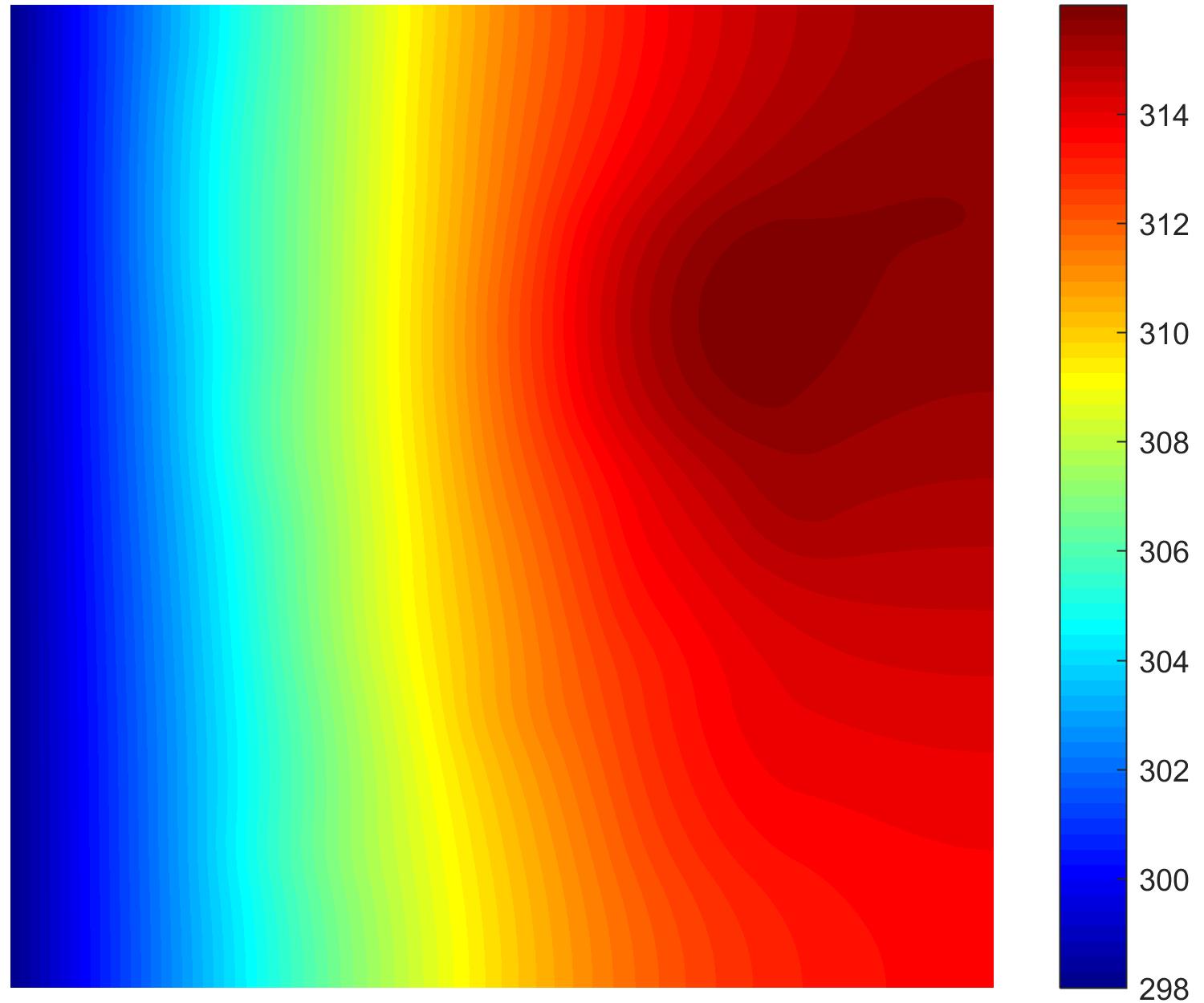}}
	\subfigure[Case 3]{\includegraphics[width=0.45\linewidth]{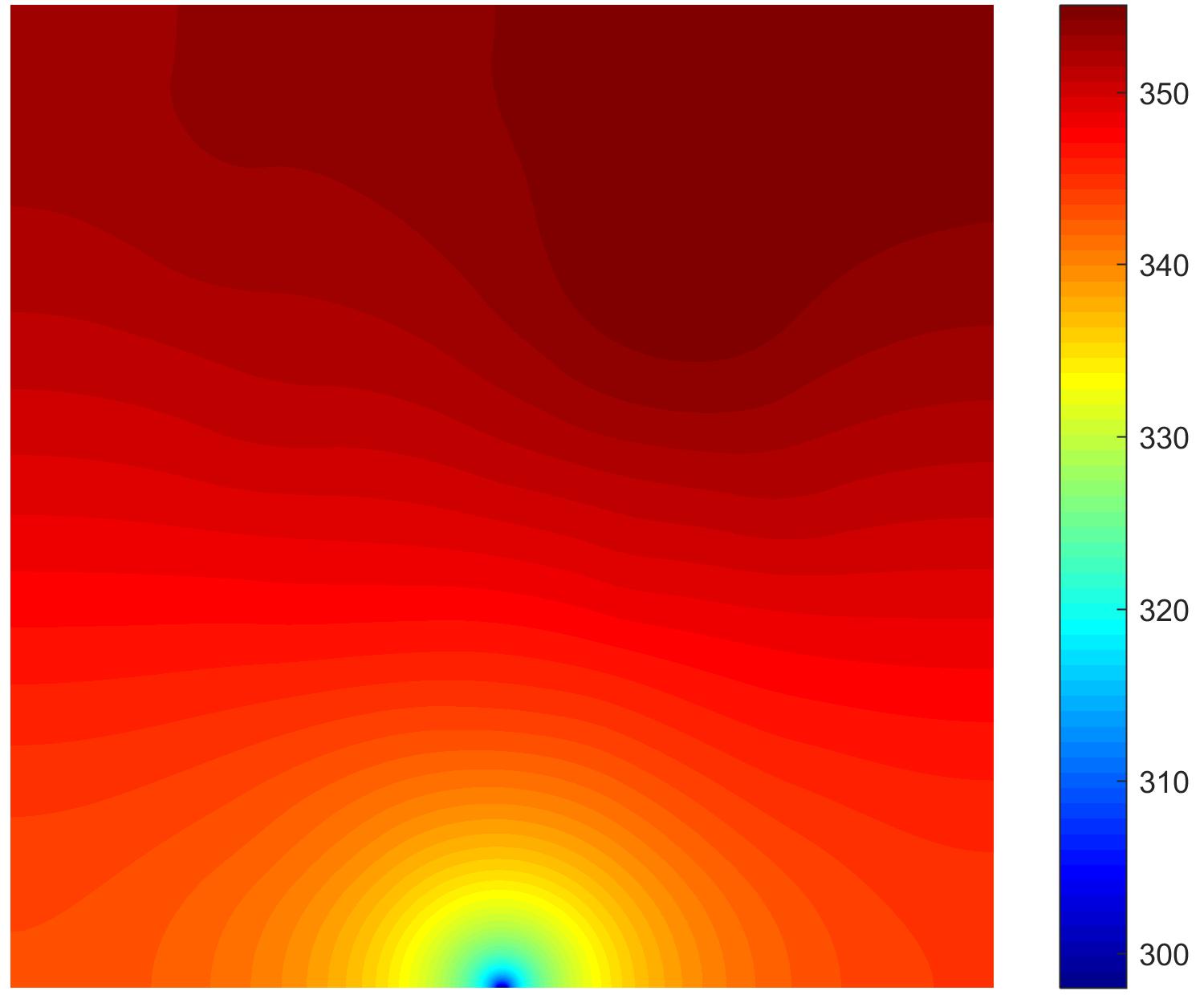}}
	\caption{The calculated temperature fields of a given layout example under three cases.}
	\label{fig:example}
\end{figure}

The steady-state temperature field corresponding to the layout scheme is calculated as the label using FEniCS, which is an open-source computing platform for solving partial differential equations based on the finite-element method and can be accessed at \url{ https://fenicsproject.org/}.
In the previous section, three layout problems which are defined under three different boundary conditions are taken as study cases in our HSLD.
To match the size of the input layout matrix, the domain also meshes as $200 \times 200$ rectangular cells for solving Poisson's equation numerically, thus generating a $200 \times 200$ temperature matrix.
Note that the uniform mesh is enough accurate for two VB problems while the adaptive refined grid around the narrow escape region is taken to guarantee the numerical simulation precision of the VP problem.
In other words, the temperature matrix for the VP problem is provided under the description of the nonuniform grid.
Taking one random layout as an example, the temperature fields under three cases are presented in Fig. \ref{fig:example}.

\subsection{Heat Source Layout Dataset (HSLD)}

To advance the state of the arts in HSL-TFP task, we construct the HSLD, a new diversity, large-scale heat source layout data set, using random layout sampling methods and special layout sampling strategies.

The new dataset is made up of three typical problems, namely VB with identical boundary conditions (Case 1), VB with different boundary conditions (Case 2), and the VP (Case 3) (See Section \ref{sec:definition} for details). For each problem, we construct the training samples and testing samples, respectively. All the training samples are obtained under the SeqLS method. Considering the tradeoff between the limited number of training samples and the performance of the deep surrogate models, we choose 2000 training samples for each problem. 
In order to ensure the completeness of samples for the test, we construct nine types of test datasets with two random sampling methods and other special sampling strategies.
The samples of test set 1-9 come from the SeqLS method,  the GibLS method, corner samples, group samples, part-space samples of $200\times 100$, part-space samples of $100\times 200$, part-space samples of $140 \times 140$, part-space samples of $120 \times 120$, and part-space samples of $100 \times 100$, respectively. 
For the SeqLS and GibLS types, we select 10000 testing samples, respectively. 
For corner samples and part-space samples in the square reduced space, we select 1000 samples for each situation, while there are $1000 \times 4$ group samples, $1000 \times 6$ part-space samples of $200 \times 100 $ and $1000 \times 6$ part-space samples of $100\times200$ in other test sets. 
Table \ref{table:number} shows the detailed compositions of the HSLD.

The developed dataset considers both the diversity of sampling strategies and the completeness of the generated samples. This brings more challenges for the HSL-TFP task and the test dataset can be more suitable for the development of deep surrogate models over the task.

\begin{table*}[htbp] 
	\centering
	\caption{The number of training and testing samples in HSLD. Training samples are obtained by the SeqLS method. Test set 1: the SeqLS samples; test set 2: the GibLS samples; test set 3: corner samples, test set 4: group samples; test set 5: part-space samples of 200$\times$100; test set 6: part-space samples of 100$\times$200; test set 7: part-space samples of 140$\times$140; test set 8: part-space samples of 120$\times$120; test set 9: part-space samples of 100$\times$100. }
	\begin{tabular}{ccccccccccc} 
		\hline
		\noalign{\smallskip}
		\multirow{2}[0]{*}{{\bf Problem}} & \multirow{2}[0]{*}{\bf TRAIN} & \multicolumn{9}{c}{{\bf TEST SET}} \\
		\cline{3-11}
		\noalign{\smallskip}
		& & \multicolumn{1}{c}{1} & \multicolumn{1}{c}{2} & \multicolumn{1}{c}{3} & \multicolumn{1}{c}{4} & \multicolumn{1}{c}{5} & \multicolumn{1}{c}{6} & \multicolumn{1}{c}{7} & \multicolumn{1}{c}{8} & \multicolumn{1}{c}{9} \\
		\hline
		\noalign{\smallskip}
	Case 1& 2000 & 10000  & 10000 & 1000 & $1000\times 4$  & $1000\times 6$  & $1000\times 6$  & 1000  & 1000 & 1000  \\
	Case 2& 2000 & 10000  & 10000 & 1000 & $1000\times 4$  & $1000\times 6$  & $1000\times 6$  & 1000  & 1000 & 1000  \\
	Case 3& 2000 & 10000  & 10000 & 1000 & $1000\times 4$  & $1000\times 6$  & $1000\times 6$  & 1000  & 1000 & 1000  \\
		\hline\noalign{\smallskip}
	\end{tabular}%
	\label{table:number}%
\end{table*}%

\subsection{A comprehensive analysis on HSLD}






In the following, we mainly analyze our HSLD from the problem definition and the obtained layout samples.

\subsubsection{Generality of Problem Definition}

In our previous work on deep learning surrogate-assisted layout optimization\citep{chen2020}, a simple heat source layout problem was investigated, where all components are identical and of the same shape (square), size and intensity.
By contrast, in this study, we propose to consider a more general and complicated heat source layout problem, which originates from engineering application, to construct the HSLD.
As listed in Table \ref{table:component}, the heat sources consist of rectangular and quadrate components.
In addition, these components maintain different sizes and intensities, which makes each of them unique.
Moreover, note that the placement resolution for components is one smallest cell, that is, 1/200 of the length of the domain, so that this layout problem can be approximated as the continuous one.
On the contrary, the previous research handled a discrete layout problem.
All these settings make the problem more general and practical, reducing the margin with the real-world layout design task.
Due to the complexity of the layout problem, it in turn brings more challenges to the HSL-TFP task based on deep learning models.

\subsubsection{Randomization and Uniformity of Layout Samples}


In our HSLD, two efficient random layout sampling methods guarantee the accessibility of a large-scale data set.
Then we use the SeqLS and GibLS method to randomly draw 10000 layout samples, respectively, aiming to discuss the randomization, uniformity, and space-filling property.

Intuitively, the position of each component is allowed to be randomly sampled in its current feasible layout region when performing the sequential placement, which basically guarantees the randomization of the SeqLS method.
For any feasible layout scheme, as long as it exists, there is always one possible process to take this out by using the SeqLS method.
As for the GibLS method, the randomization of sampling can also be achieved.
It is because given any two layout schemes, there is always one available state transition path to change from one layout to another due to a low spatial layout density ($<$25$\%$) in our problem.
In other words, there is no isolated layout scheme that cannot be generated by state transition in our problem.
In general, the randomization of random layout sampling methods guarantees a fair collection of samples in the entire layout space without any preference.

\begin{figure*}[!htbp]
	\centering	
	\subfigure[10000 Samples by SeqLS]{\label{fig:div_seq}\includegraphics[width=0.45\linewidth]{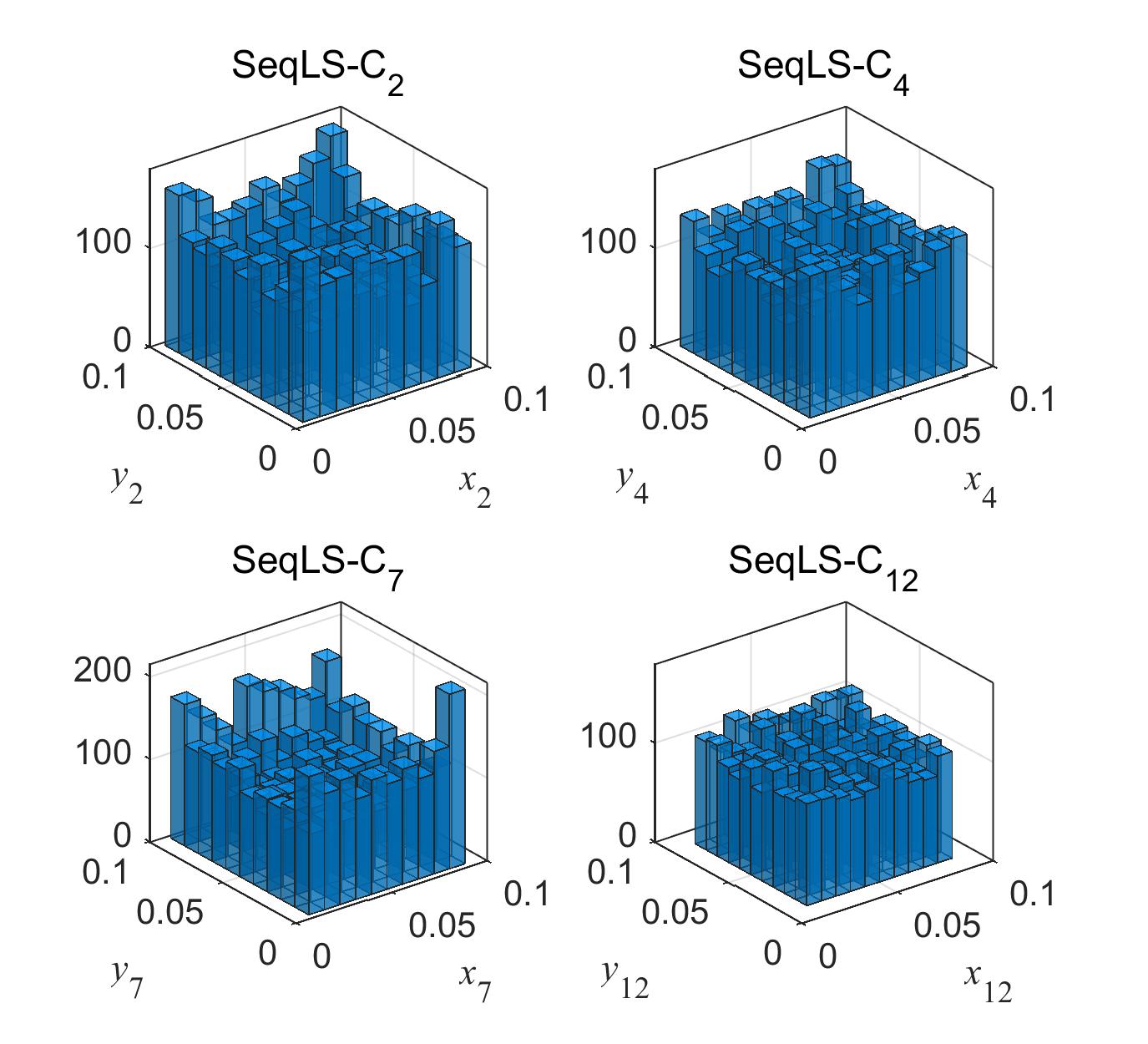}}
	\subfigure[10000 Samples by GibLS]{\label{fig:div_gib}\includegraphics[width=0.45\linewidth]{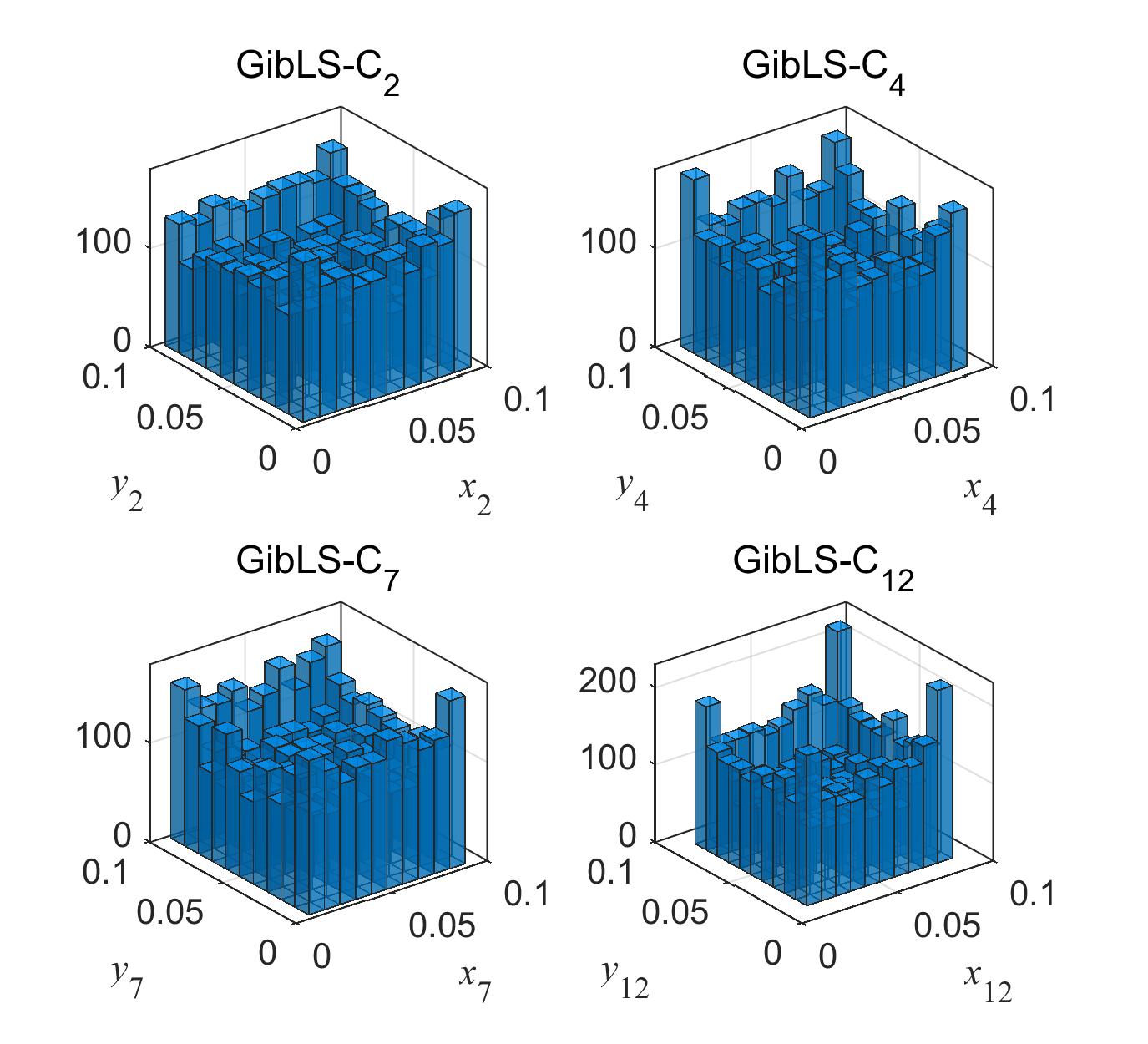}}
	\caption{The statistical distributions of the positions of central points of component 2, 4, 7 and 12 in 10000 samples obtained by the SeqLS and GibLS method. The layout domain for each component is divided into 10 $\times$ 10 bins to make these histograms.}
	\label{fig:diversity}
\end{figure*}

The uniformity of layout samples is investigated by seeing the position distribution of each single component.
By analyzing 20000 samples (10000 obtained by each method), it is found that the mean values of position coordinates of components are almost equal to 0.5, which is the middle of the sample interval.
This indicates that there is no obvious preference or imbalance on the component placement in the layout domain.
Besides, during the acquisition of 10000 samples, the SeqLS method maintains a 100$\%$ sampling success rate in our problem.
Thereby, the position of the firstly placed component, which is component 12 in our setting, should follow a uniform distribution in the two-dimensional plane.
This can be verified by seeing the position distribution of component 12 (SeqLS-$C_{12}$) in Fig. \ref{fig:div_seq}.
Apart from this, it is also observed from Fig. \ref{fig:diversity} that the position distributions for other components in the samples by the SeqLS and GibLS method maintain a higher frequency near the domain boundary.
It illustrates that in the feasible layout space, the probability of components being placed near the boundary is relatively higher than in the middle area.
We speculate one possible explanation that when components get away from each other, that is, move towards the domain boundary, they are more likely to satisfy the non-overlapping constraint.
In other words, under the effect of the non-overlapping constraint between components, the layout samples where components are scattered over the entire domain accounts for a large proportion in the entire feasible layout space.


\subsubsection{Analysis on the composition of HSLD}

As stated before, the generality of the investigated layout case makes the problem more practical, and in turn, increases the complexity of the layout problem.
Simultaneously, the layout space is also too complicated to be described owing to the non-overlapping constraint.
In this work, we mainly develop two efficient useful layout sampling methods to solve the constrained sampling difficulty, which allows us to acquire massive training or test data for further advancement of deep surrogate models.
However, the feasible layout space is so immense that it is hard to completely represent this design space using a finite number of samples.
Therefore, we leave the research on the training data with better space-filling property in future work.

Despite this, the evaluation of surrogates requires a comprehensive reflection on the entire layout space, while the completely random sampling cannot cover this space.
In order to alleviate this difficulty, we heuristically propose to construct some special samples from the aspect of neural network modeling and the layout optimization task.
The provided special layout samples are of great use to increase the diversity of our generated data set, which with random samples compose a relatively comprehensive test data set, as elucidated in Section \ref{sec: special samples}.

Until now, we have made an overall description of the benchmark problem and conducted a feasible construction for a reasonable HSLD.
The baseline deep surrogate models for learning the underlying laws from these data are applied and discussed in the following section.

\section{Deep Neural Network Surrogate Models}
\label{sec:baseline}


This section tries to develop a set of deep neural network (DNN) surrogate methods as baselines to advance the HSL-TFP task. As section \ref{sec:definition} shows, the HSL-TFP aims to predict the temperature value at each discrete position according to the heat source layout. 
One should observe that the underlying assumption of the HSL-TFP task is that the input and output can be treated as images and it can be considered as an image-to-image regression problem.
Due to the limited performance of general machine learning-based surrogate models and the excellent performance of deep models in other fields \citep{badrinarayanan2017segnet}, using the deep regression methods as deep surrogate models for the HSL-TFP task tends to be indispensable to boost the state-of-the-art performance for the HSL-TFP task.

In what follows, we introduce the process to construct the set of deep surrogate models for the HSL-TFP task in detail, including the data-preprocessing, the backbone networks, the deep regression frameworks, and the normalization methods used in the deep surrogate models.

\subsection{Data Pre-processing}
\label{subsec:preprocessing}

Data pre-processing, which can accelerate the convergence of the deep model and even improve the performance of the learned model, is an essential process for the training of deep models.
For the current task, the data pre-processing methods are also used for the training of deep surrogate models. 
For the HSL-TFP task, the input ${X}$ represents the intensity of the different components with a size of 200$\times$200 matrix, and output $Y$ describes the corresponding temperature field which has the same size. 
To make the input and output of the network internally consistent and benefit the gradient descent process during training, this work uses data standardization and feature standardization technology.
The input ${X}$ of the network is pre-processed as
\begin{equation}
	X_0 = \frac{X - X_m}{X_{std}}
\end{equation}
where $X_m$ denotes the mean of the input and $X_{std}$ represents the standard deviation of the input data.
Similar to the input $X$, the output $Y$ is pre-processed as
\begin{equation}
	Y_0 = \frac{Y - Y_m}{Y_{std}}
\end{equation}
where $Y_m$ denotes the mean of the output and $Y_{std}$ represents the standard deviation of the output data.
It should be noted that the standardization of the output is implemented through the sigmoid layer in the deep model and can provide end-to-end training.

In the training of deep surrogate models, $X_m$ and $X_{std}$ are set to 0 and 1000 to keep the input in a smaller range. To re-scale the output with the distribution value between 0 and 1, $Y_m$ and $Y_{std}$ are set to 298 and 50 for Case 1 and Case 2, while 298 and 100 for Case 3.
With the data pre-processing process, the deep surrogate models can be trained more steadily.

\subsection{Backbone Networks}

The backbone networks which extract the discriminative features from the input images play a quite important and essential role in deep regression methods. 
Due to excellent performance, some backbone networks become the standard component in various deep network frameworks, and they are expediently replaced with each other on account of the structural similarity.
Generally, deep surrogate models with different backbones would present differences in performance. 
Therefore, in order to construct the set of deep surrogate models for the -TFP task, this work mainly selects the most used deep networks, namely the AlexNet, VGG, and ResNet as the backbone network in the constructed models.
The following will introduce these backbone networks respectively.

\subsubsection{AlexNet}
\begin{figure}
	\centering
	\includegraphics[width=1.0\linewidth]{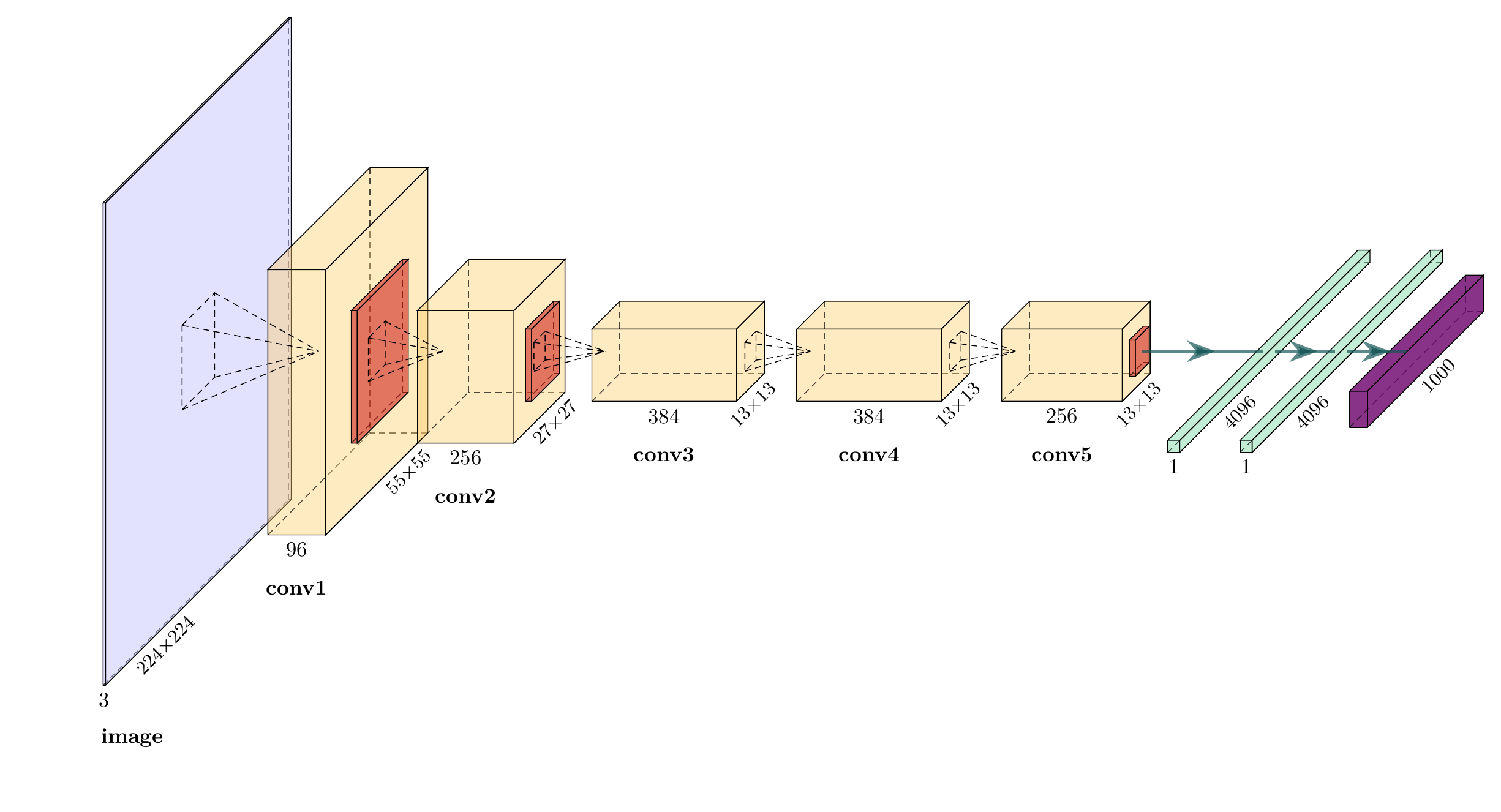}
	\caption{The architecture of AlexNet.}
	\label{fig:alexnet}
\end{figure}

AlexNet \citep{2012ImageNet} was first proposed by Krizhevsky in ILSVRC-2012 competition and obtained excellent performance when compared with other traditional machine learning methods.
It contains eight learned layers, including five convolution layers, three fully connected layers. The architectures of Alexnet are shown in Fig.~\ref{fig:alexnet}.
As the figure shows, it adopts 11$\times$11 kernel in the first convolution layer to capture large range of objects, and the kernel sizes are reduced to 5$\times$5 and 3$\times$3 in the rest convolution layers. The training process of the deep model effectively alleviates the overfitting problem with the dropout technology.
In addition, the sigmoid activation function is replaced by the Rectified Linear Units (ReLUs) to avoid the vanishing gradient.
The network has achieved great success in many real-world applications \citep{ballester2016} and this work will use this specific network as one of the backbone networks in the deep surrogate models for the HSL-TFP task.

\subsubsection{VGG}

\begin{figure}
	\centering
	\includegraphics[width=1.0\linewidth]{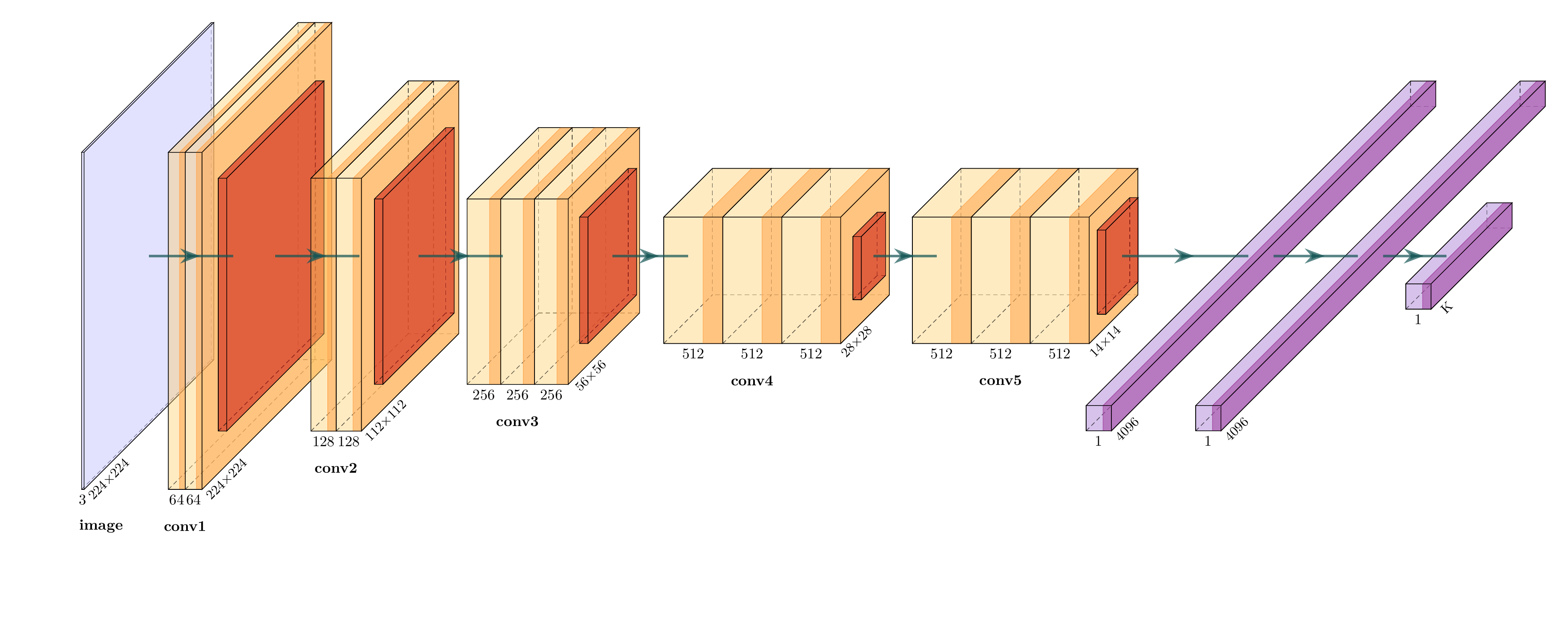}
	\caption{The architecture of VGG-16.}
	\label{fig:vgg16}
\end{figure}

Instead of large kernel convolution in AlexNet, VGG \citep{2014Very} uses the stack of small kernel convolutions to form the large kernel.
Fig.~\ref{fig:vgg16} shows the architecture of VGG-16, which is the most used form of VGG.
In each block, multiple convolution layers with 3$\times$3 kernel and one-pixel padding are stacked.
It can be find that the same receptive field is obtained between two consecutive 3$\times$3 convolution layers and one 5$\times$5 convolution layers, similarly for three consecutive 3$\times$3 convolution layers and one 7$\times$7 convolution layers.
However, this special structure in VGG can reduce the network parameters and increase the depth of the block without sacrificing the size of the feature map.
In addition, max-pooling is followed by each stacked block with a stride of 2 to reduce the obtained feature map.
Considering the merits of the VGG network, this work selects the VGG-16 as another backbone network for the HSL-TFP task.

\subsubsection{ResNet}
ResNet \citep{he2016deep} is the milestone in the development of CNN.
It is well-known that with the increase of the depth, deep network always suffers from serious performance degradation. 
Generally, after the network is overlaid with more layers, the performance drops rapidly.
However, benefit from deep residual learning mechanism, ResNet can effectively avoid such degradation with a deeper network structure.
The ResNet also takes advantage of the stack of small kernel convolutions just as the VGG network.
While different from the VGG blocks, the residual blocks in ResNet do not directly learn a desired underlying mapping $\mathcal{H}(x)$, but introduce identity skip mapping to copy the current inputs to the next block, and fit another mapping of $\mathcal{F}(x):=\mathcal{H}(x)-x$. Then, the original mapping is recast into $\mathcal{F}(x) + x$.
It is easier to optimize the residual mapping than to optimize the original, unreferenced mapping \citep{he2016deep}.
\begin{figure}
	\centering
	\subfigure[Basic residual block]{
		\includegraphics[width=1.0\linewidth]{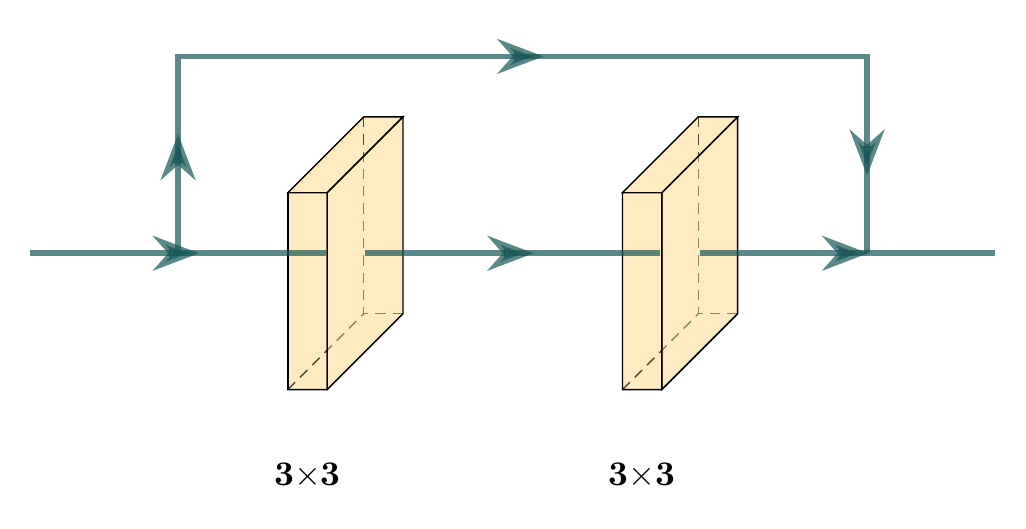}
	}
	\quad
	\subfigure[Bottleneck residual block]{
		\includegraphics[width=1.0\linewidth]{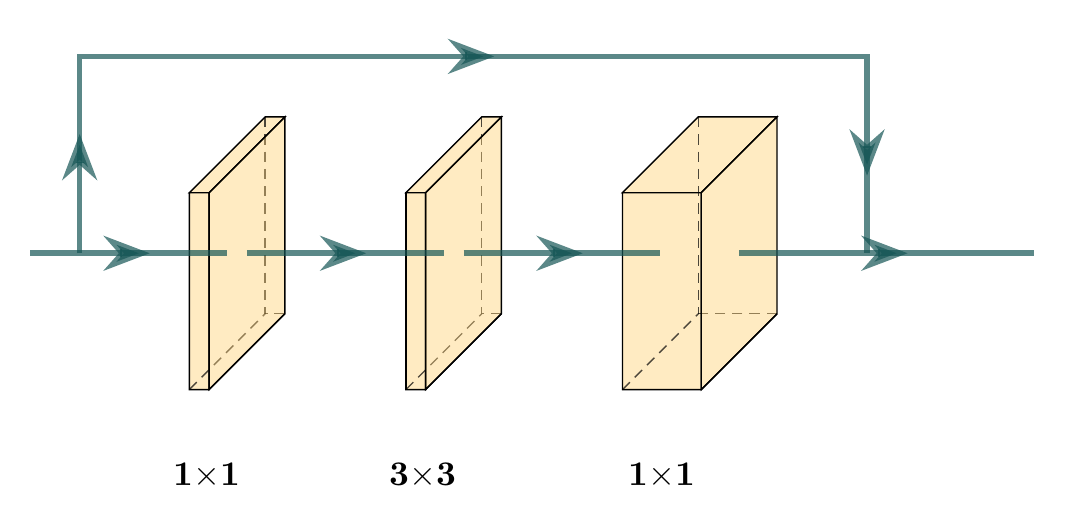}
	}
	\caption{Two kinds of ResNet building block: Basic and Bottleneck residual block.}
	\label{fig:resnet_block}
\end{figure}
Fig. \ref{fig:resnet_block} shows different building blocks of Resnet.
Benefit by the Basic and Bottleneck residual architectures which can overcome the vanishing gradients problem, the depth of ResNet can be up to 152 layers. 
Due to the significant performance of ResNet, this work chooses ResNet as the backbone to learn representation from different layouts.

In addition to these former networks, there exist some other common architectures, such as GoogleNet \citep{szegedy2015}, DenseNet \citep{huang2017} and ReNet \citep{visin2015}. 
On the basis of the mission analysis, this work selects these three potential networks (e.g., AlexNet, VGG, and ResNet) as representative backbone networks to construct DNN surrogate models for the HSL-TFP task.
Researchers are also encouraged to develop more effective backbone networks for this image-to-image surrogate modeling.

\subsection{Deep Regression Frameworks for HSL-TFP task}

As the former shows, the HSL-TFP task can be modeled as an image-to-image regression task and therefore can be divided into the encoder process and the decoder process. Using different encoder and decoder methods, deep regression frameworks can be divided into several classes, such as the fully convolutional networks (FCNs), the SegNet, the Unet, the feature pyramid networks (FPNs), and others. This work attempts to construct the set of deep surrogate models based on these deep regression frameworks and the following will introduce these frameworks for the HSL-TFP task in detail.

\subsubsection{Fully Convolutional Network (FCN) for HSL-TFP task}
Fully convolutional networks (FCNs) \citep{2015Fully}, which only perform convolution operations (downsampling or upsampling),  are a general framework to solve the image-to-image regression task.
Generally speaking, a fully convolutional network is achieved by replacing the parameter-rich fully connected layers in standard CNN architectures with convolutional layers with $1 \times 1$ kernels.
The key to generate dense prediction outputs in FCNs is to use deconvolution layers, which are just convolutional layers with input and output swapped.
In addition, FCNs contatenate different layers of the network, combining deep, coarse layer's semantic information with shallow, fine layer's appearance information.
By employing multi-scale features, FCNs could achieve a larger receptive field and make a more accurate prediction.
Taking advantage of these architectures, 
FCNs can efficiently learn to make dense predictions of input.

\begin{figure*}
	\centering
	\includegraphics[width=1.0\linewidth]{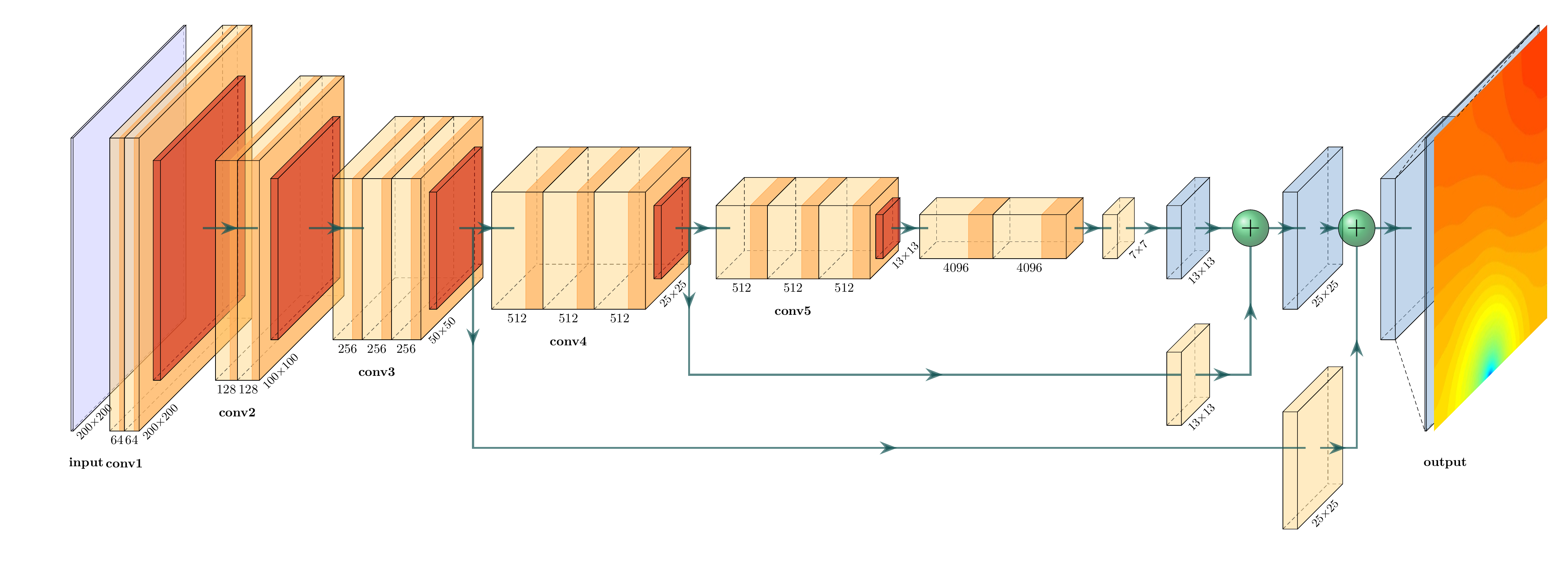}
	\caption{The architecture of FCN with VGG16 backbone.}
	\label{fig:fcn8}
\end{figure*}

In this paper, we propose to take advantage of the FCNs for end-to-end prediction of the HSL-TFP task. 
Based on the former strategies of FCNs, we construct FCNs with different backbone networks, namely the AlexNet, VGG16, and ResNet.
Fig.~\ref{fig:fcn8} shows the architecture of FCN with VGG16 backbone.
As former introduces, there are five pooling layers in VGG16, orderly denoted as pool1, pool2, pool3, pool4, and pool5. The predictions from pool3, pool4, and pool5 are respectively at stride 8, 16, and 32. FCN-8s combine the predictions from pool5 and pool4 to get feature maps at stride 16, and then combine the outputs with the predictions from pool3 to get feature maps at stride 8. 
During combination, 1$\times$1 filters is adopted to ensure the same channels for different feature maps, and block with 3$\times$3 filters and 2 times linear interpolation to make consistency of size.
Finally, 8 times linear interpolation is employed to provide 200$\times$200 temperature field predictions.
FCNs based on AlexNet and ResNet follow similar construction methods.
Notably, for FCN with AlexNet backbone, the kernel size of the first layers in AlexNet is set to 7 instead of 11, and the padding of max-pooling layers is set to 1. 

\subsubsection{SegNet for HSL-TFP task}

\begin{figure*}
	\centering
	\includegraphics[width=1\linewidth]{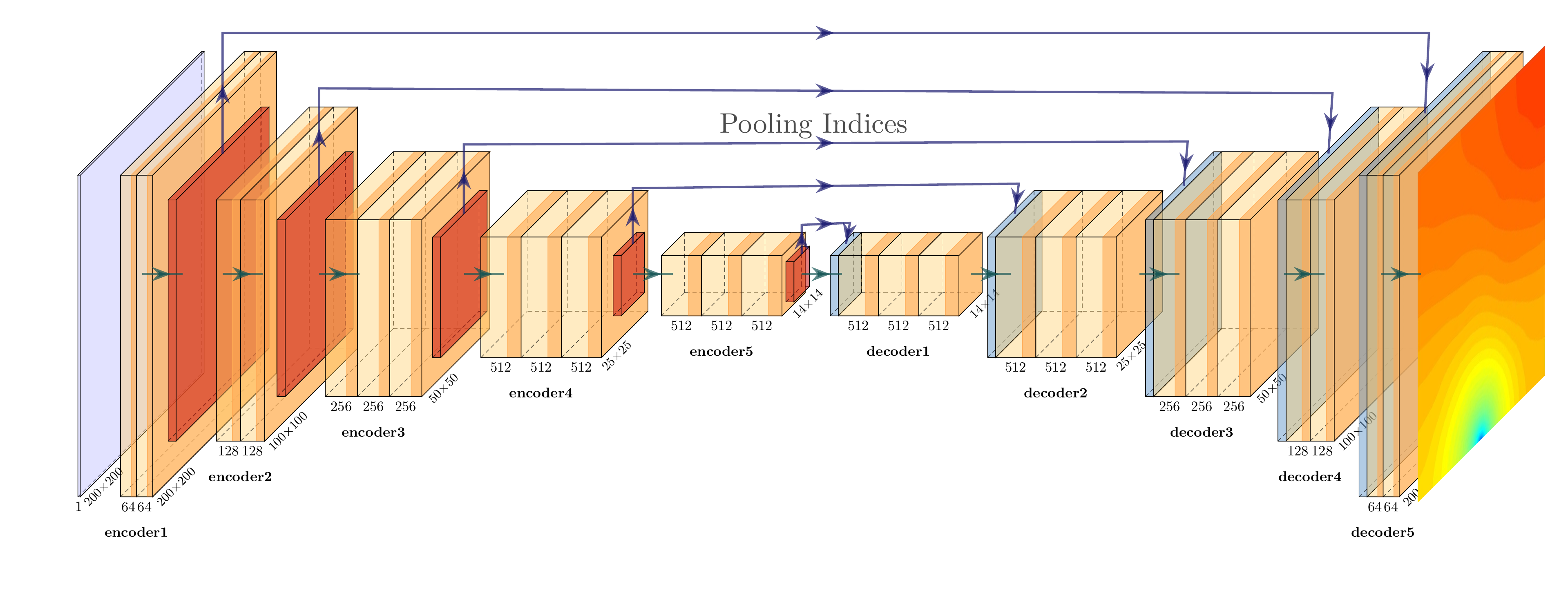}
	\caption{The architecture of Segnet. In the stage of decoder, pooling indices transfered from encoder are used to upsample the input to produce sparse feature maps.}
	\label{fig:segnet}
\end{figure*}

SegNet \citep{badrinarayanan2017segnet} is a typical convolutional encoder-decoder architecture for pixel-wise classification. 
Fig.~\ref{fig:segnet} shows the architecture of Segnet based on the VGG network, including the encoder stage and the decoder stage.
The encoder stage is the representative convolutional networks without their fully connected layers in vanilla CNNs. Because of several pooling layers in the encoder, the input image is down-sampled to low-resolution feature maps. 
In contrast to the encoder stage, the decoder stage aims to upsample those low-resolution images to high-resolution predictions. 
In Segnet, the decoder is composed of a set of upsampling and convolution layers, and the architecture of decoder blocks are corresponding to the encoder. 
In contrast to FCN, the skip architecture to utilize information of shallow layers is abandoned in SegNet. Furthermore, SegNet utilize a new upsample technique, which uses the max-pooling indices to upsample the feature maps.

For the current task, we take advantage of the AlexNet, VGG net, and ResNet as the encoder of Segnet, and the convolution blocks in the decoder are corresponding to the encoder stage.
Specially, for SegNet with AlexNet backbone, we have removed the first max-pooling layers in the encoder and adopt double deconvolutions with kernel size 2 and stride 2 to realize 4 times upsampling in the final stage of the decoder.
Detailed architectures of the SegNet can be seen in our released code on the website.

\subsubsection{Unet for HSL-TFP task}

Unet \citep{ronneberger2015u} is initially presented for biomedical image segmentation and is another convolutional encoder-decoder deep regression framework. 
Similar to SegNet, the Unet architecture consists of an encoder to capture context and a symmetric decoder to make high-resolution pixel-wise predictions, which are respectively called contracting path and expanding path. 
In addition, just as the FCN, skip connection architecture is also used in Unet to combine the features from the contracting path with the upsampled feature maps from the corresponding expanding path. 

\begin{figure*}
	\centering
	\includegraphics[width=1\linewidth]{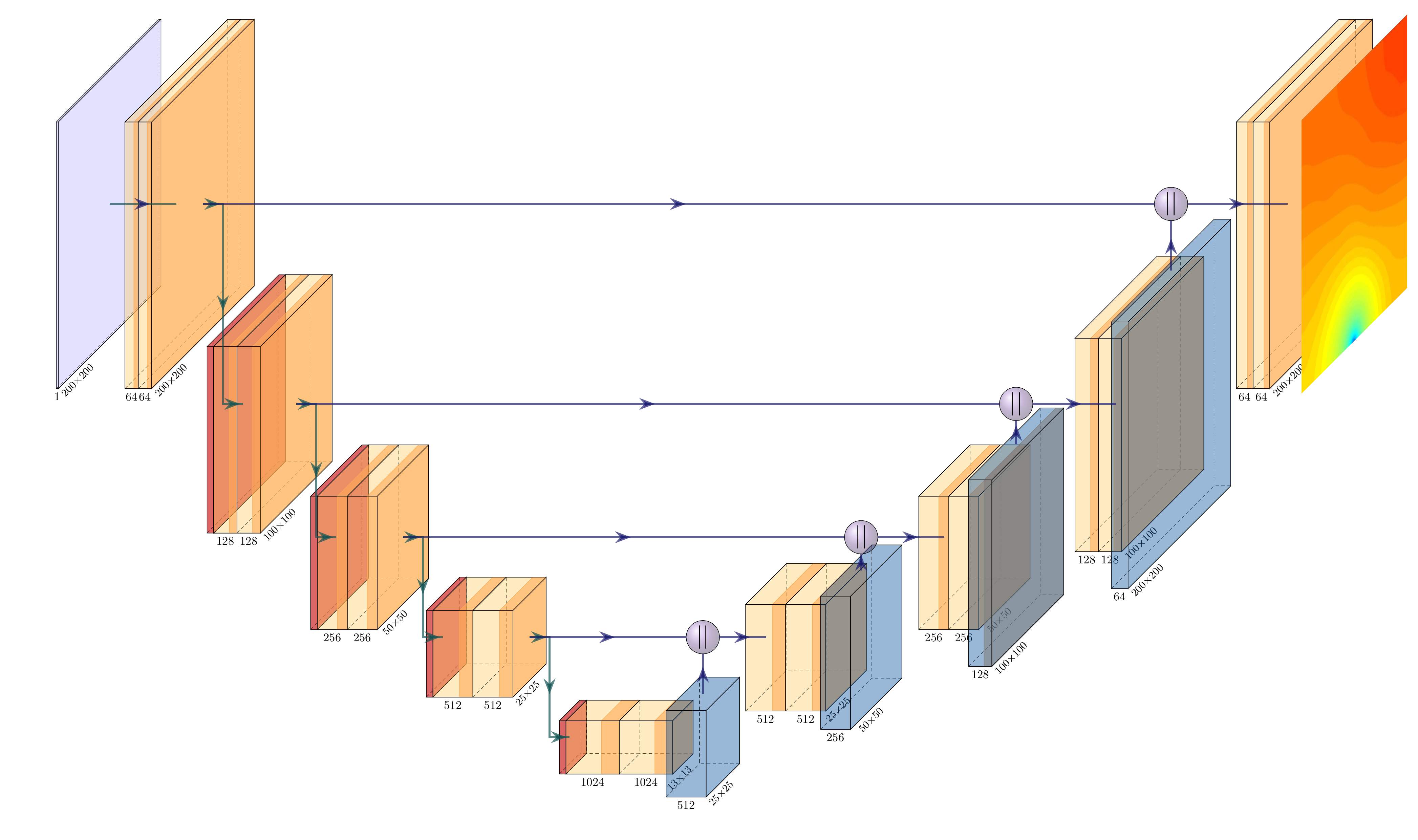}
	\caption{The architecture of Unet.}
	\label{fig:unet}
\end{figure*}

Fig. \ref{fig:unet} shows the typical architecture of Unet.
As shown in Fig.~\ref{fig:unet}, there are four max-pooling layers to reduce feature map size, and skip connections are conducted before each pooling layer to concatenate feature maps in the channel dimension.
The Unet supplements a usual contracting network by successive layers, where pooling operations are replaced by upsampling operators to increase the resolution of the output. The successive convolutional layer can then learn to assemble a precise dense prediction based on this information.


To fit the HSL-TFP task, the 3$\times$3 convolutions with one padding is adopted in the Unet. In addition, normalization layers are performed after the 3$\times$3 convolution layer.
 
\subsubsection{Feature Pyramid Networks (FPN)  for HSL-TFP task}

FPN \citep{lin2017feature} is a special type of fully convolutional networks to take advantage of multi-scale information through feature pyramid structures. 
To build high-level semantic feature maps at different scales, the FPN utilizes the top-down path as well as the lateral connections. 
By these simple connections, FPN can build pyramid representations without substantially intensive computation and memory.
   
\begin{figure*}
	\centering
	\includegraphics[width=1\linewidth]{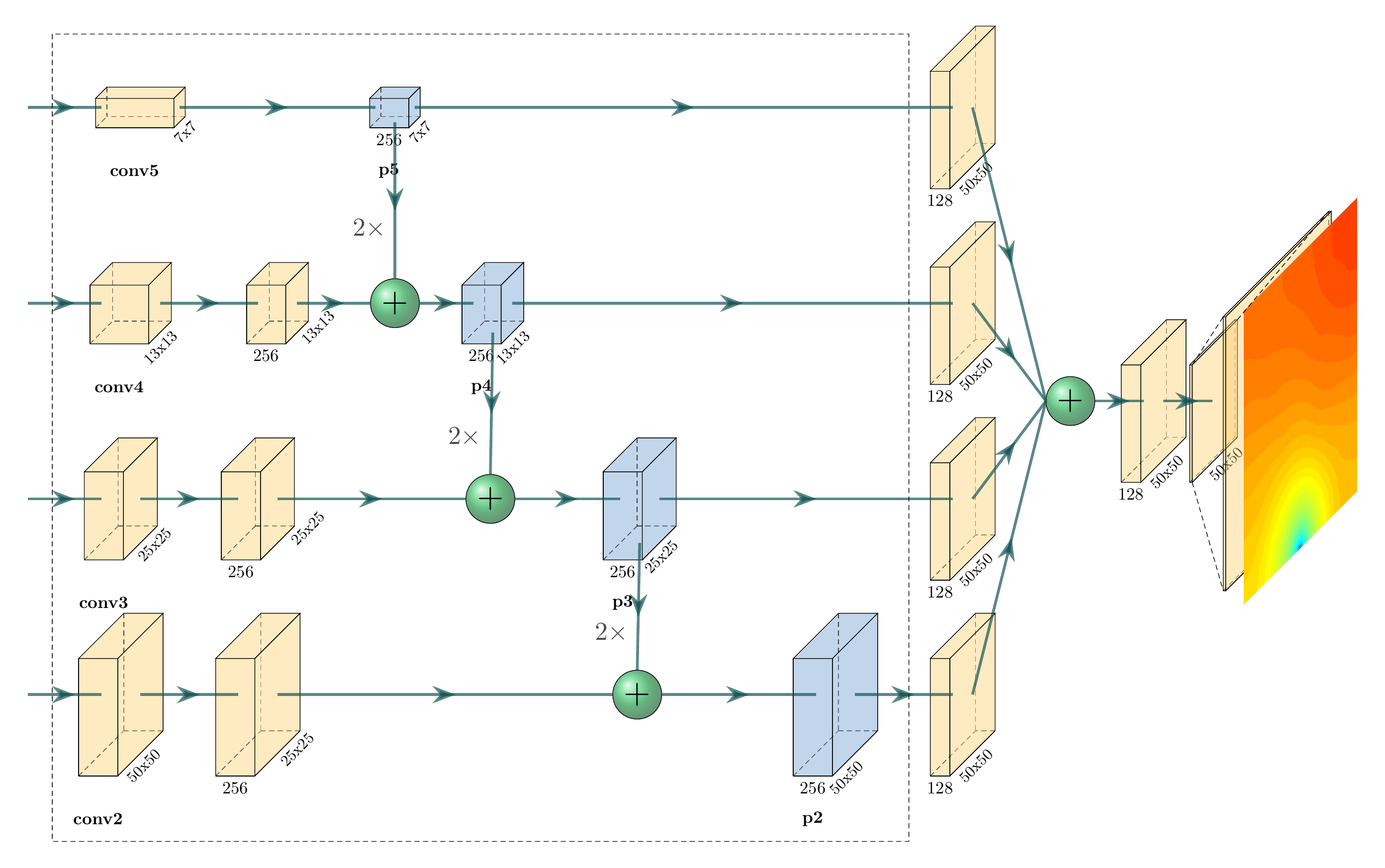}
	\caption{The architecture of FPN. The architecture consists of the FPN backbone to produce multi-scale features and the head to combine features to make predictions.}
	\label{fig:fpn}
\end{figure*}

The architecture of FPN is shown in Fig.~\ref{fig:fpn}.
FPN is composed of three parts: bottom-up pathway, top-down pathway, and lateral connections. 
Note that since the bottom-up pathway is the vanilla CNNs, such as the ResNet and VGG, we do not present in the figure.
As Fig. \ref{fig:fpn} shows, different resolution feature maps are obtained by stack of convolution and pooling layers. 
The output maps of the same size are regarded as the same network stage. 
Taking ResNet as an example, the output of the last residual block in each stage is used to construct a feature pyramid in FPN. 
There exist four feature maps, respectively with strides 4, 8, 16, and 32, which can construct the four-level feature pyramid. 
Then the top-down pathway aims to transform the low-resolution feature maps to higher resolution by upsampling. 
With the lateral connections, the feature maps from bottom-up and top-down pathways of the same spatial size are merged to combine both the high-level and low-level semantic information. 
Therefore, the FPN can take full use of multi-scale information and provide accurate dense prediction.

For this specific task, we develop the architecture to utilize the multi-scale features based on FPN. 
First, the multi-scale feature maps are expanded to the same size by successive two times upsamping and 3$\times$3 convolutions, and then these features would be fused by add operation. 
The final predictions are produced by 1$\times$1 convolutions and upsampling. 
Besides, we take advantage of the ResNet-18, ResNet-50, and ResNet-101 as the backbone network in the bottom-up pathway. More details of the implementation can be seen in the released code.

This work mainly takes advantage of these special deep regression frameworks as representative baselines to construct the set of the deep surrogate models in the HSL-TFP task. Interested researchers can explore other deep surrogate models to advance the state-of-the-art methods for the HSL-TFP task.

\subsection{Normalization}

Normalization of input data is the common data processing method to accelerate the training of deep networks. However, with the increasing depth of networks, the unstable numerical calculation makes it notoriously hard to train deep models.
Such model normalization, e.g. batch normalization and group normalization, is a milestone technique of deep learning, which can accelerate deep network training by reducing internal covariate shift.
In the construction of the set of deep surrogate models for the HSL-TFP task, batch normalization (BN) and group normalization (GN) are also applied and played an important role in the training of the deep surrogate models. Following we will introduce the two normalizations in detail.

\subsubsection{Batch Normalization}

Batch Normalization (BN) \citep{ioffe2015batch} aims to normalize layers' inputs using the mini-batch statistics. Given a layer with $d$-dimensional input $x=(x^{(1)}...x^{(d)})$ and a mini-batch $\mathcal{B}=\left\{ {{x_{1},\cdots, x_{m}}} \right\}$ of size $m$, the statistics for each dimension $x_i^{(j)}$ are denoted as
\begin{equation}
\mu _\mathcal{B}^{(j)} = \frac{1}{m}\sum\nolimits_{i = 1}^m {x_i^{\left(j\right)}} 
\end{equation}
\begin{equation}
\left({\sigma {_\mathcal{B}^{(j)}}}\right)^2 = \frac{1}{m}{\sum\nolimits_{i = 1}^m {\left( {x_i^{\left(j\right)} - {\mu _\mathcal{B}}} \right)} ^2}
\end{equation}
Then, the normalized value can be calculated by
\begin{equation}
\hat x_i^{\left( j \right)} = \frac{{x_i^{\left( j \right)} - \mu _\mathcal{B}^{\left( j \right)}}}{{\sqrt {{{\left( {\sigma _\mathcal{B}^{\left( j \right)}} \right)}^2} + \varepsilon } }}
\end{equation}
In addition, pair of parameters, namely $\gamma _i^{\left( j \right)}$ and $\beta _i^{\left( j \right)}$, are learned by network training to scale and shift the normalized value.
\begin{equation}
y_i^{\left( j \right)} = \gamma _i^{\left( j \right)}\hat x_i^{(j)} + \beta _i^{\left( j \right)}
\end{equation}
For CNN, BN is usually employed between the convolutional layer and activation function. If there exist multi-channel outputs, each channel applys BN.

\subsubsection{Group Normalization}

Group Normalization (GN) \citep{wu2018group} is developed to overcome the inaccurate batch statistics estimation problem by BN, especially for training processes with small batch sizes. In contrast with BN, GN divides the channels into groups and the mean and variance are computed in each group for normalization. Since this computation is independent of batch sizes, the normalization is more stable with the change of batch sizes. 

For the HSL-TFP task, the batch normalization and group normalization will be used in the deep surrogate models to accelerate the training process. In the experiments, we will test the performance of the proposed deep surrogate models with different normalizations.

\section{Evaluation Metrics}
\label{sec:metric}

To compare the performance of different deep learning surrogates quantitatively, we propose to evaluate models from the following three main categories of metrics: pixel-level metrics, image-level metrics, and batch-level metrics.

\subsection{Pixel-level Metrics}


In the layout optimization task driven by the thermal performance, the maximum temperature of the layout domain is usually regarded as one important variable that designers concern about.
Hence, two specific metrics are defined here to evaluate the performance on this aspect, which is summarized as pixel-level metrics.
This kind of metrics illustrates the local performance of different models.

\medskip
\noindent
\textbf{The Absolute Error of the Maximum Temperature (MT-AE).} 
MT-AE is defined as one of pixel-level metrics, which can be calculated as
\begin{equation}
	\text{MT-AE} = |\max(\hat{Y}) - \max(Y)|
\end{equation}
where $\hat{Y}$ represents the predicted temperature matrix while $Y$ is the corresponding groud-truth temperature matrix.

\smallskip
\noindent
\textbf{The Position Absolute Error of the Maximum Temperature (MT-PAE).}
MT-PAE is defined as the other pixel-level metric, representing the deviation between the point where the maximum temperature happens in the predicted temperature field and that in the label temperature field.
The specific points in $\hat{Y}$ and $Y$ are denoted by $\hat{P}_{max}(\hat{x}, \hat{y})$ and $P_{max}(x, y)$, respectively, which are described using grid coordinates in the $200 \times 200$ grid system.
MT-PAE is calculated as the Euclidean distance of $\hat{P}_{max}$ and $P_{max}$, which can be written as
\begin{equation}
	\text{MT-PAE} = \sqrt{(\hat{x} - x)^2 + (\hat{y} - y)^2}
\end{equation}

\subsection{Image-level Metrics}

In contrast with pixel-level metrics, image-level metrics are developed to put more emphasis on evaluating the global performance on the temperature prediction task of different models.

\medskip
\noindent
\textbf{MAE.} 
MAE is the mean absolute error of the prediction on the whole temperature field, which can be represented as
\begin{equation}
	\text{MAE} = \frac{1}{N^2} \sum_{i=1}^{N} \sum_{j=1}^{N} |\hat{Y}_{ij} - Y_{ij}|
\end{equation}
where $N$ represents the mesh size along one direction and is a constant in our setting, that is, $N=200$.

\medskip
\noindent
\textbf{Max AE.}
Max AE is the maximum absolute error of the prediction on the temperature field, which can be written as 
\begin{equation}
	\text{Max AE} = \max_{i,j} |\hat{Y}_{ij} - Y_{ij}|
\end{equation}
.

\smallskip
\noindent
\textbf{Boundary-constrained MAE (BMAE).}
BMAE is defined to describe the prediction accuracy on the boundary of the layout domain.
In our problem, two types of boundaries including Dirichlet BC and Neumann BC are involved in Case 2 and 3 while only Dirichlet BC is involved in Case 1.
Note that the boundary segments with different BCs are evaluated separately.
When calculating this metric, the boundary segments are firstly extracted based on problem settings to divide these nodes on the boundary into different BCs.
In case 3, the nonuniform mesh should be taken into consideration when identifying different boundary nodes.
For convenience, we define a mask with the same size as the temperature matrix denoted by $M$ to calculate the specific metric, where only boundary nodes belong to this type of BCs maintain the value of 1 and the other is 0.
Therefore, BMAE for two BCs (BMAE$_D$ and BMAE$_N$ for Dirichlet and Neumann BC, respectively) can be computed as
\begin{equation}
	\text{BMAE}_D = \frac{1}{sum(M_D)} \sum \left( M_D \otimes |\hat{Y} - Y| \right)
\end{equation}
\begin{equation}
	\text{BMAE}_N = \frac{1}{sum(M_N)} \sum \left( M_N \otimes |\hat{Y} - Y| \right)
\end{equation}
where $M_D$ and $M_N$ represent the mask of Dirichlet and Neumann BCs, respectively; $\otimes$ stands for the element-wise multiplication and the same below; $sum(M_D)$ is the sum of the value of all elements of $M_D$ and the same as $sum(M_N)$.

\smallskip
\noindent
\textbf{Component-constrained MAE (CMAE).}
CMAE is defined to describe the temperature prediction precision of the area that components cover. 
Following the same computational procedures as BMAE, the nodes occupied by the components should be identified first and then expressed using masks.
In case 3, this identification should be carefully handled.
The mask for different components is denoted by $M_i(i=1,2,...,12)$.
Therefore, CMAE can be calculated as
\begin{equation}
	\text{CMAE}_i = \frac{1}{sum(M_i)} \sum \left( M_i \otimes |\hat{Y} - Y| \right)
\end{equation}
where CMAE$_i(i=1,2,...,12)$ represents the CMAE for component $i$. 

To concisely represent this indicator, CMAE metric is taken as the maximum one among CMAE$_i$, which is $ \text{CMAE} = \max_{i} (\text{CMAE}_i) $.

\medskip
\noindent
\textbf{Gradient MAE (G-MAE).}
The metric about the physical gradient (unit: K/m) is proposed here to assess whether the surrogate learns the feature on distribution variations embedded in the provided training dataset.
The gradient field of the temperature distribution is computed using central differences along two directions (the horizontal axis and the longitudinal axis), of which the formula is provided as (2.2) in \citep{Reimer2013}.
This formula has second-order accuracy in representing the gradient for two VB problems while first-order accuracy for the VP problem, since the uniform mesh is used in VB problems while the non-uniform mesh in VP problem.
To avoid the inaccurate representation in the boundary, only the gradient of the internal temperature field is compared.
Thus, the returned gradient matrix along the horizontal axis (denoted by $G_x$) maintains the size of $(N-1) \times (N-1)$, as well as the gradient matrix along the longitudinal axis (denoted by $G_y$).
Then we calculate the summation of the MAE of the gradient along two axes as our indicator denoted by G-MAE.
Hence, we evaluate the prediction accuracy on the gradient by using the metric G-MAE, which can be expressed as
\begin{equation}
	\text{G-MAE} = \frac{1}{(N-1)^2} \sum \left (|\hat{G}_{x} - G_{x}| + |\hat{G}_{y} - G_{y}| \right)
\end{equation}
where $(G_x, G_y)$ are calculated based on the provided ground-truth temperature field and $(\hat{G}_x, \hat{G}_y)$ based on the predicted one.

\medskip
\noindent
\textbf{Laplace MAE (Lap-MAE).}
The Laplace MAE (Lap-MAE) metric is proposed to reveal the capability of the surrogate learning the inherent laws behind the provided pairs of datasets.
As stated before, the steady-state temperature field follows Poisson's equation, which involves a Laplace operator $\Delta$ on the temperature field in our case study.
Therefore, we define the Lap-MAE as the MAE between the Laplace operator on the predicted temperature field $\Delta \hat{Y}$ and that on the provided labeled temperature field $\Delta Y$.
In this metric, the second-order derivative is approximately calculated following the central difference formula (2.3) defined in \citep{Reimer2013}, which has a second-order accuracy for a uniform mesh and first-order otherwise.
Similarly, the second-order derivative value of boundary points in the temperature matrix is neglected and thus the returned $\Delta \hat{Y}$ and $\Delta Y$ have the size of $(N-1) \times (N-1)$.
Besides, it is worth noting that it is unreasonable to directly testify the consistency of $\Delta \hat{Y}$ with $\phi(x,y)$ in Poisson's equation, as the labeled $Y$ is not necessarily obtained using the finite-difference method based on the above central difference pattern.

The Lap-MAE can be written as
\begin{equation}
	\text{Lap-MAE} = \frac{1}{(N-1)^2} \sum |\Delta \hat{Y} - \Delta Y|
\end{equation}
.


\subsection{Batch-level Metrics}

Pixel-level metrics and image-level metrics aim to provide a comprehensive review based on the absolute data of the predicted temperature field while the developed batch-level metric here aims to present the ranking capability of different surrogate models by performing correlation analysis based on the ranks of the data.
In order to quantitatively analyze the ranking ability, \textbf{Spearman rank-order correlation coefficient} $\rho_s$ \citep{kokoska2000crc}, a non-parametric statistical measure, is applied.
The range of the Spearman coefficient is $[-1, 1]$ with 0 implying no correlation. 
The closer to 1 $\rho_s$ is, the higher the positive monotonic correlation between two datasets is.
The closer to -1, the higher the negative monotonic correlation is.
Therefore, as we expect for surrogates, the closer to 1, the better.
The ranking data is chosen to be the maximum temperature in the whole field, which is usually the minimizing objective for layout optimization.
It means that the rank-order correlation analysis between the predicted and the true maximum temperature for each surrogate is performed, thus denoting the metric by $\rho_{MT}$.
The batch size of samples for evaluating correlation is chosen as 100 and all test samples are reviewed by repeating this process multiple times.
Hence, we can assess each surrogate by comparing its mean and standard deviation of Spearman coefficient $\rho_s$, thus distinguishing the ranking ability of different models.

\section{Experimental Studies}
\label{sec:experiments}

In this section, we mainly investigate the prediction performance of different deep regression frameworks-based surrogates, including FCN, SegNet, FPN, and Unet. 
Firstly, two feature normalization methods, involving GN and BN, are discussed by studying the training efficacy. 
Then, with the choice of the proper normalization method, different DNN surrogates are created by integrating different backbone networks, including AlexNet, VGG, and ResNet, in the deep regression frameworks. 
By calculating pixel-level metrics, image-level metrics, and batch-level metrics, an overall evaluation on the prediction performance of DNN surrogates is performed in nine test sets.
Apart from comparing the performance of different surrogates, their characteristics are exhaustively discussed and some useful insights about the surrogate modeling are further provided.
Moreover, the model efficiency is also investigated by presenting the number of model parameters and computing their forward inference latency on different hardware.


\subsection{Experimental Setups}
In our experiments, some adjustments over different deep surrogate frameworks have been made to adapt to the HSL-TFP task.
DNN surrogates of FCN based on backbone of AlexNet, VGG-16, and ResNet-18, SegNet based on AlexNet, VGG-16, and ResNet-18, Unet, FPN based on ResNet-18, ResNet-50, and ResNet-101, have been described as FCN-AlexNet, FCN-VGG, FCN-ResNet, SegNet-AlexNet, SegNet-VGG, SegNet-ResNet, Unet, FPN-ResNet18, FPN-ResNet50, and FPN-ResNet101, respectively.
In addition, the softmax activation function of each model in the last layer has been replaced by sigmoid to implement the regression task and simultaneously realize the normalization of the output. 

DNN surrogate models are implemented using an open-source deep learning framework Pytorch. 
Each model is trained for 400 epochs with a batch of preprocessed data.
The batch size is set as 32. 
We use the Adam solver with an initial learning rate of $1.0\times10^{-3}$. 
The polynomial decay policy with 0.99 multiplicative factor is conducted during the training process and the momentum is set to 0.9.
It should be noted that all the results are presented by calculating the average value of five independent repeated runs of the training and testing process.

\subsection{Results and Discussion}

This section is organized into three parts: the comparison of two feature normalization methods, the comparison and discussion of ten representative surrogates in nine test sets using three types of evaluation metrics, and the investigation on the model efficiency.

\subsubsection{Comparison of two feature normalization methods}

\begin{figure*}[tp]
	\centering
	\subfigure[]{
		\includegraphics[width=0.235\linewidth]{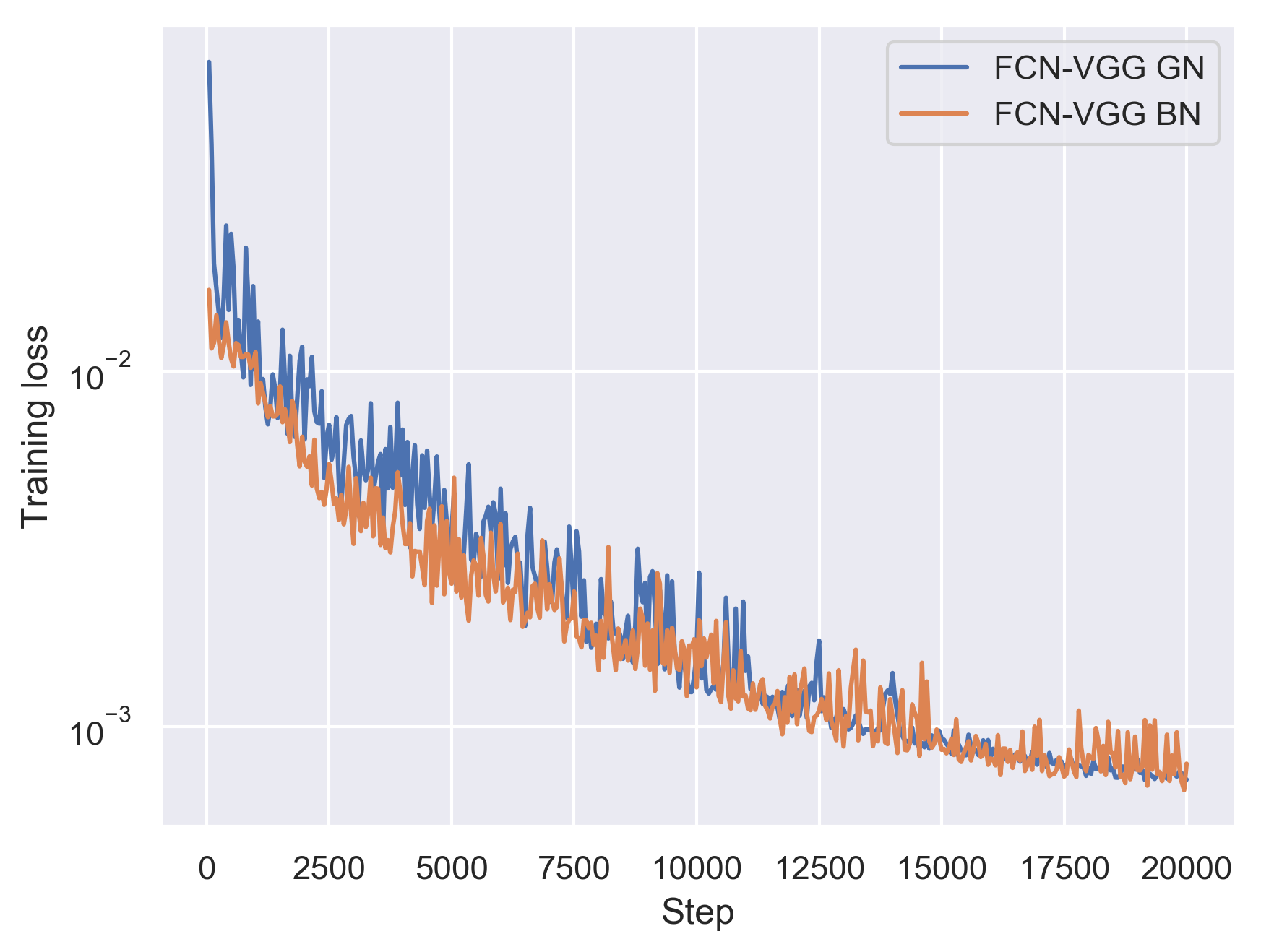}}
	\subfigure[]{
		\includegraphics[width=0.235\linewidth]{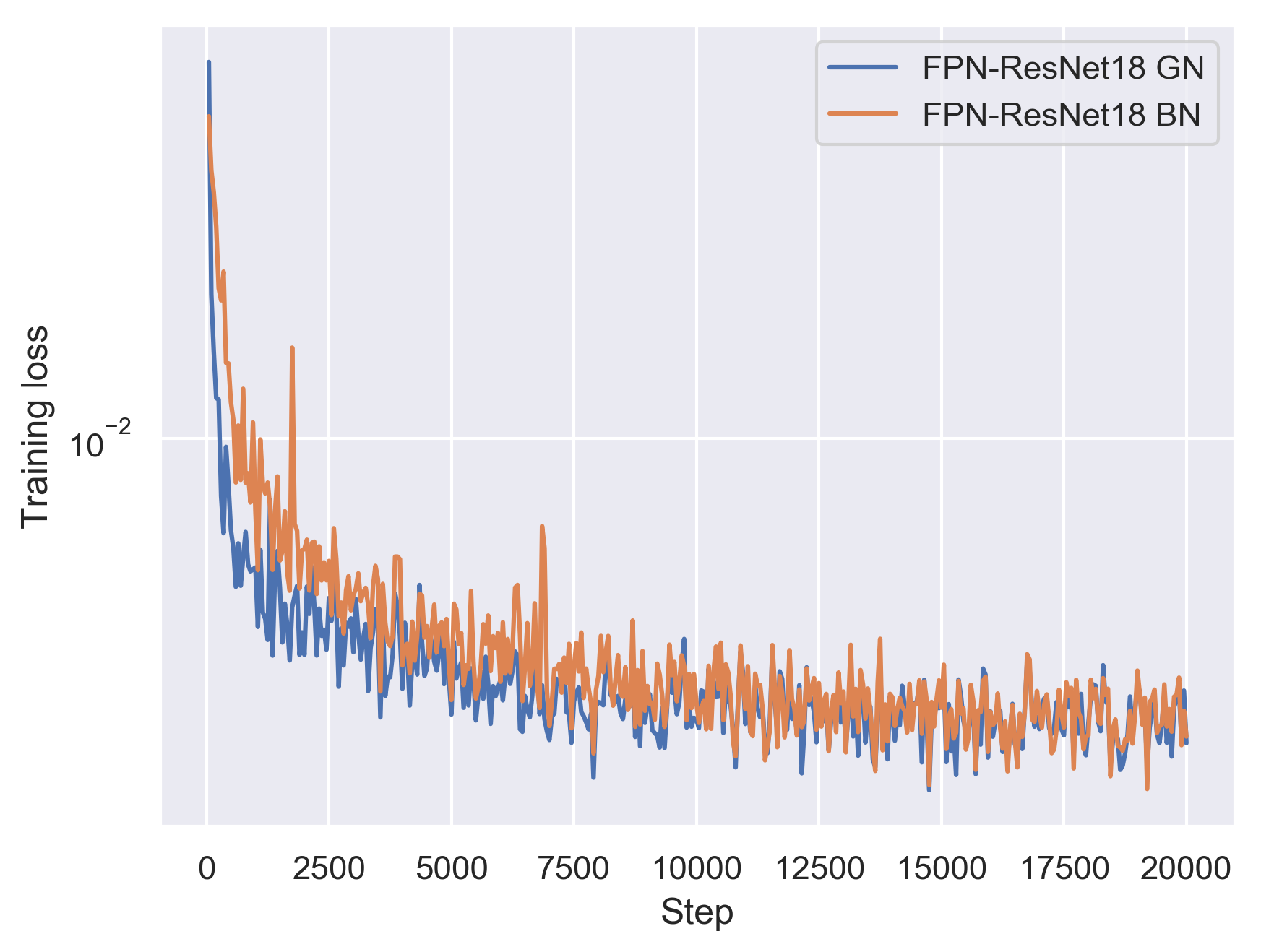}}
	\subfigure[]{
		\includegraphics[width=0.235\linewidth]{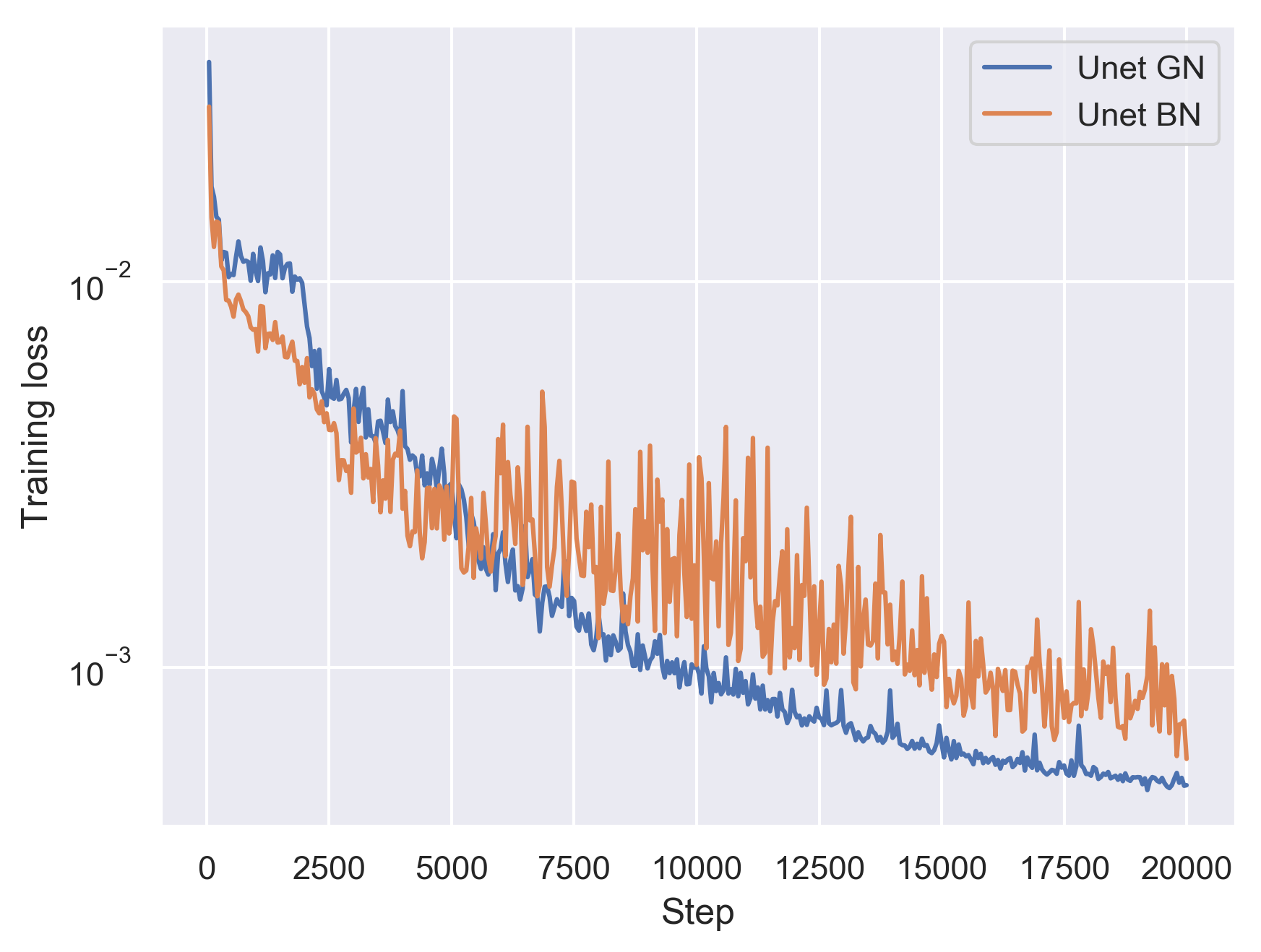}}
	\subfigure[]{
		\includegraphics[width=0.235\linewidth]{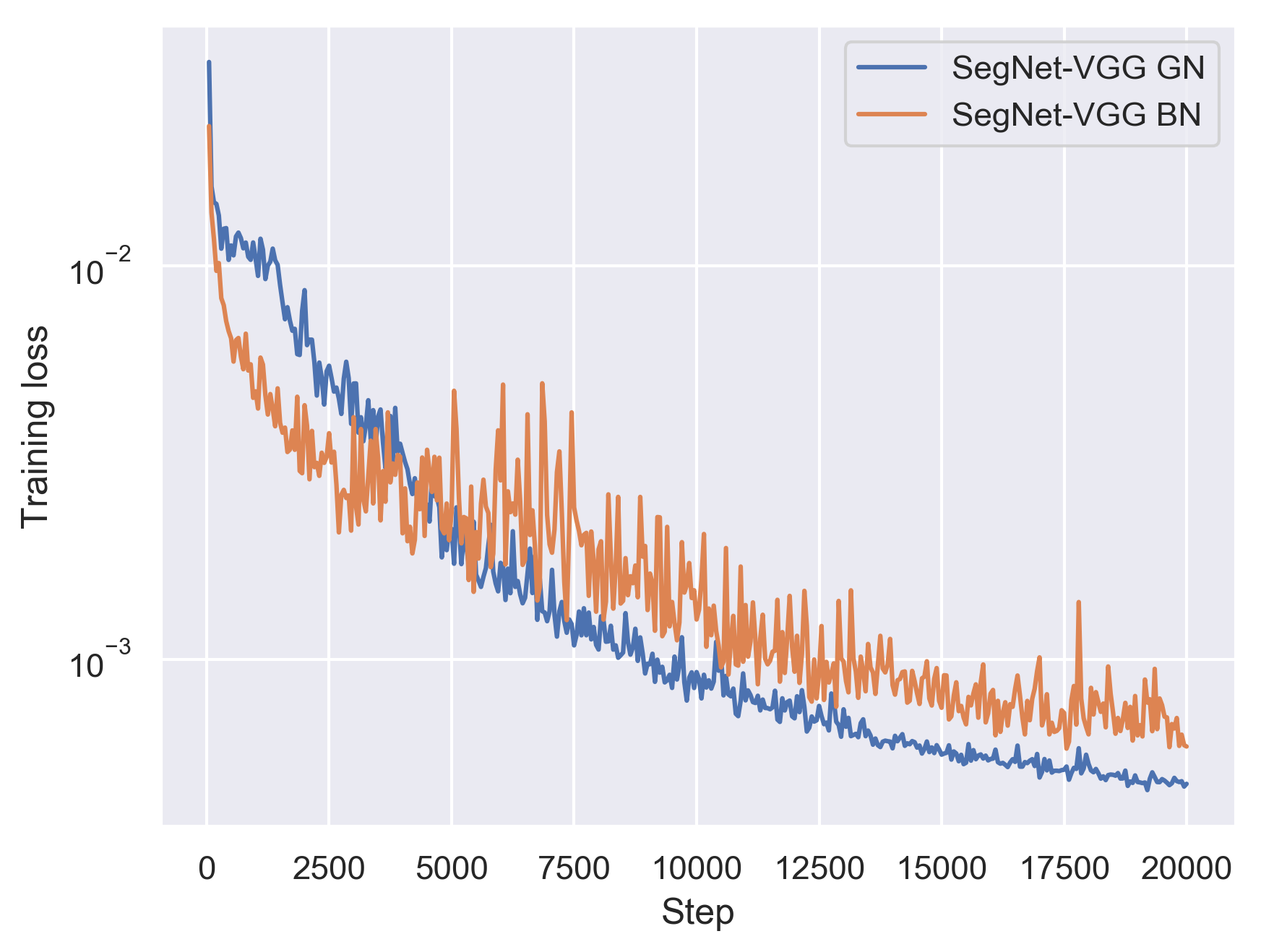}}
	\\
	\subfigure[]{
		\includegraphics[width=0.235\linewidth]{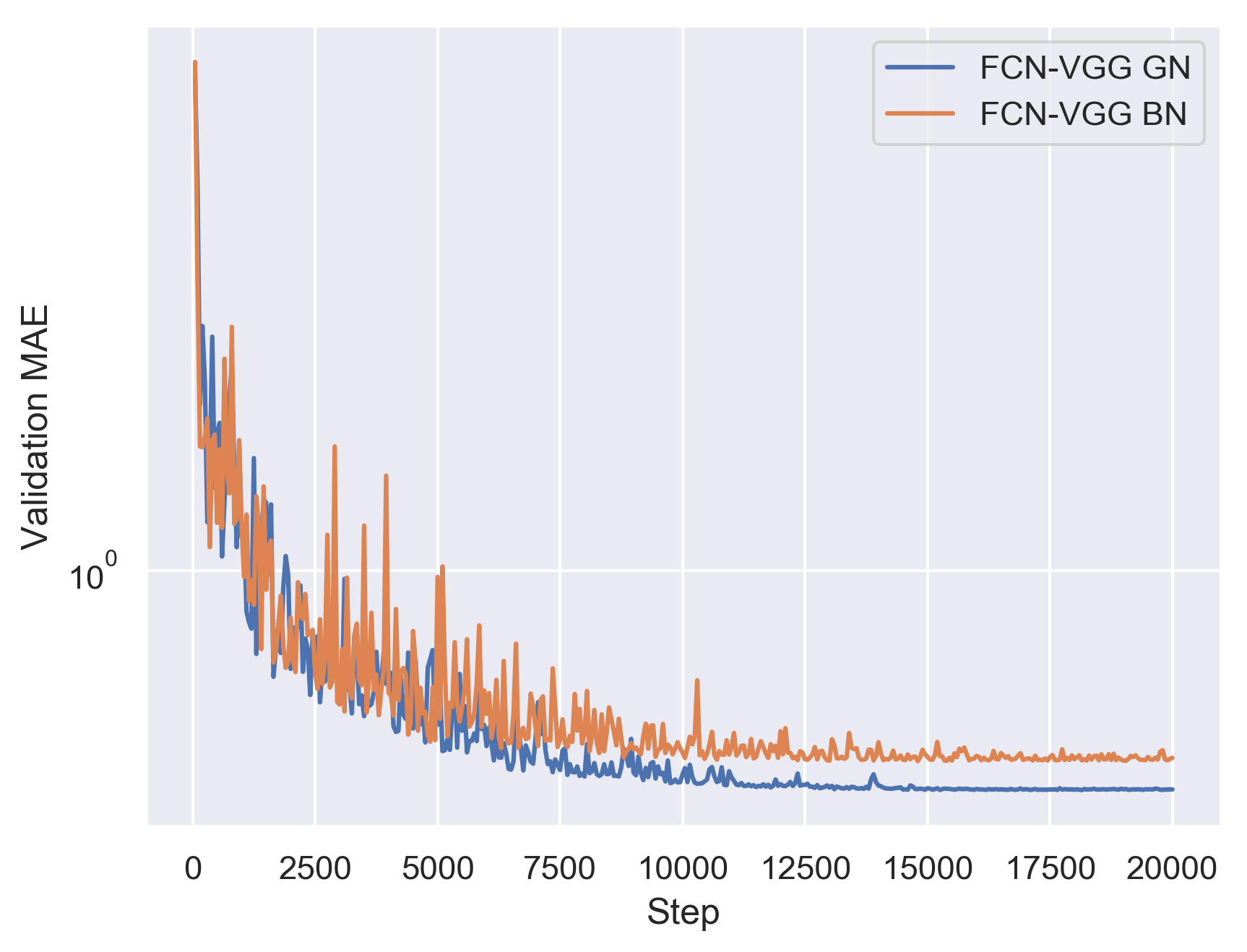}}
	\subfigure[]{
		\includegraphics[width=0.235\linewidth]{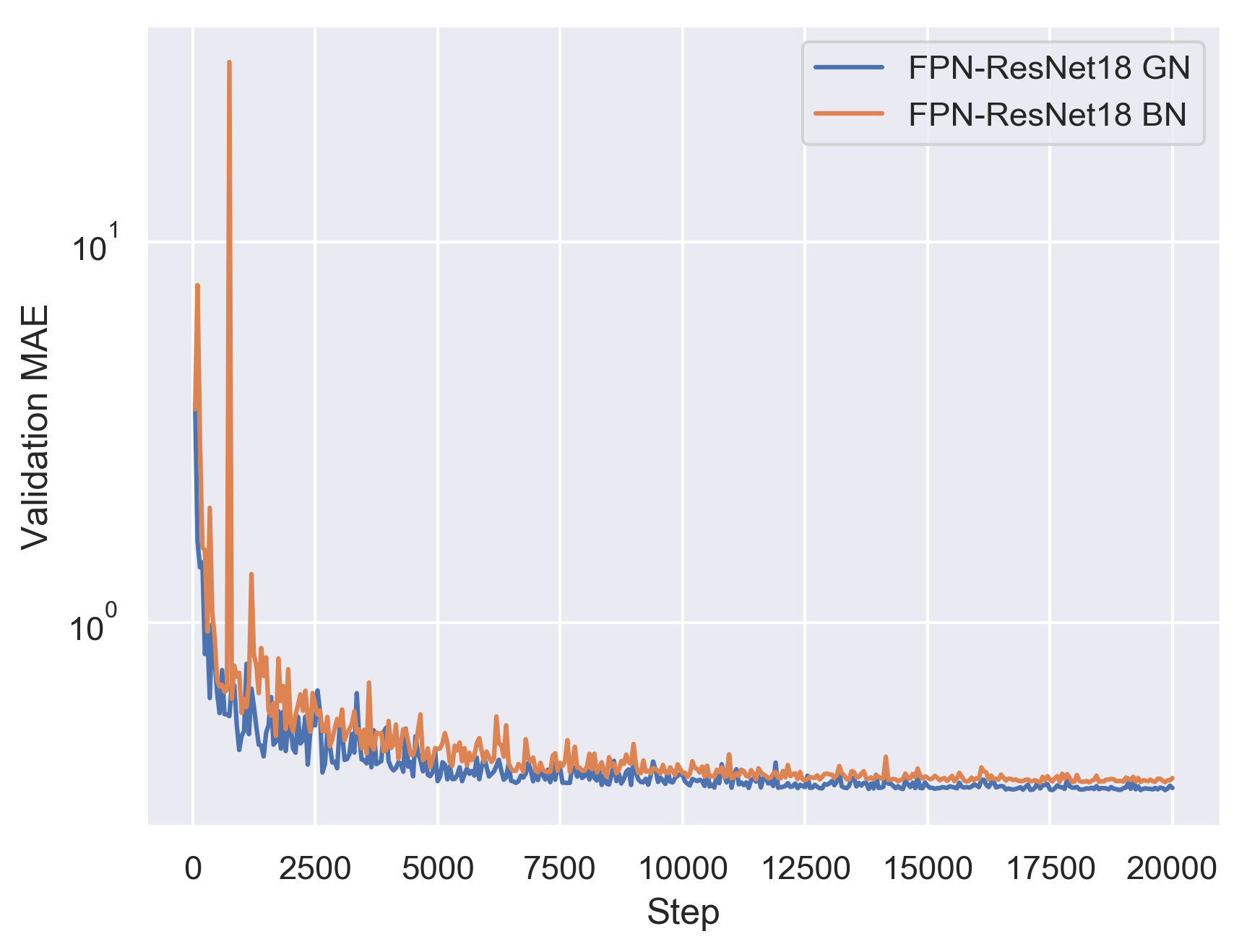}}
	\subfigure[]{
		\includegraphics[width=0.235\linewidth]{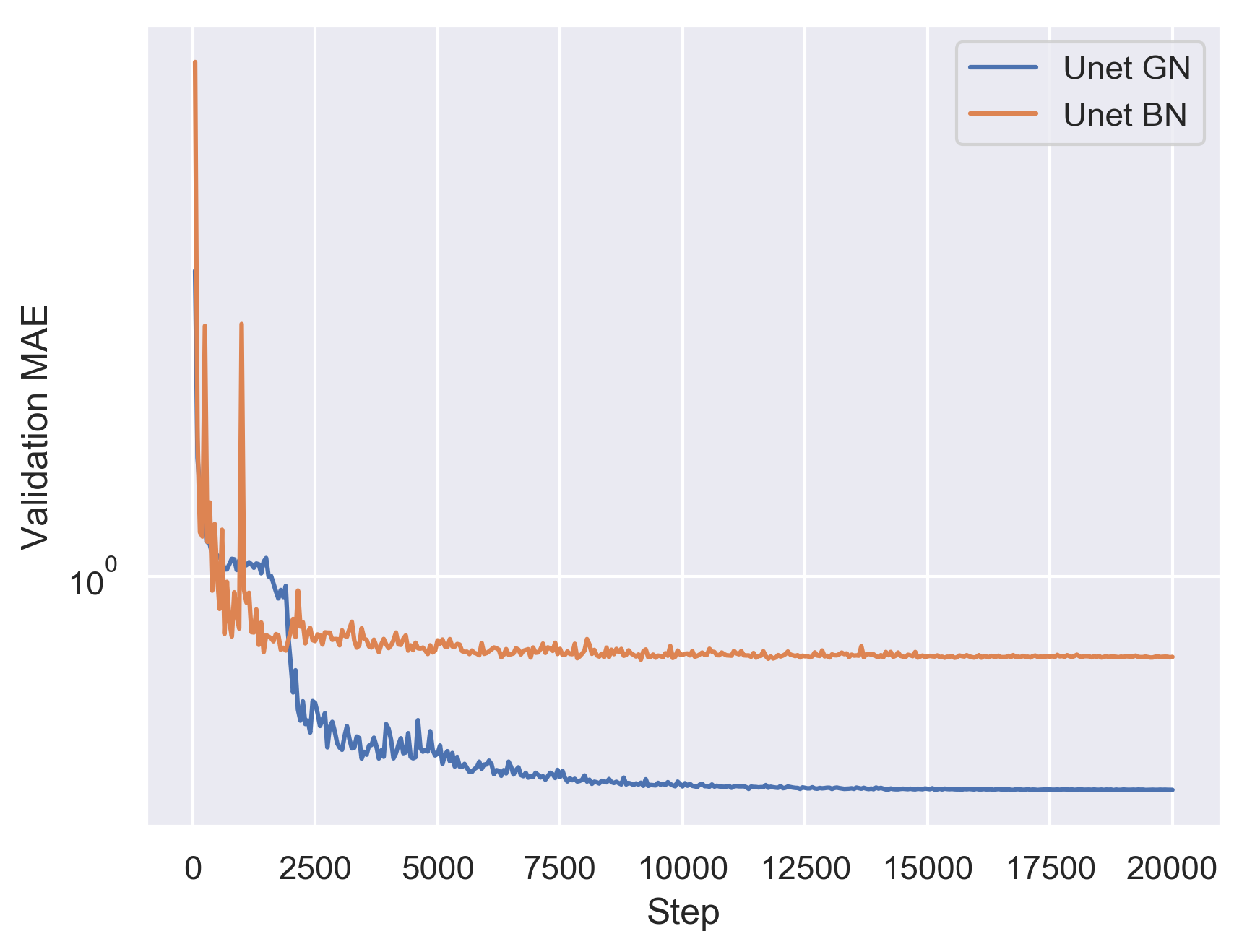}}
	\subfigure[]{
		\includegraphics[width=0.235\linewidth]{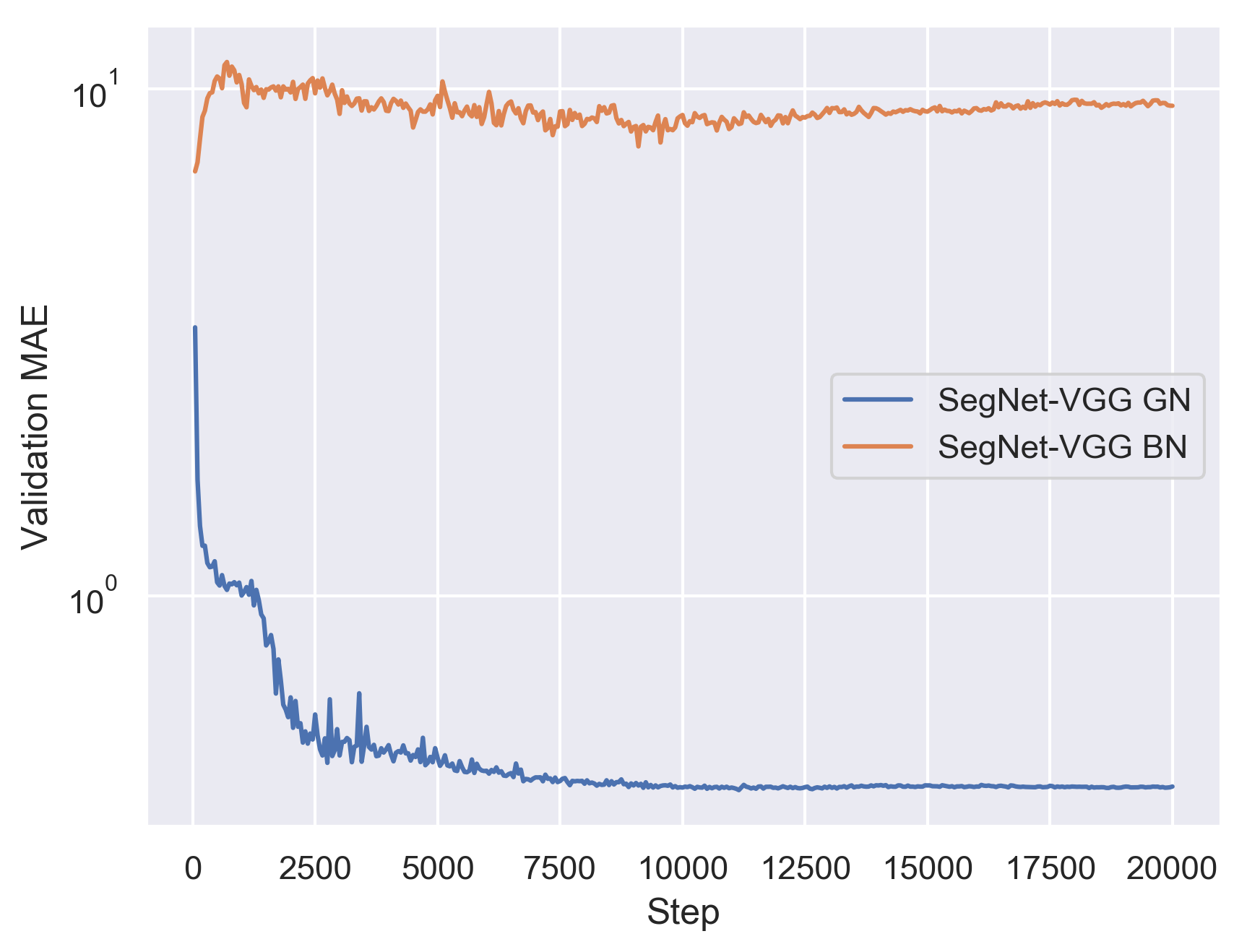}}
	\caption{The comparison of GN and BN for different network architectures. (a-d) show the trend of the loss function in the training set, and (e-f) show the changes of testing MAE in the validation set.}
	\label{fig:bganggn}
\end{figure*}

Fig.~\ref{fig:bganggn} illustrates the variation trend of training loss function and validation MAE for different network architectures with BN and GN. 
A general trend in the training loss of the comparison between GN and BN is that DNN surrogates based on GN converge faster and stably than those based on BN, indicating that GN is more beneficial to speeding up the training process.
Sometimes, GN can make the model achieve a slightly better performance than using BN.
In addition, by investigating the validation error in Fig.~\ref{fig:bganggn}(e-f), it can be obviously observed that the generalization capability of SegNet and Unet models with BN has experienced a clear drop, especially by seeing the severe model degradation for BN-based SegNet.
The reason may be that the inputs of these models, namely the layout images of components in our task, are sparser than natural images, causing the statistical errors among different mini-batches and further probably leading to a large deviation in the model inference results.
Therefore, GN, that is, the group normalization method would be more suitable and helpful for the construction of DNN surrogate models in the HSL-TFP task.

\begin{figure*}[!htbp]
	\centering
	\subfigure[MAE in Case 1]{
		\includegraphics[width=0.31\linewidth]{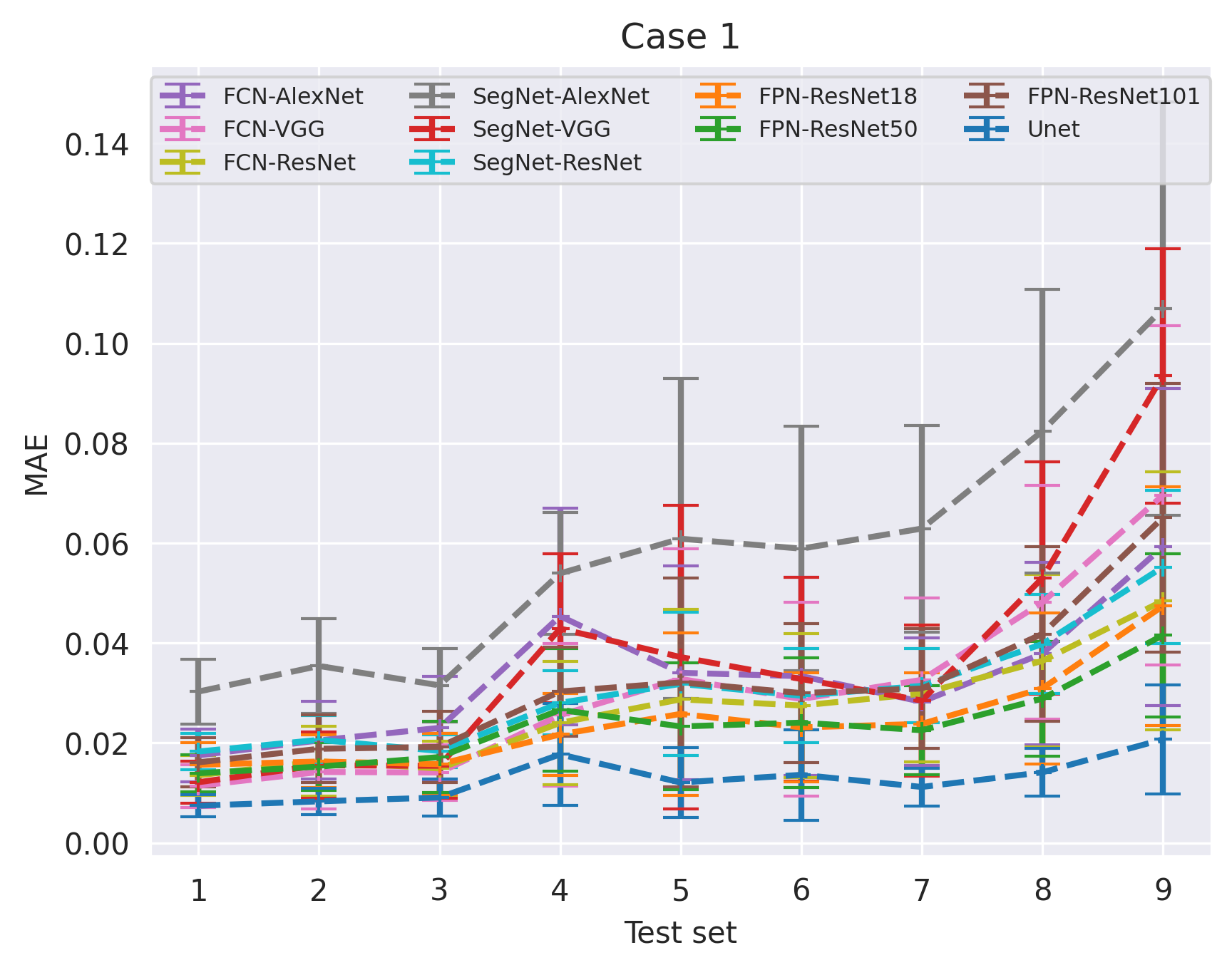}
		\label{fig:mae:case1}
	}
	\subfigure[MAE in Case 2]{
		\includegraphics[width=0.31\linewidth]{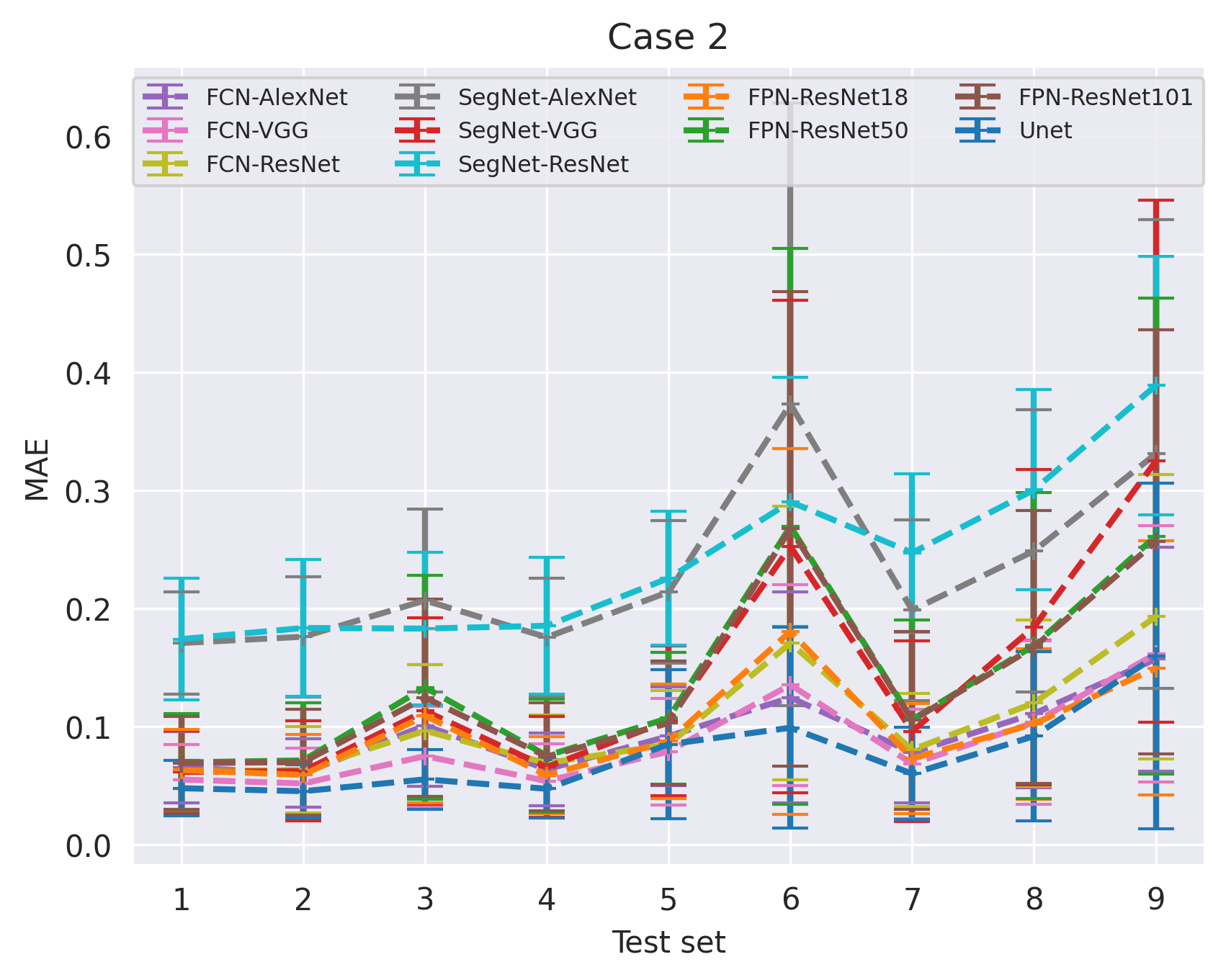}
		\label{fig:mae:case2}
	}
	\subfigure[MAE in Case 3]{
		\includegraphics[width=0.31\linewidth]{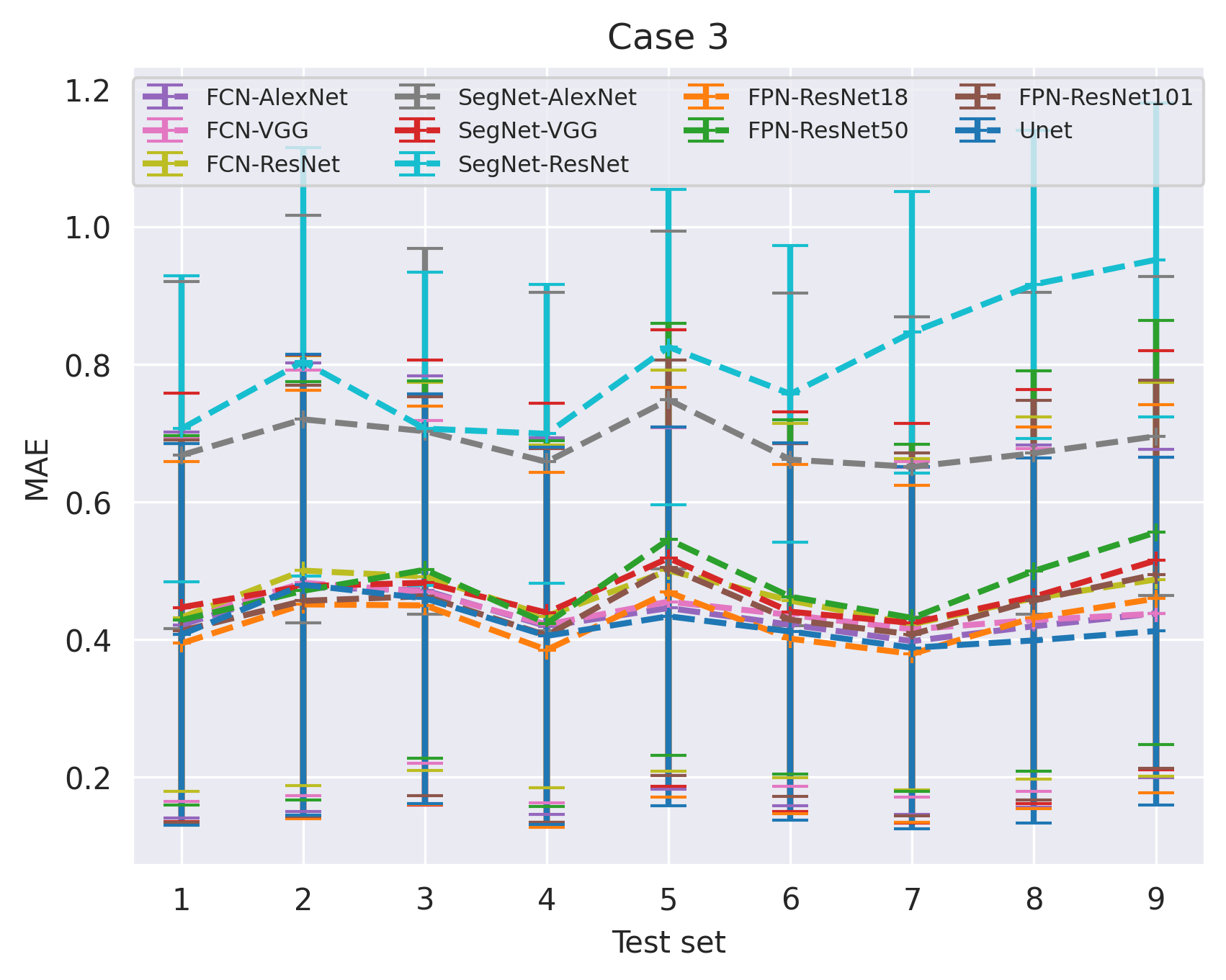}
		\label{fig:mae:case3}
	}
	\\
	\subfigure[Max AE in Case 1]{
		\includegraphics[width=0.31\linewidth]{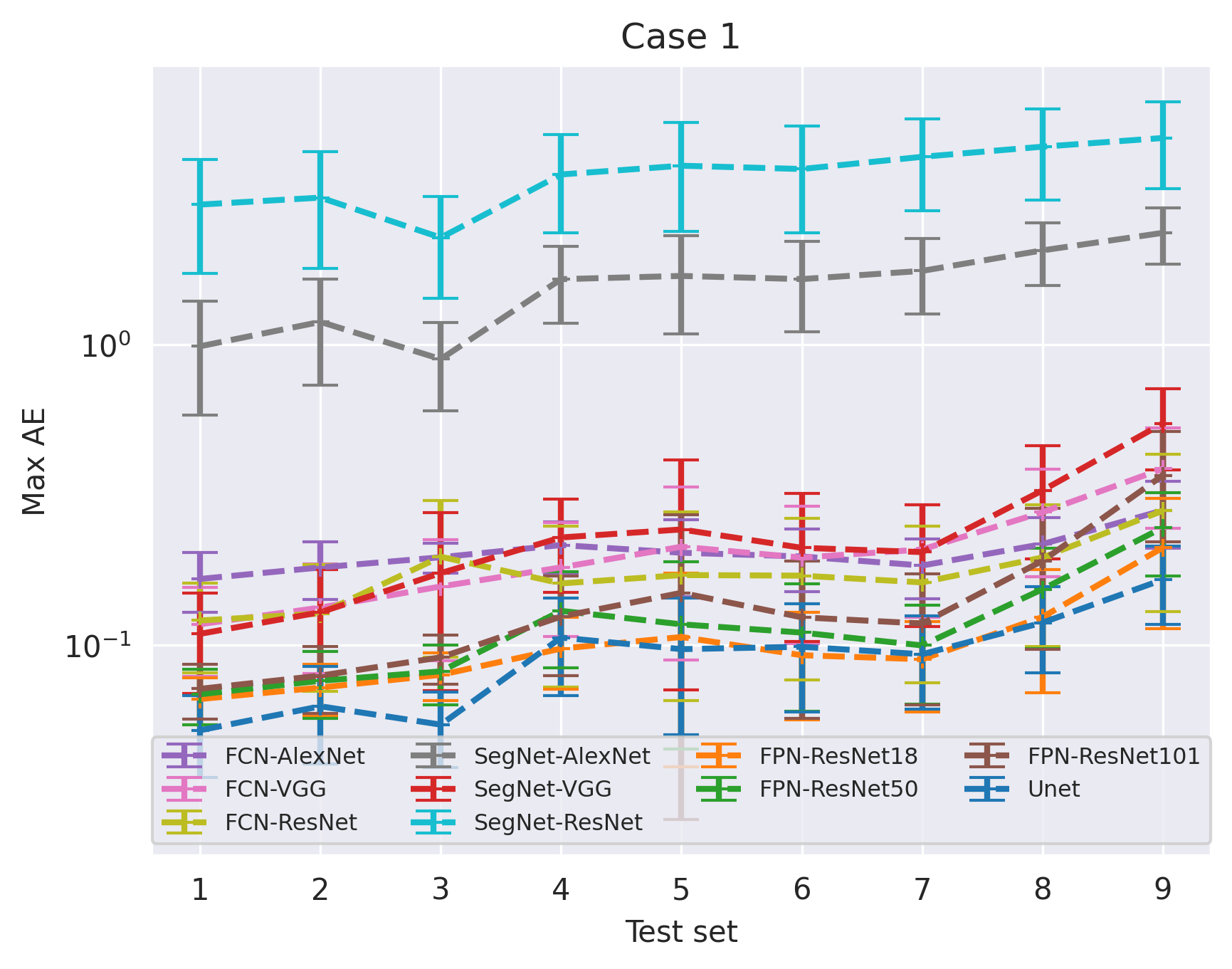}
		\label{fig:maxae:case1}
	}
	\subfigure[Max AE in Case 2]{
		\includegraphics[width=0.31\linewidth]{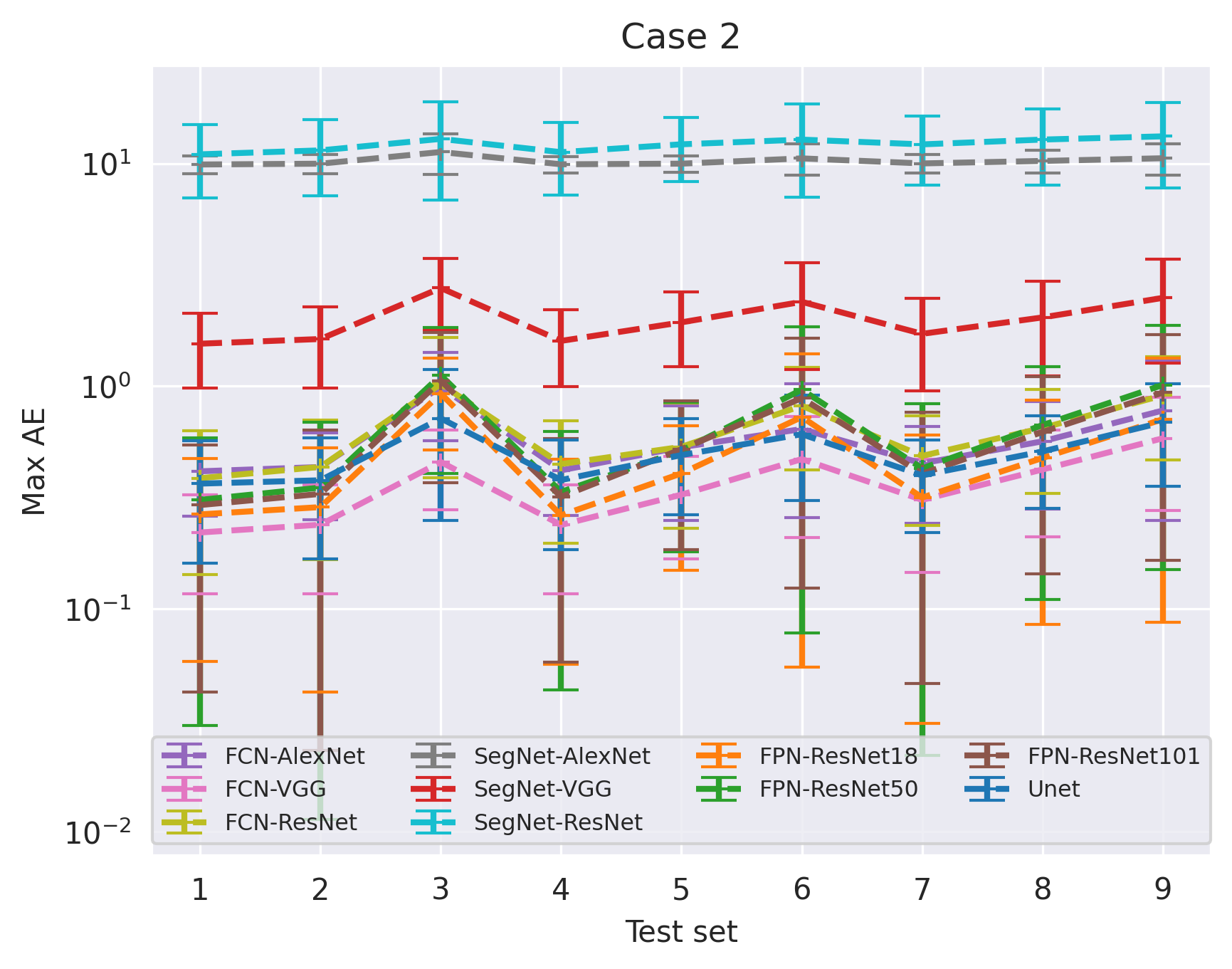}
		\label{fig:maxae:case2}
	}
	\subfigure[Max AE in Case 3]{
		\includegraphics[width=0.31\linewidth]{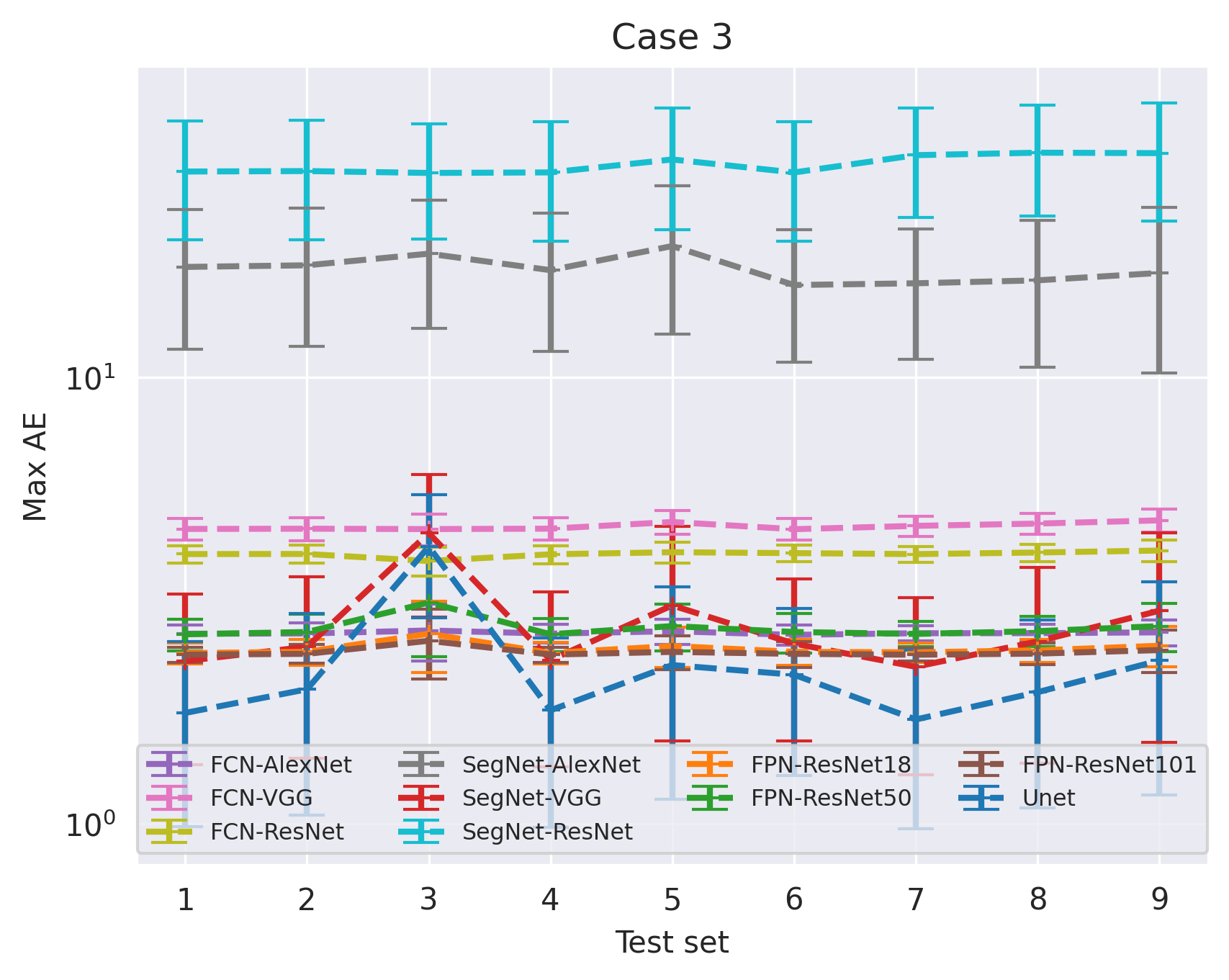}
		\label{fig:maxae:case3}
	}
	\caption{The comparison of the overall prediction performance in nine test sets of ten representative DNN surrogates in three cases. (a-c) MAE of surrogates in case 1, 2 and 3. (d-f) Max AE of surrogates in case 1, 2 and 3. (unit: K)}
	\label{fig:mae-maxae}
\end{figure*}

\begin{figure*}[!htbp]
	\centering
	\subfigure[MT-AE in Case 1]{
		\includegraphics[width=0.31\linewidth]{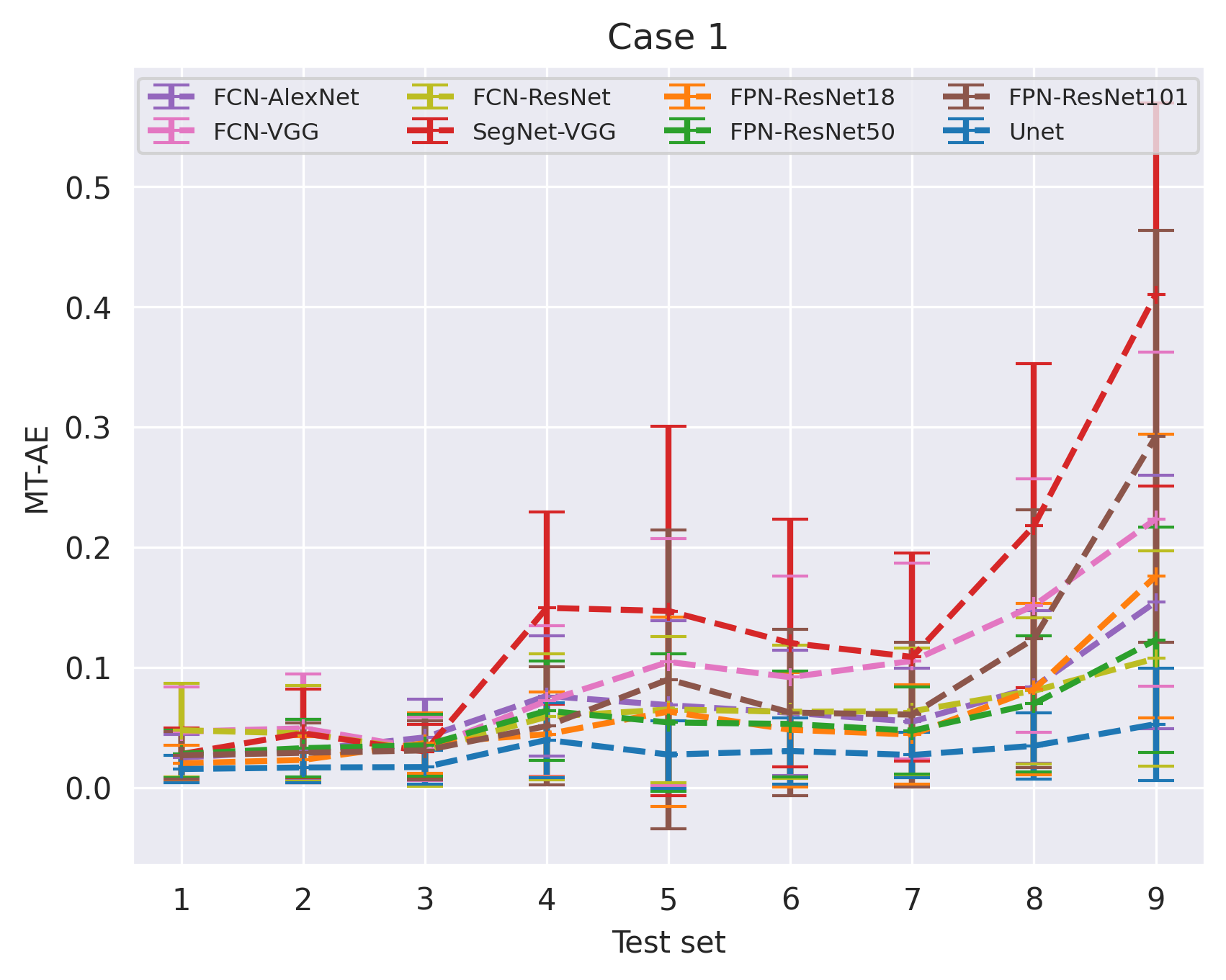}
		\label{fig:mtae:case1}
	}
	\subfigure[MT-AE in Case 2]{
		\includegraphics[width=0.31\linewidth]{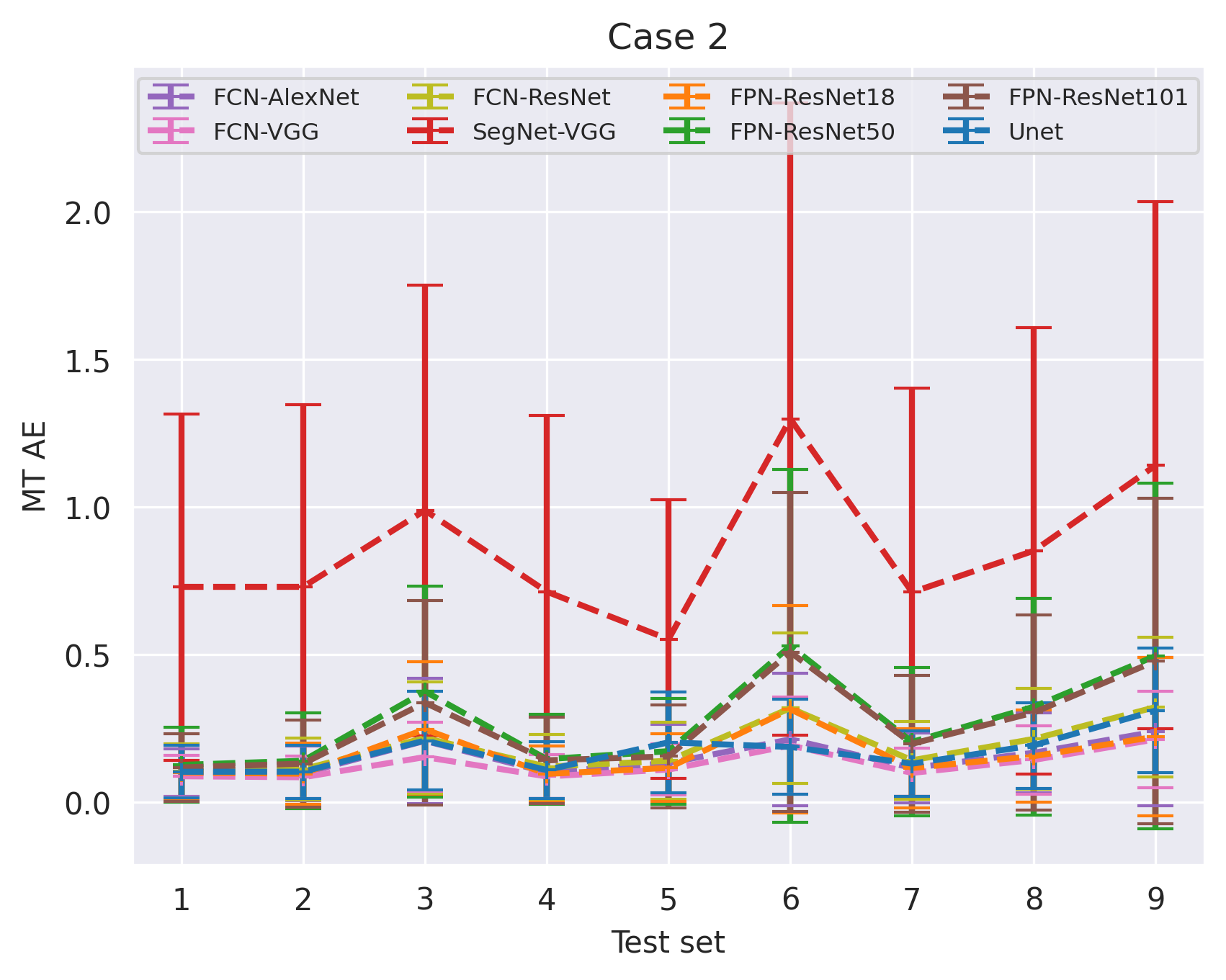}
		\label{fig:mtae:case2}
	}
	\subfigure[MT-AE in Case 3]{
		\includegraphics[width=0.31\linewidth]{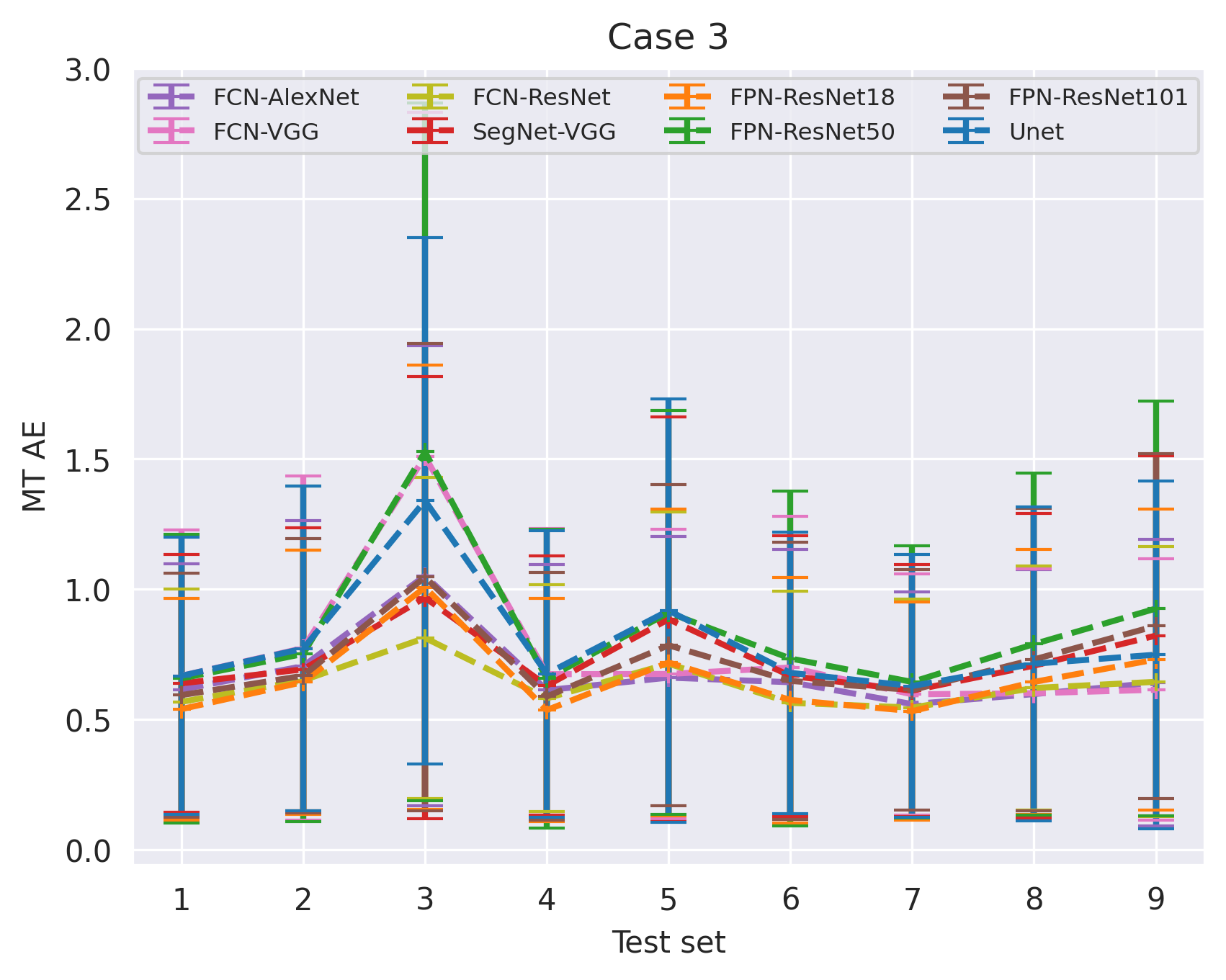}
		\label{fig:mtae:case3}
	}
	\\
	\subfigure[MT-PAE in Case 1]{
		\includegraphics[width=0.31\linewidth]{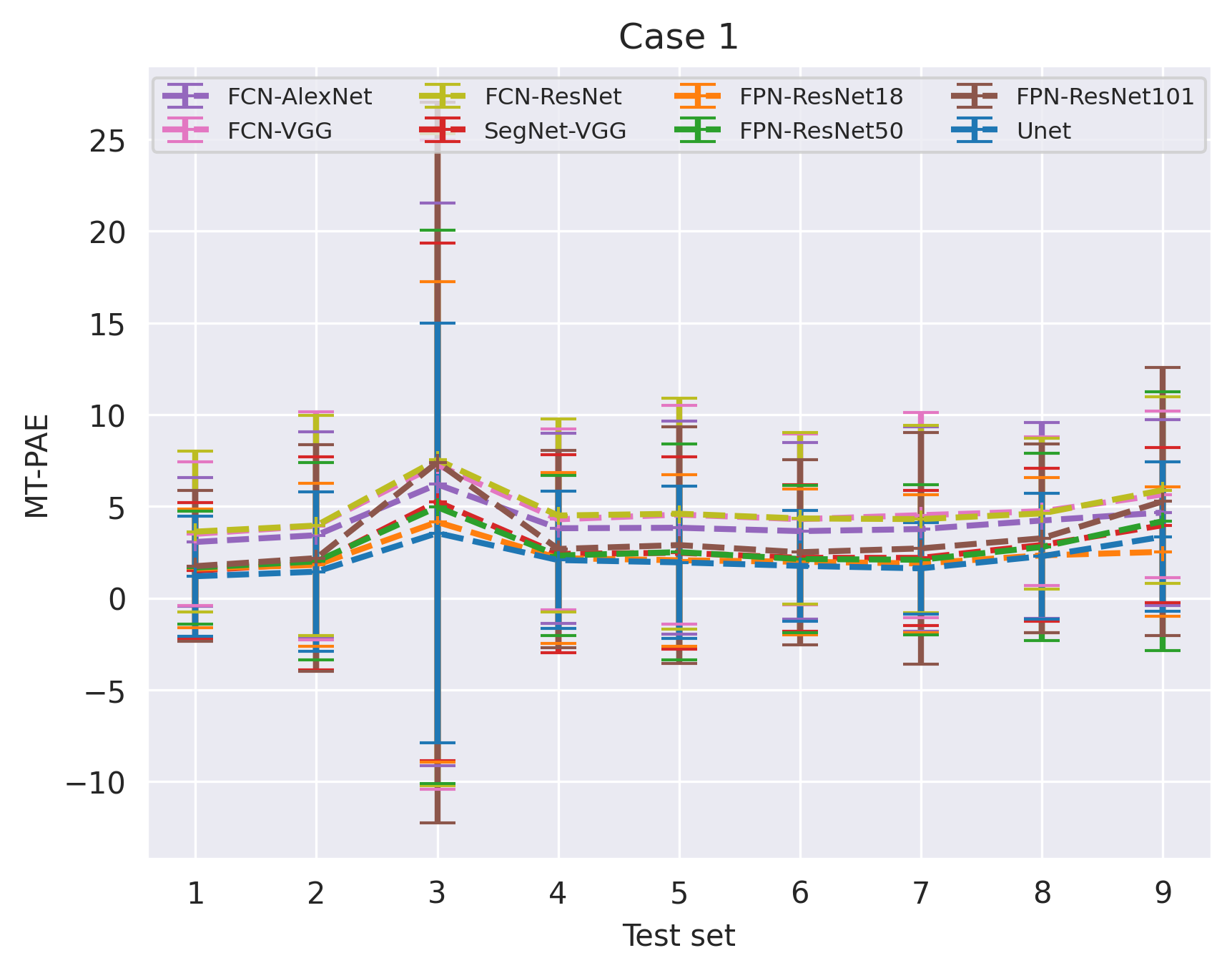}
		\label{fig:mtpae:case1}
	}
	\subfigure[MT-PAE in Case 2]{
		\includegraphics[width=0.31\linewidth]{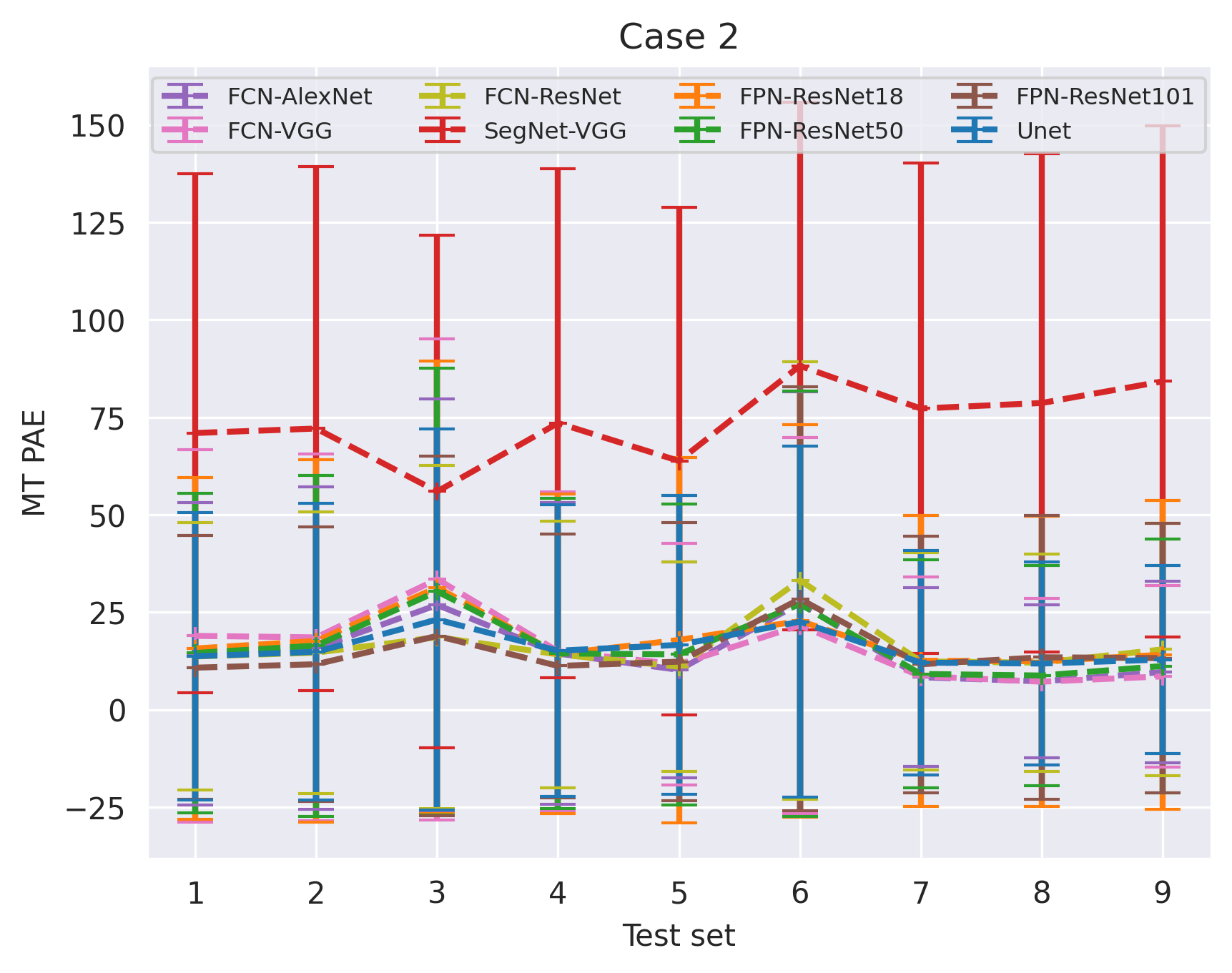}
		\label{fig:mtpae:case2}
	}
	\subfigure[MAE in Case 3]{
		\includegraphics[width=0.31\linewidth]{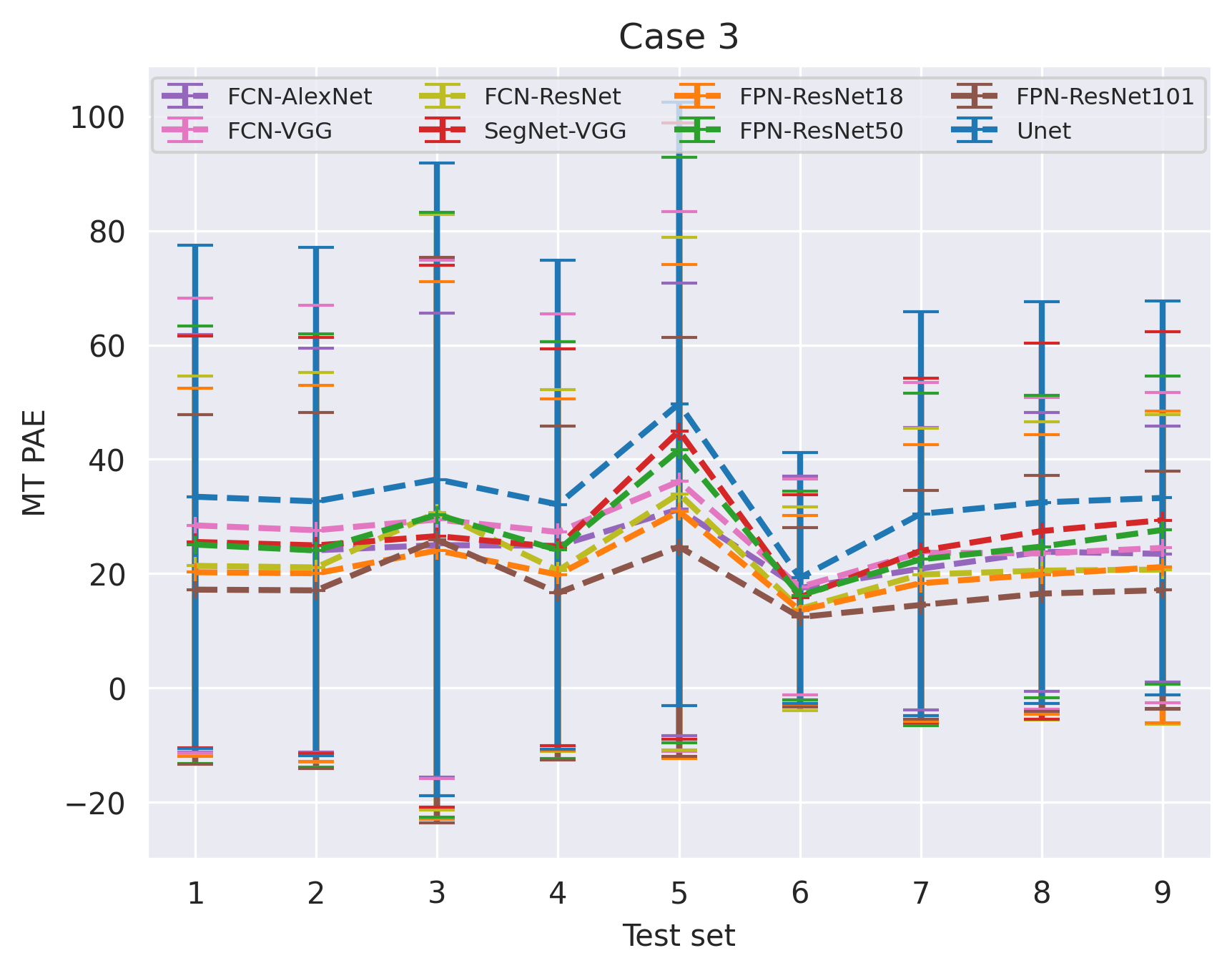}
		\label{fig:mtpae:case3}
	}
	\caption{The comparison of the prediction performance on the maximum temperature in nine test sets of eight representative DNN surrogates in three cases. (a-c) MT-AE of surrogates in case 1, 2 and 3 (unit: K). (d-f) MT-PAE of surrogates in case 1, 2 and 3 (unit: cell).}
	\label{fig:mtae-mtpae}
\end{figure*}

\begin{figure*}[!htbp]
	\centering
	\subfigure[BMAE$_D$ in Case 1]{
		\includegraphics[width=0.31\linewidth]{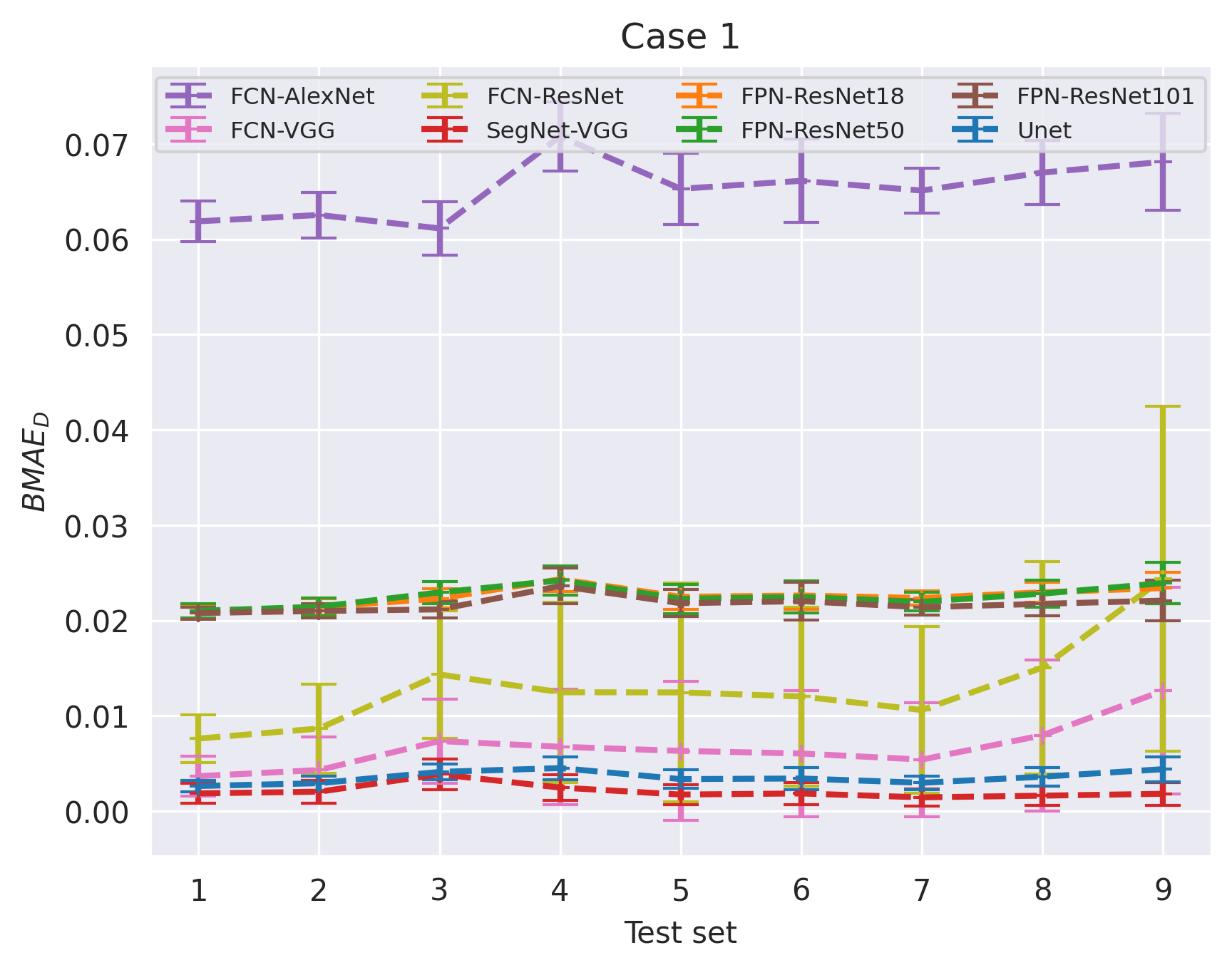}
		\label{fig:bmaed:case1}
	}
	\subfigure[BMAE$_D$ in Case 2]{
		\includegraphics[width=0.31\linewidth]{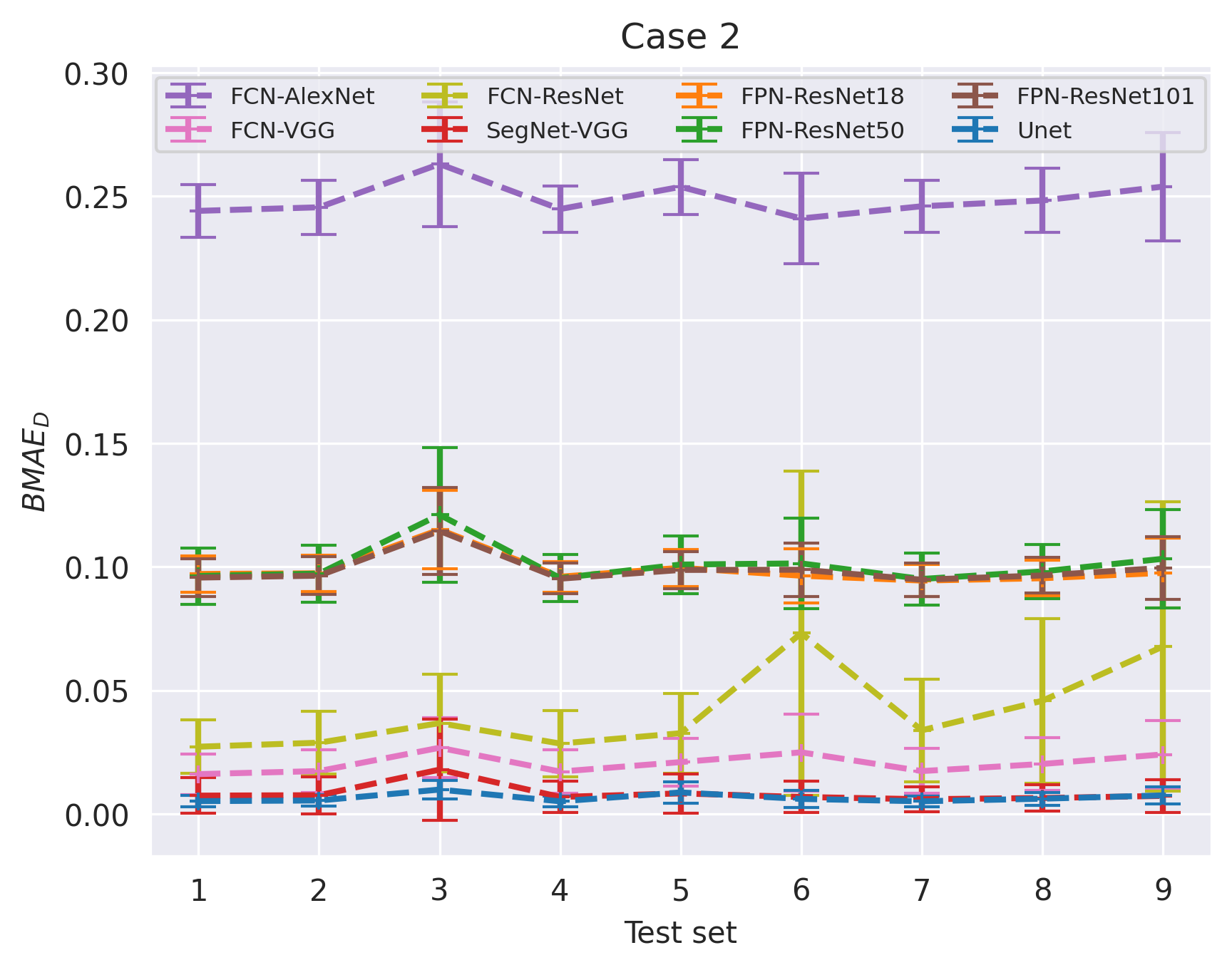}
		\label{fig:bmaed:case2}
	}
	\subfigure[BMAE$_D$ in Case 3]{
		\includegraphics[width=0.31\linewidth]{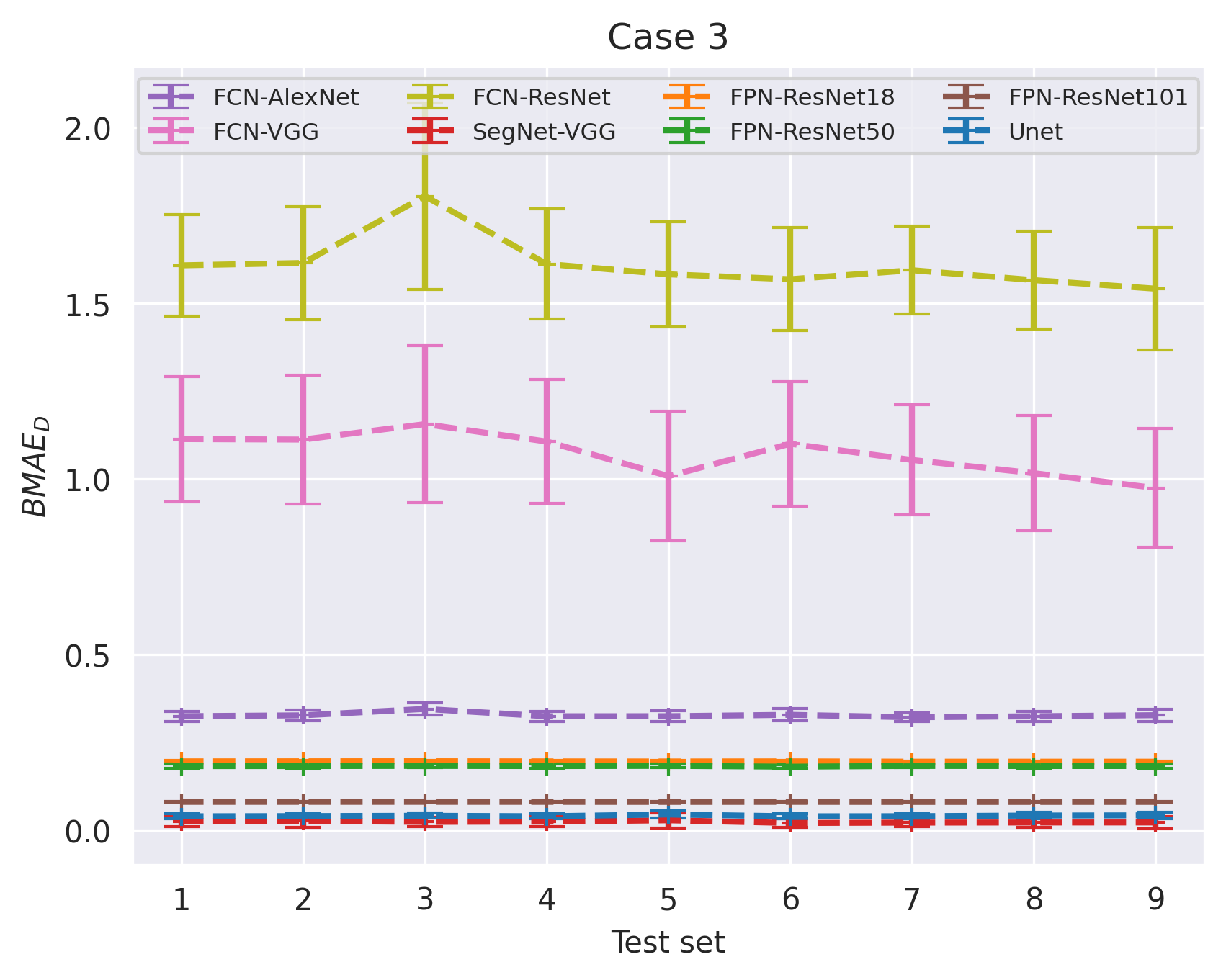}
		\label{fig:bmaed:case3}
	}
	\\
	\subfigure[BMAE$_N$ in Case 1]{
		\includegraphics[width=0.31\linewidth]{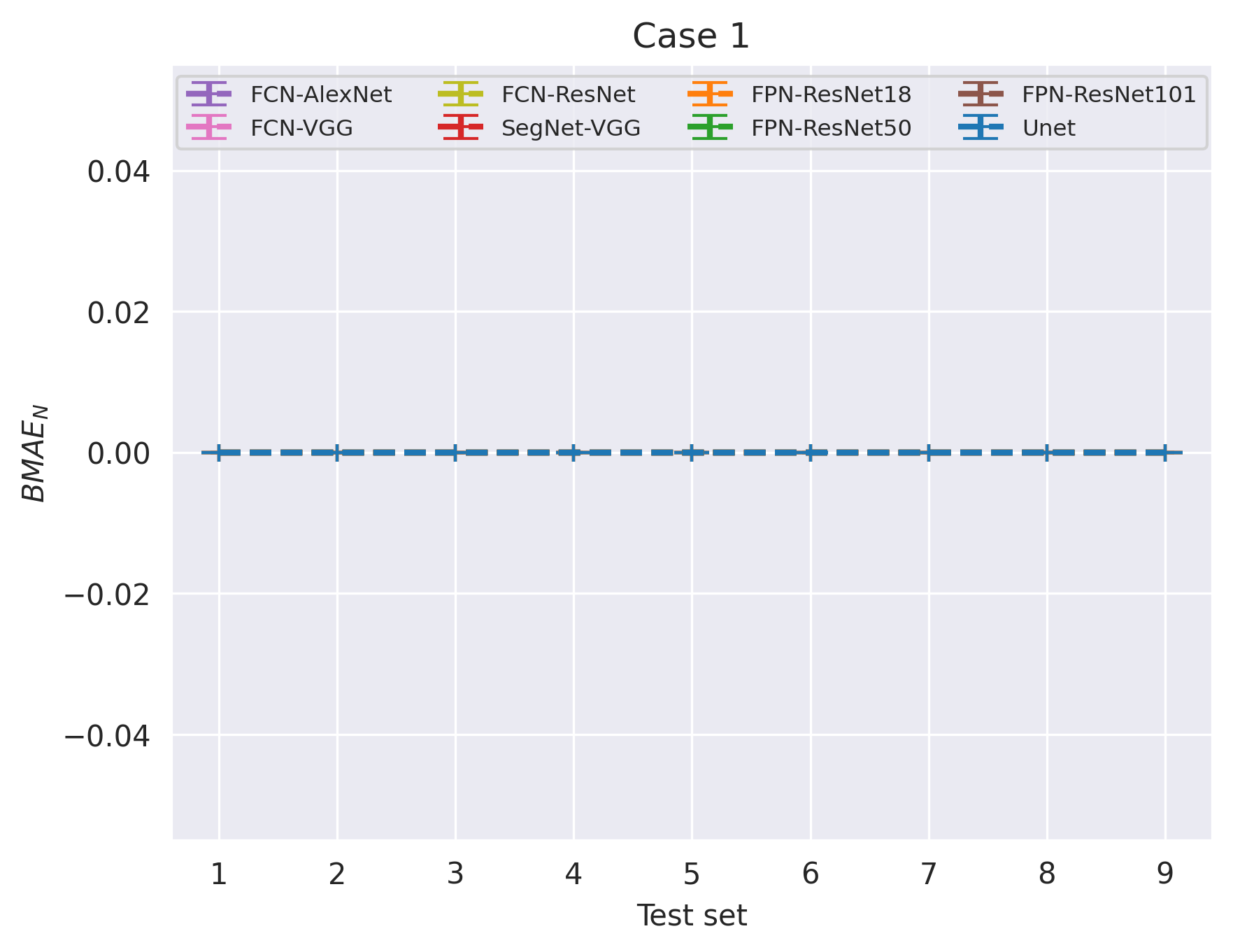}
		\label{fig:bmaen:case1}
	}
	\subfigure[BMAE$_N$ in Case 2]{
		\includegraphics[width=0.31\linewidth]{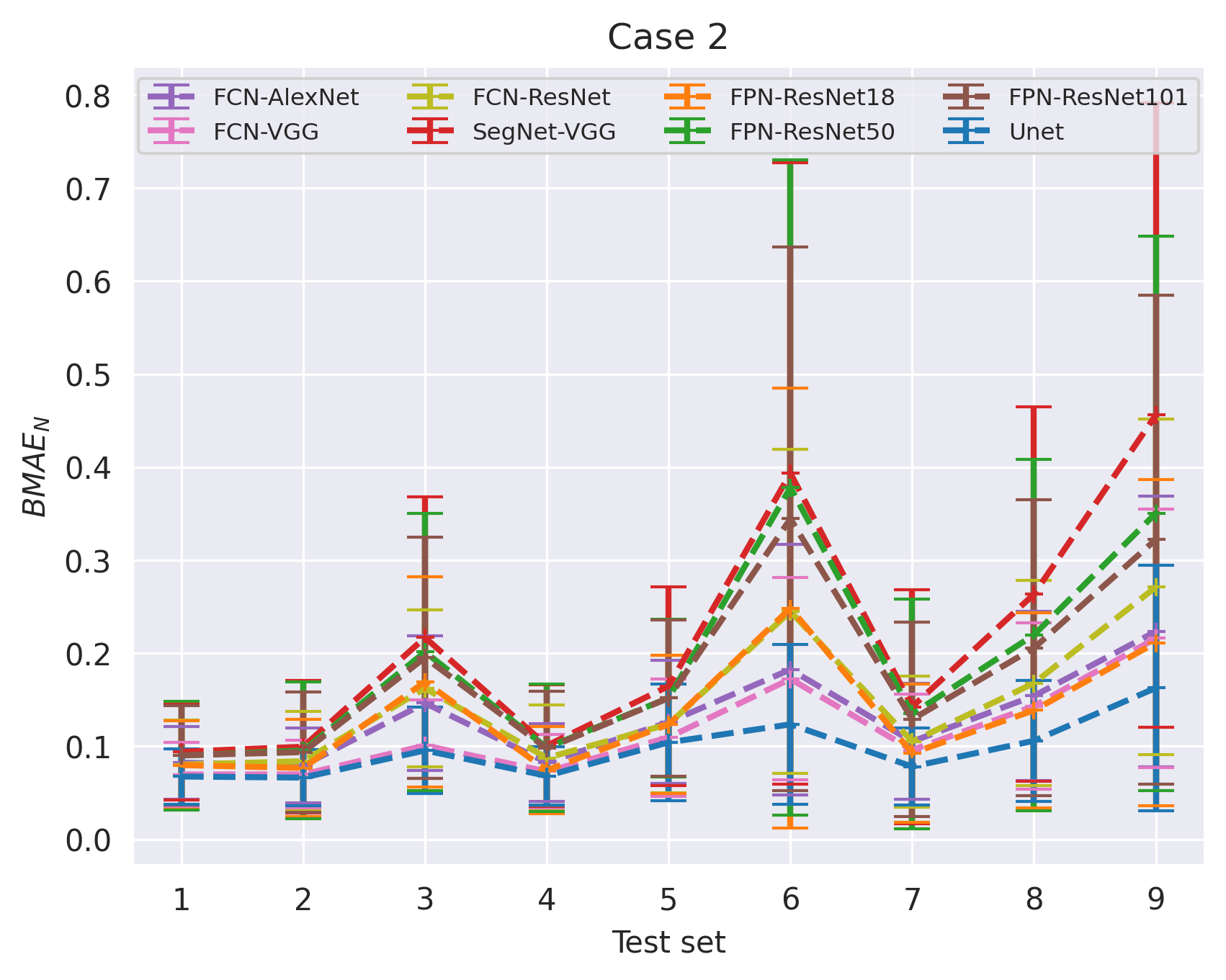}
		\label{fig:bmaen:case2}
	}
	\subfigure[BMAE$_N$ in Case 3]{
		\includegraphics[width=0.31\linewidth]{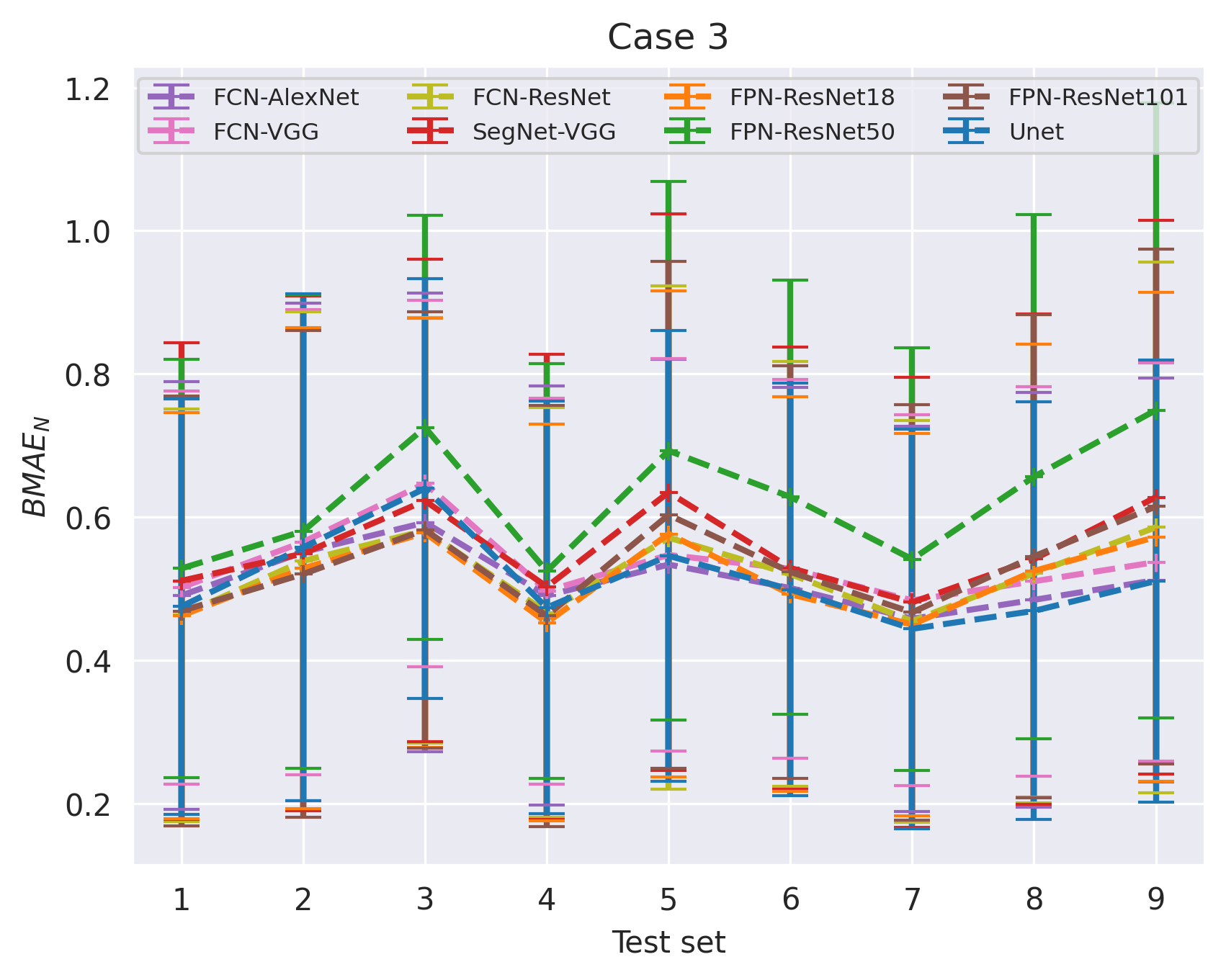}
		\label{fig:bmaen:case3}
	}
	\caption{The comparison of the prediction performance on the boundary temperature in nine test sets of eight representative DNN surrogates in three cases. (a-c) BMAE$_D$ for Dirichlet boundary of surrogates in case 1, 2 and 3. (d-f) BMAE$_N$ for Neumann boundary of surrogates in case 1, 2 and 3. (unit: K)}
	\label{fig:bmaed-bmaen}
\end{figure*}

\begin{figure*}[!htbp]
	\centering
	\subfigure[CMAE in Case 1]{
		\includegraphics[width=0.31\linewidth]{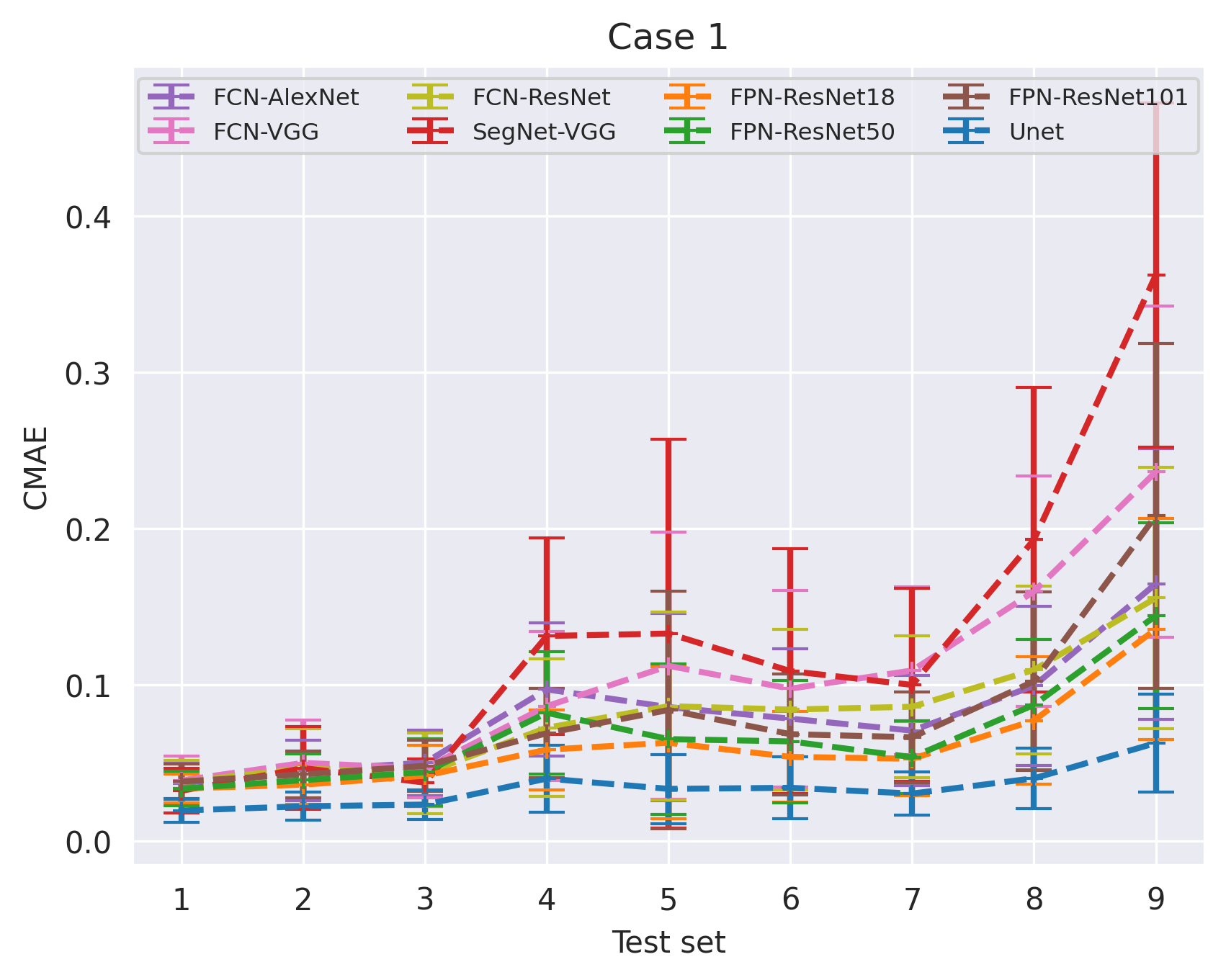}
		\label{fig:cmae:case1}
	}
	\subfigure[CMAE in Case 2]{
		\includegraphics[width=0.31\linewidth]{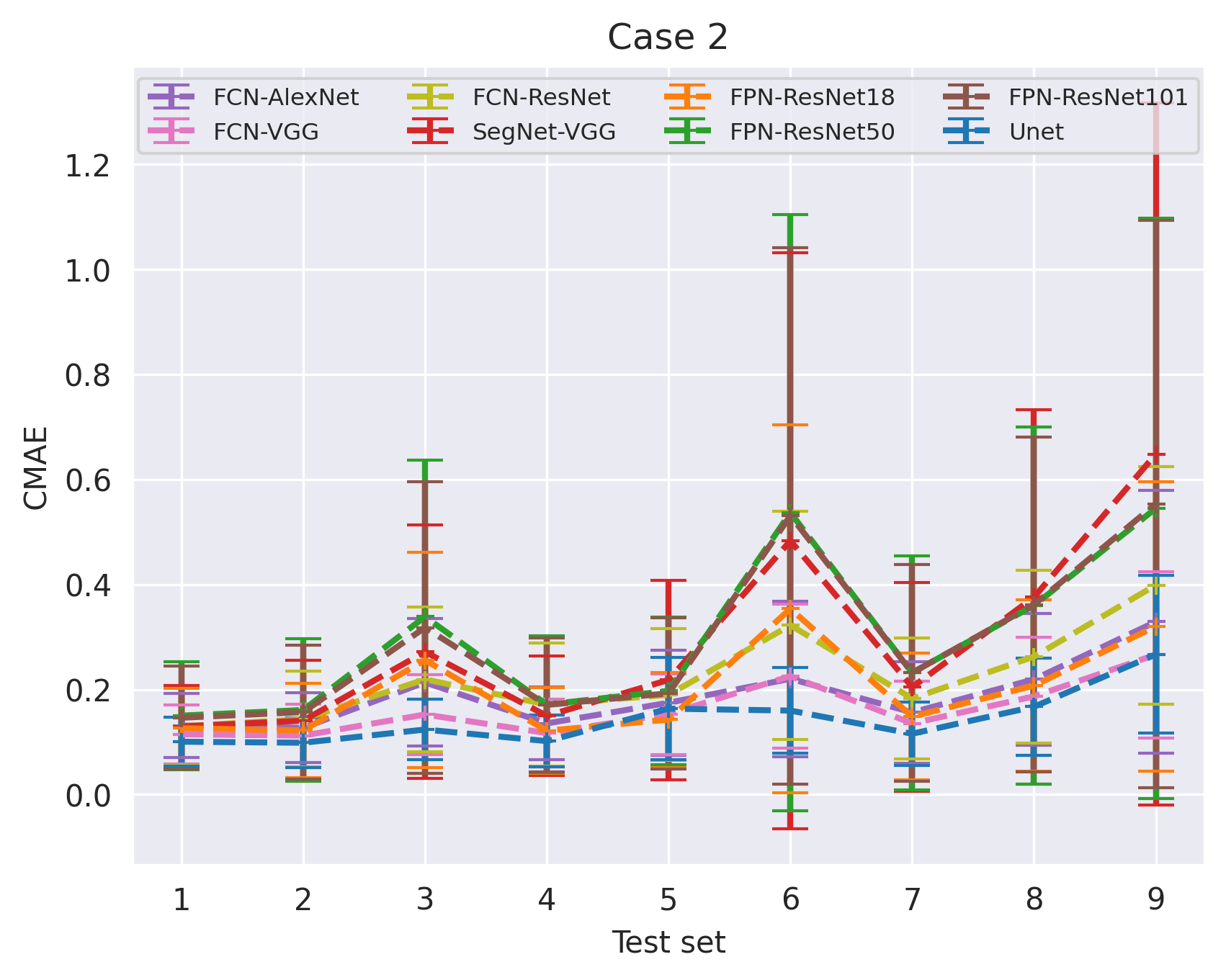}
		\label{fig:cmae:case2}
	}
	\subfigure[CMAE in Case 3]{
		\includegraphics[width=0.31\linewidth]{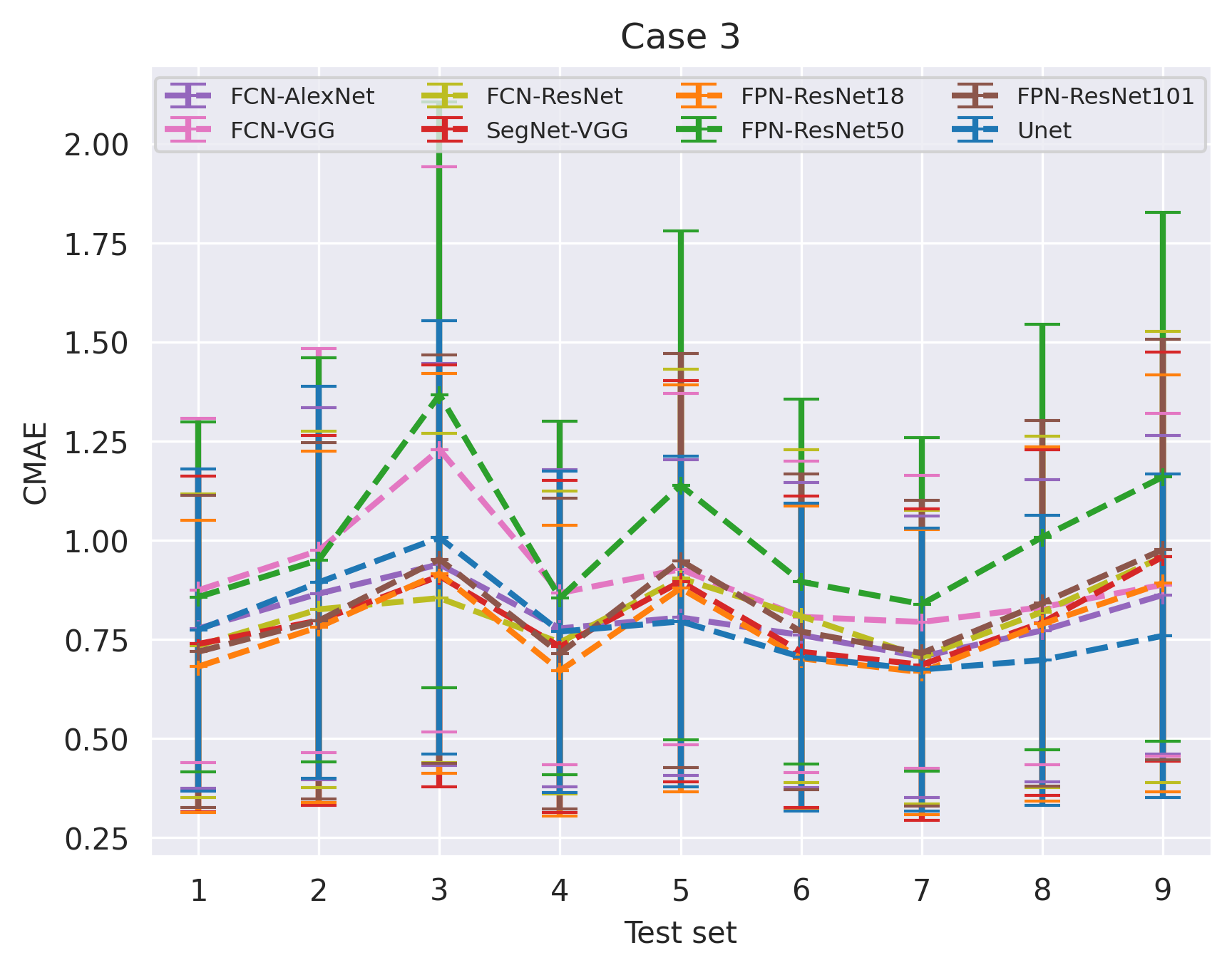}
		\label{fig:cmae:case3}
	}
	\caption{The comparison of the prediction performance on the component temperature in nine test sets of eight representative DNN surrogates in three cases. (a-c) CMAE of surrogates in case 1, 2 and 3. (unit: K)}
	\label{fig:cmae}
\end{figure*}

\begin{figure*}[!htbp]
	\centering
	\subfigure[G-MAE in Case 1]{
		\includegraphics[width=0.31\linewidth]{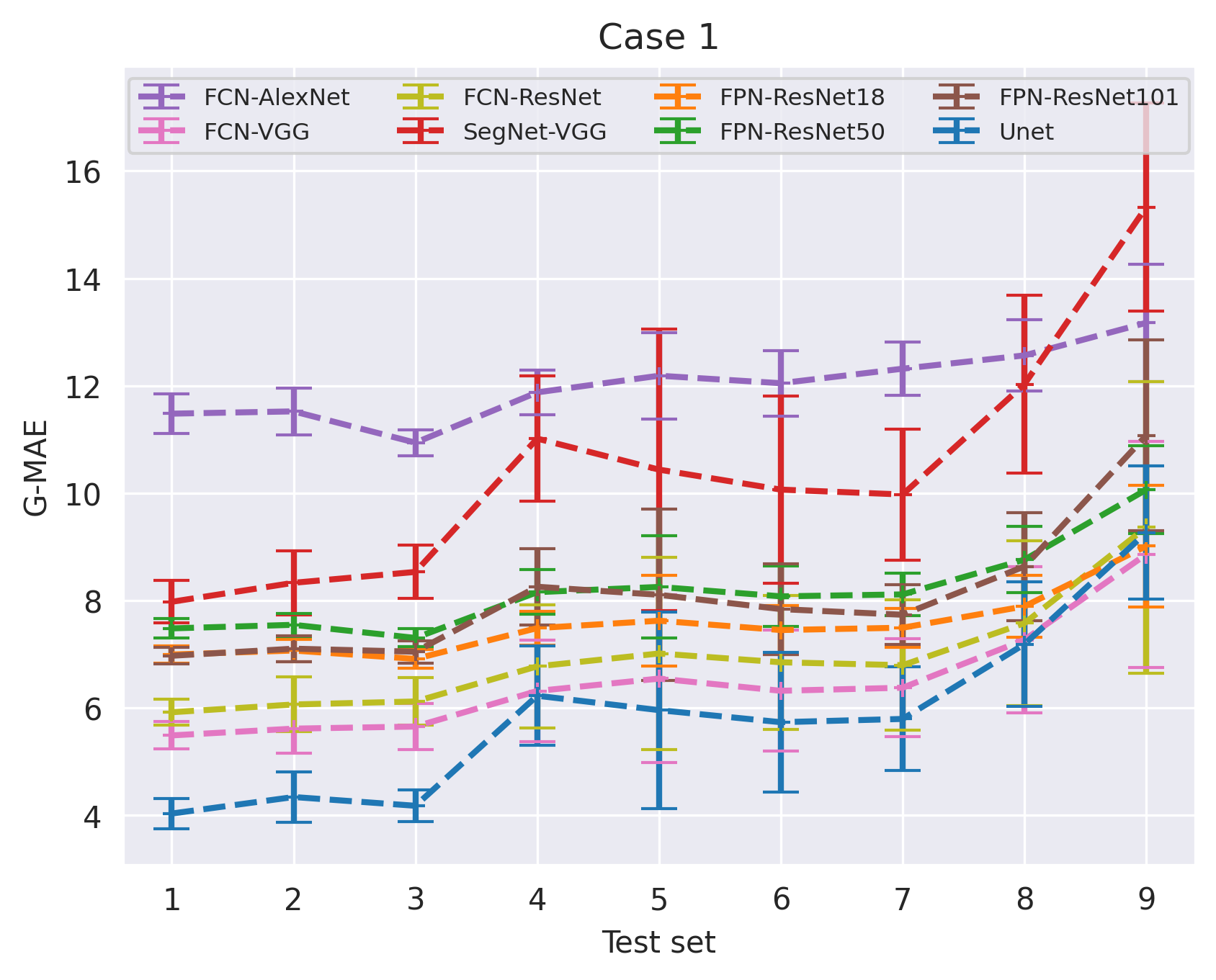}
		\label{fig:gmae:case1}
	}
	\subfigure[G-MAE in Case 2]{
		\includegraphics[width=0.31\linewidth]{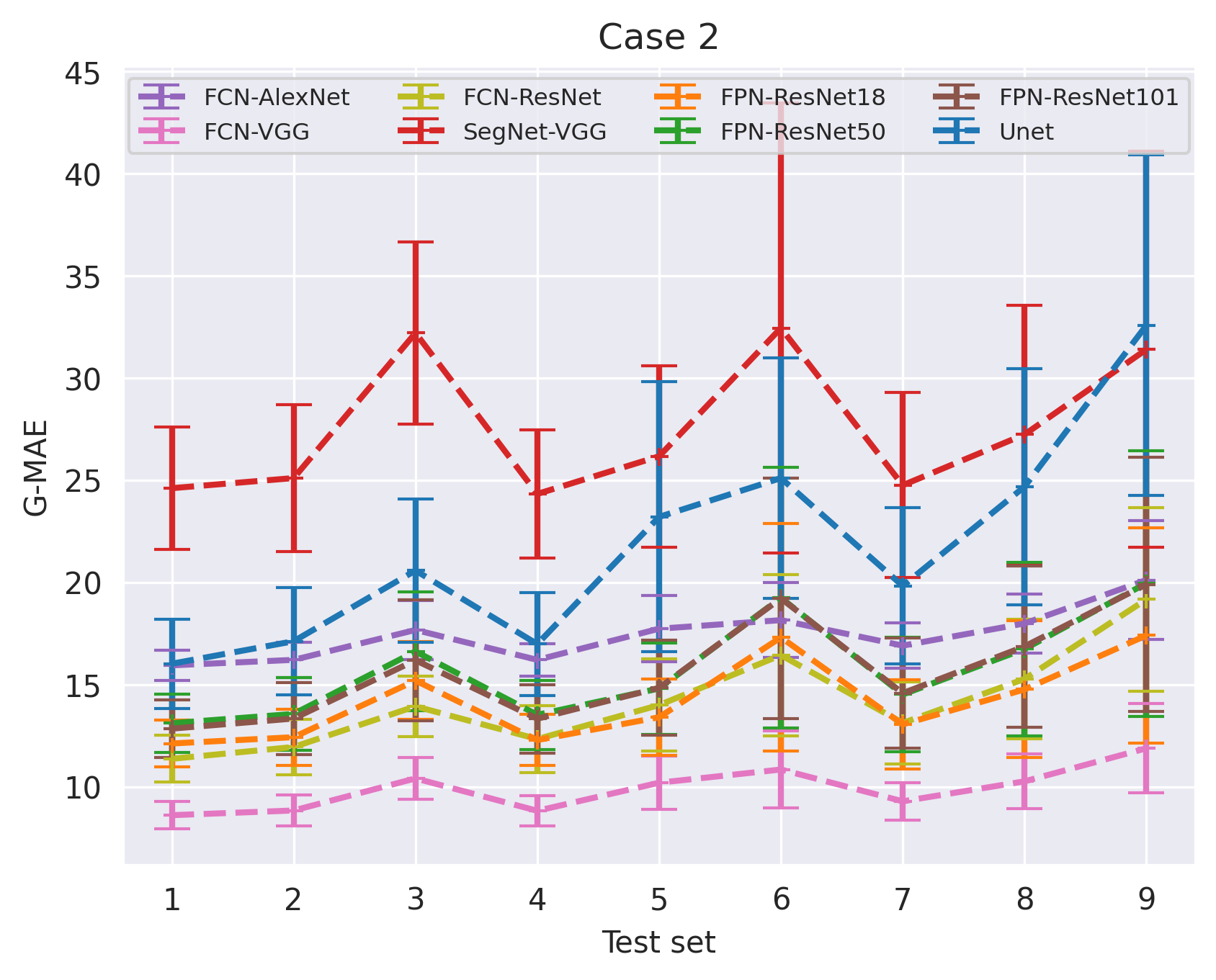}
		\label{fig:gmae:case2}
	}
	\subfigure[G-MAE in Case 3]{
		\includegraphics[width=0.31\linewidth]{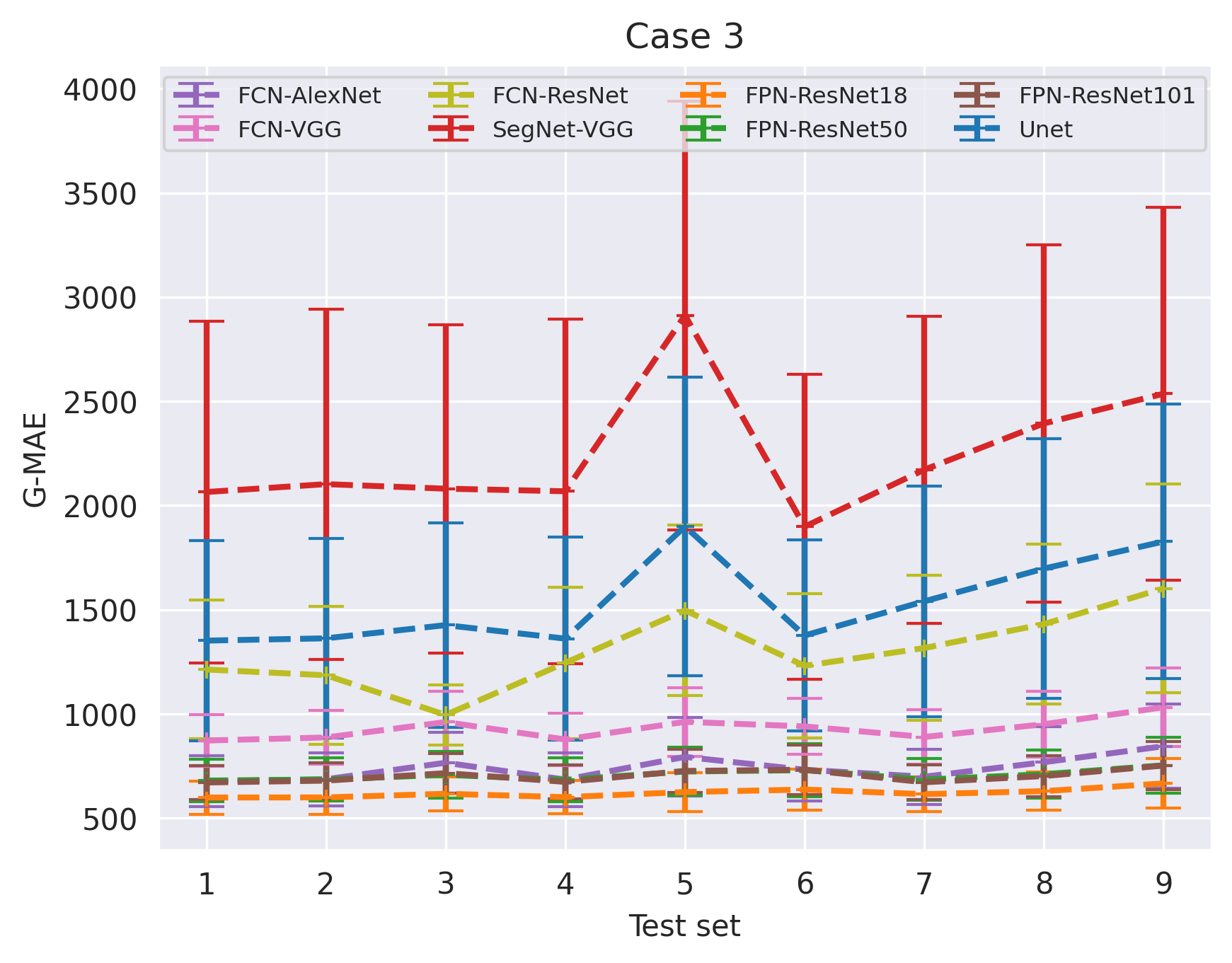}
		\label{fig:gmae:case3}
	}
	\caption{The comparison of the prediction performance on the gradient field in nine test sets of eight representative DNN surrogates in three cases. (a-c) G-MAE of surrogates in case 1, 2 and 3. (unit: K/m)}
	\label{fig:gmae}
\end{figure*}

\begin{figure*}[!htbp]
	\centering
	\subfigure[Lap-MAE in Case 1]{
		\includegraphics[width=0.31\linewidth]{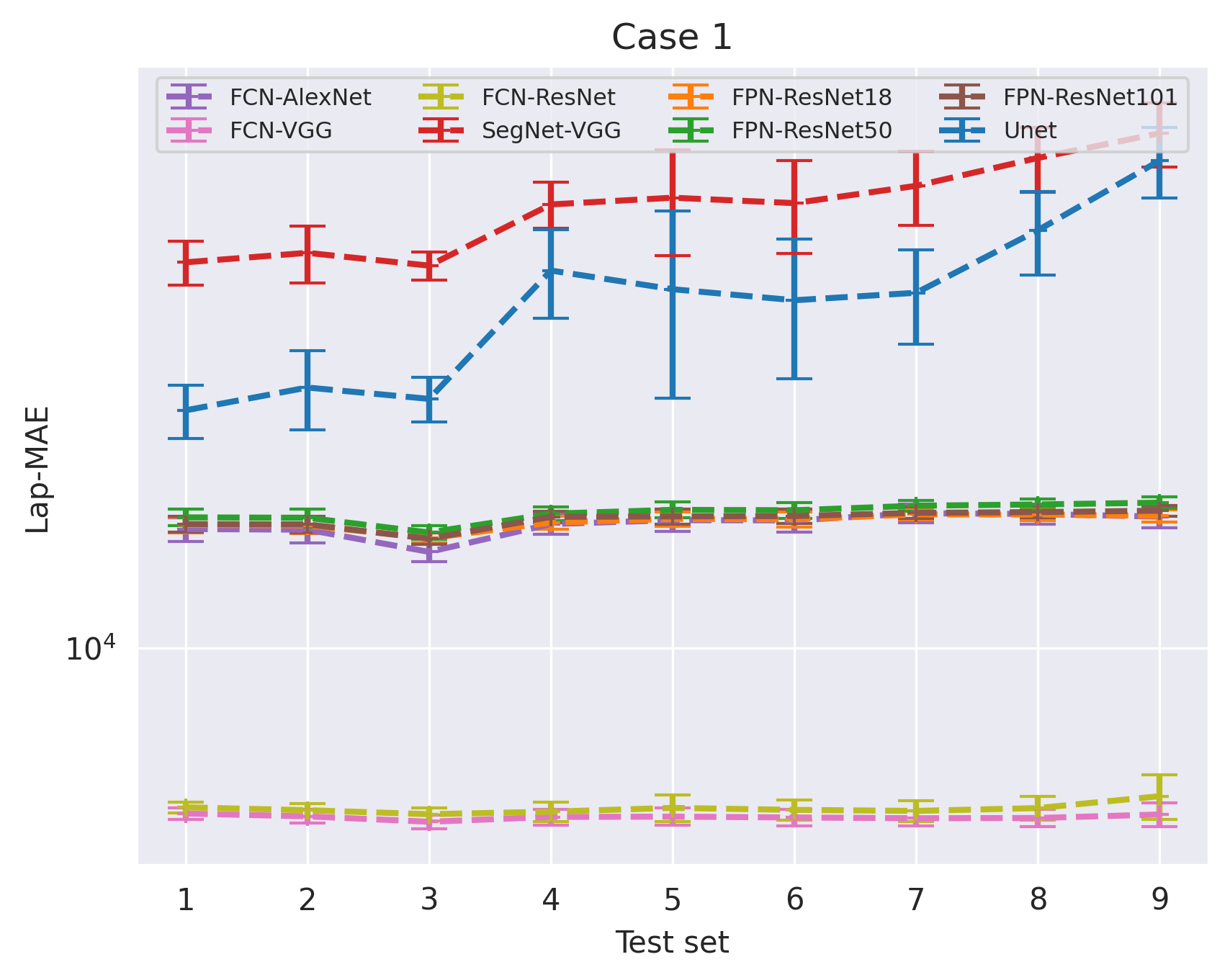}
		\label{fig:lapmae:case1}
	}
	\subfigure[Lap-MAE in Case 2]{
		\includegraphics[width=0.31\linewidth]{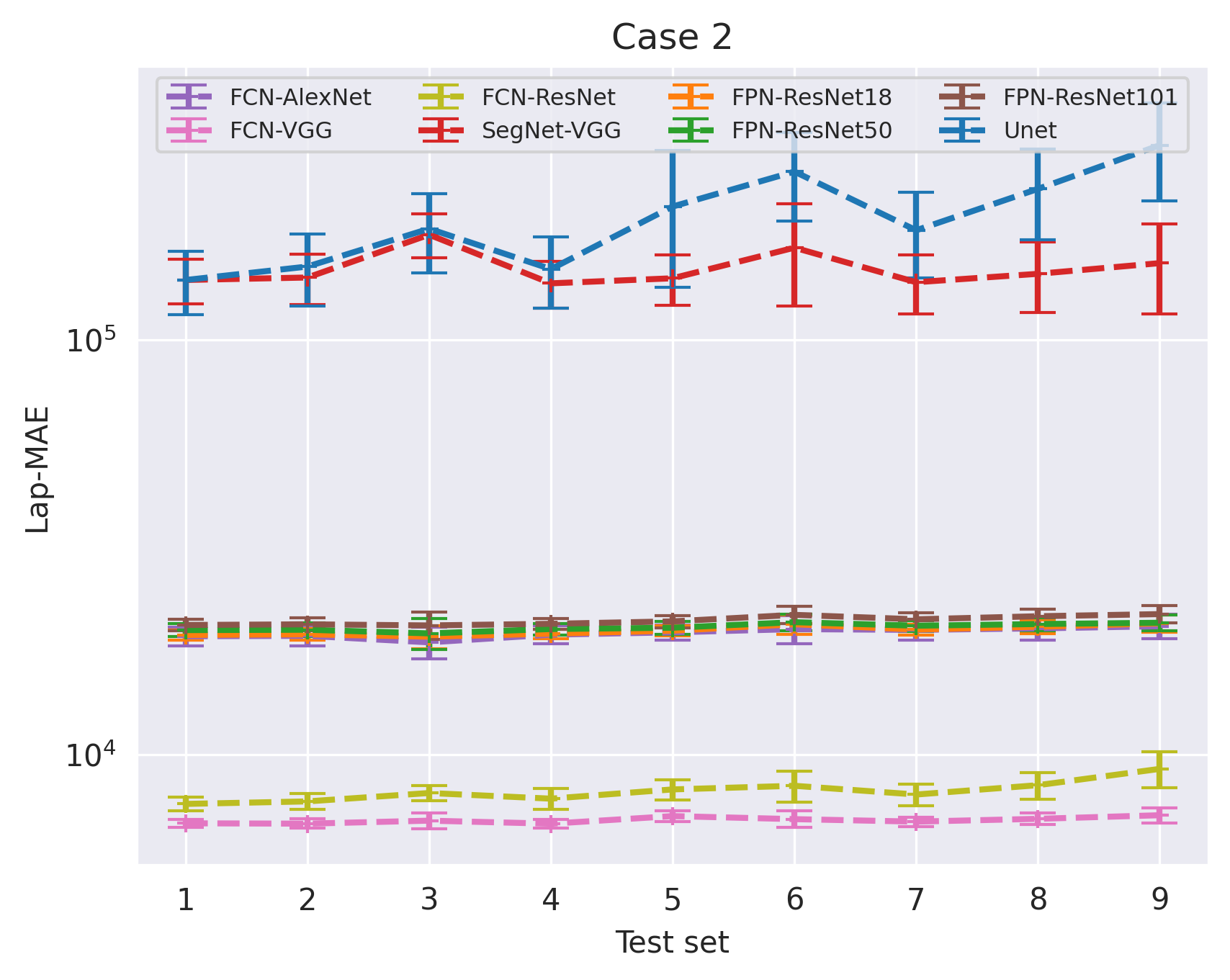}
		\label{fig:lapmae:case2}
	}
	\subfigure[Lap-MAE in Case 3]{
		\includegraphics[width=0.31\linewidth]{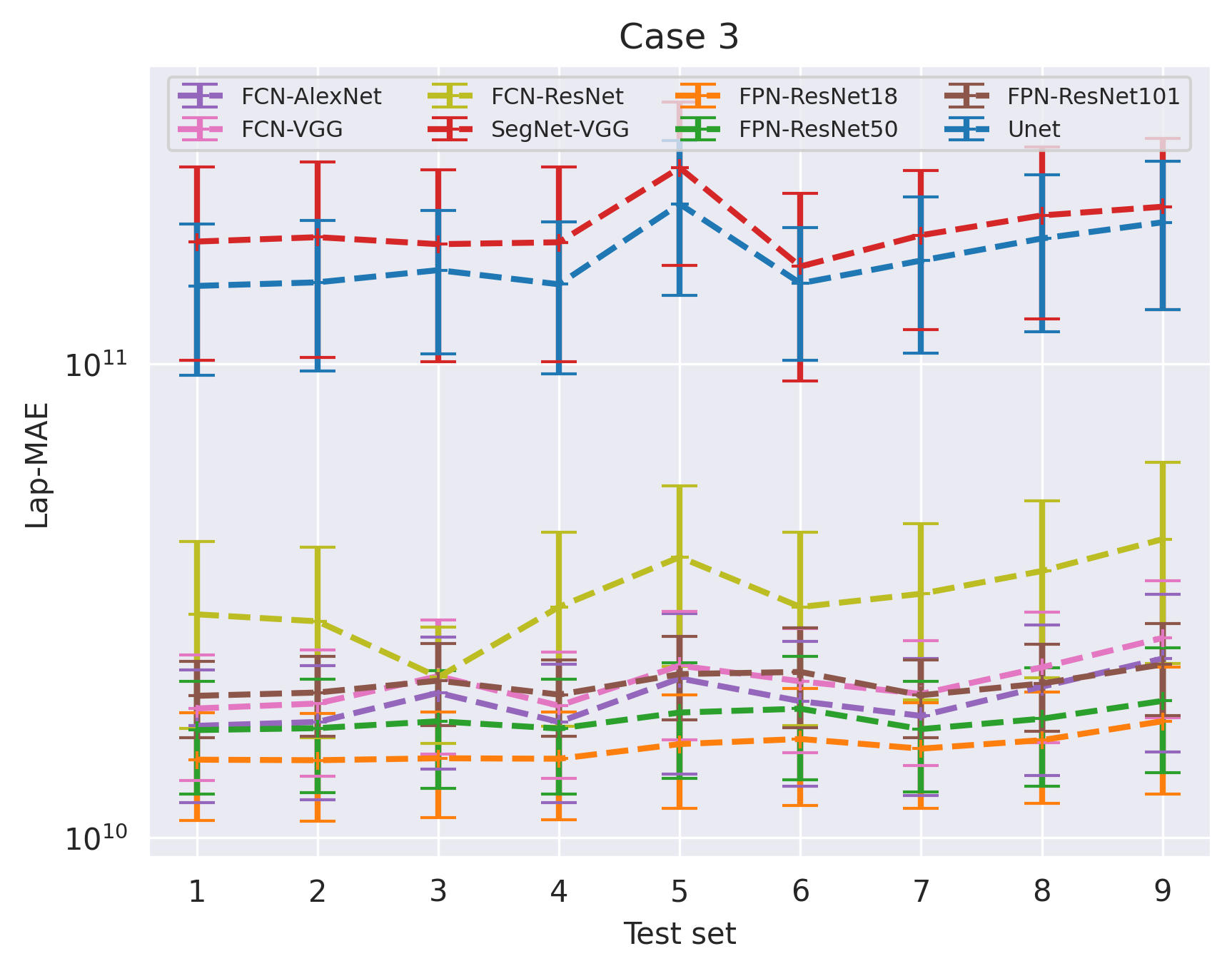}
		\label{fig:lapmae:case3}
	}
	\caption{The comparison of the prediction performance on the Lap-MAE in nine test sets of eight representative DNN surrogates in three cases. (a-c) Lap-MAE of surrogates in case 1, 2 and 3. (unit: W/m$^2$)}
	\label{fig:lapmae}
\end{figure*}

\begin{figure*}[!htbp]
	\centering
	\subfigure[$\rho_{MT}$ in Case 1]{
		\includegraphics[width=0.31\linewidth]{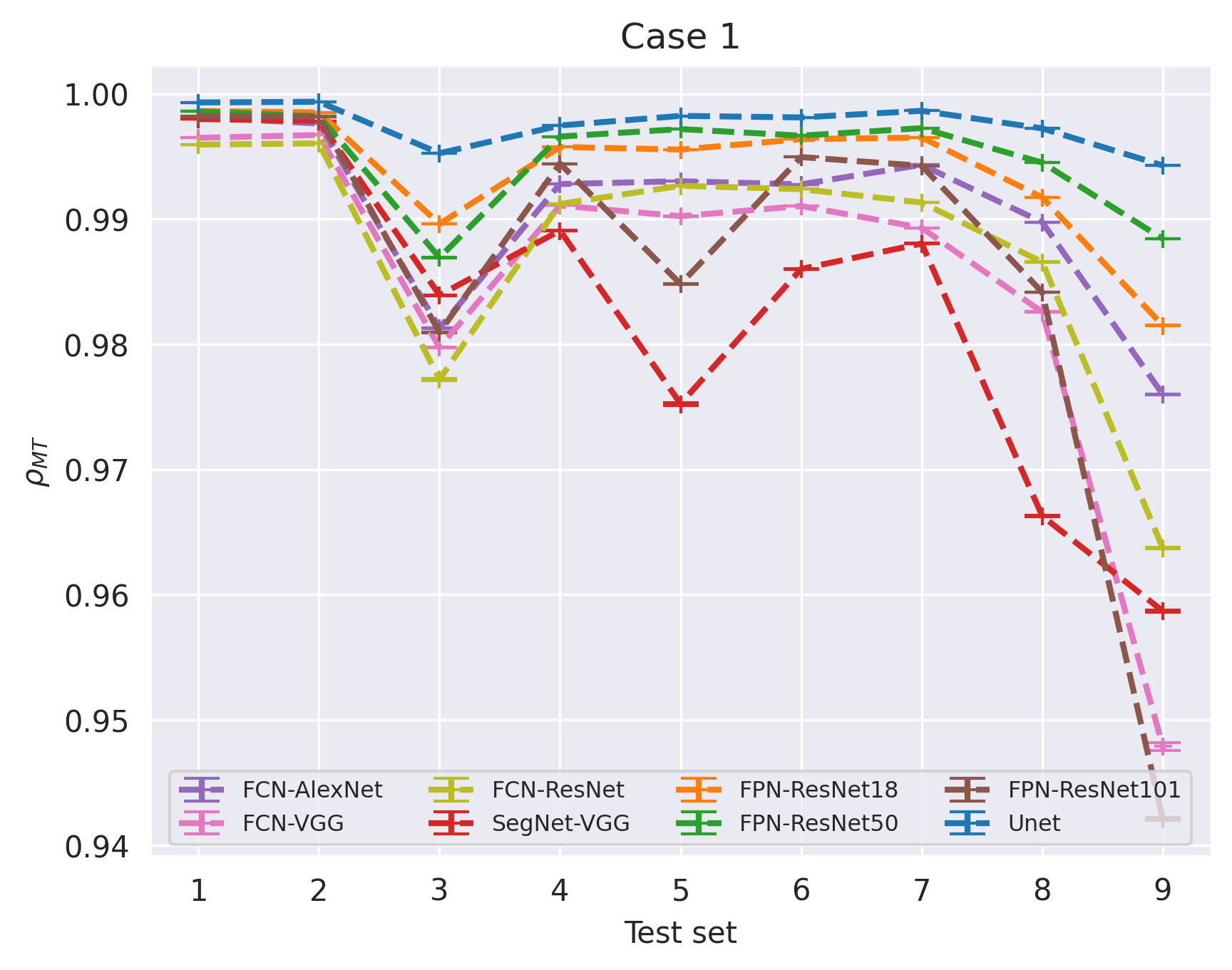}
		\label{fig:rhomae:case1}
	}
	\subfigure[$\rho_{MT}$ in Case 2]{
		\includegraphics[width=0.31\linewidth]{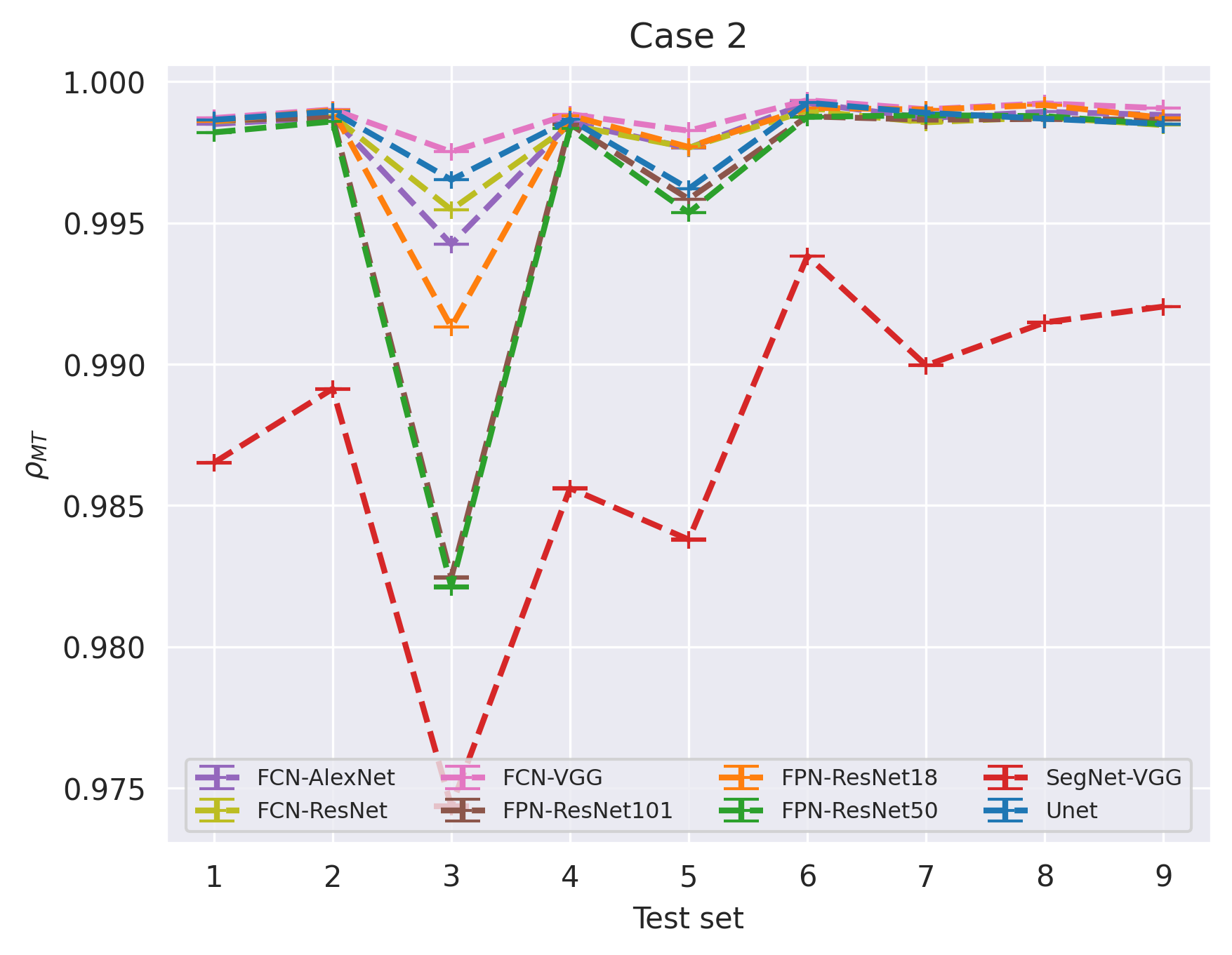}
		\label{fig:rhomae:case2}
	}
	\subfigure[$\rho_{MT}$ in Case 3]{
		\includegraphics[width=0.31\linewidth]{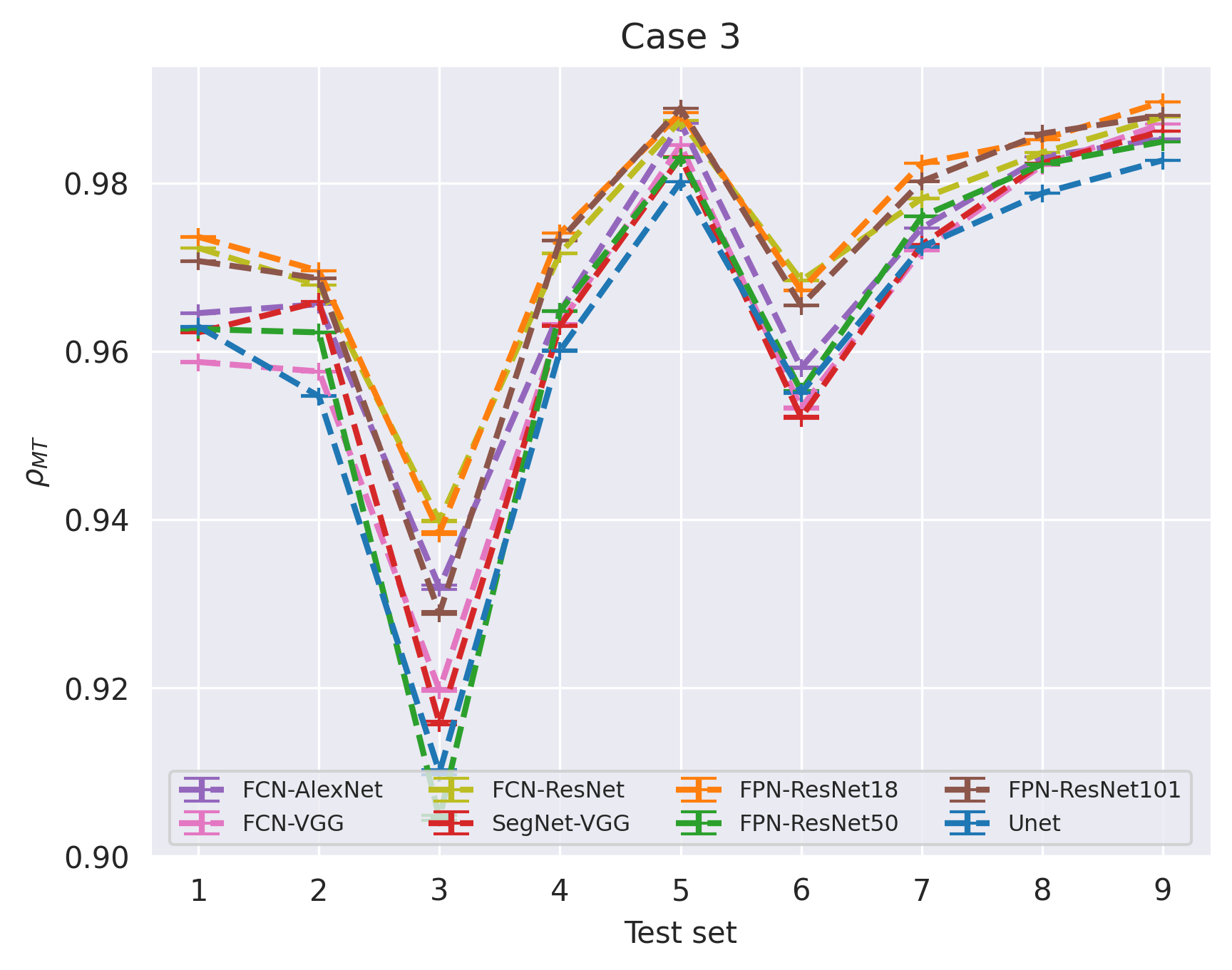}
		\label{fig:rhomae:case3}
	}
	\caption{The comparison of the Spearman coefficient $\rho_{MT}$ in nine test sets of eight representative DNN surrogates in three cases. (a-c) $\rho_{MT}$ of surrogates in case 1, 2 and 3.}
	\label{fig:rhomae}
\end{figure*}

\begin{figure*}[!htbp]
	\centering
	\subfigure[MAE in test set 6 of Case 2]{
		\includegraphics[width=0.31\linewidth]{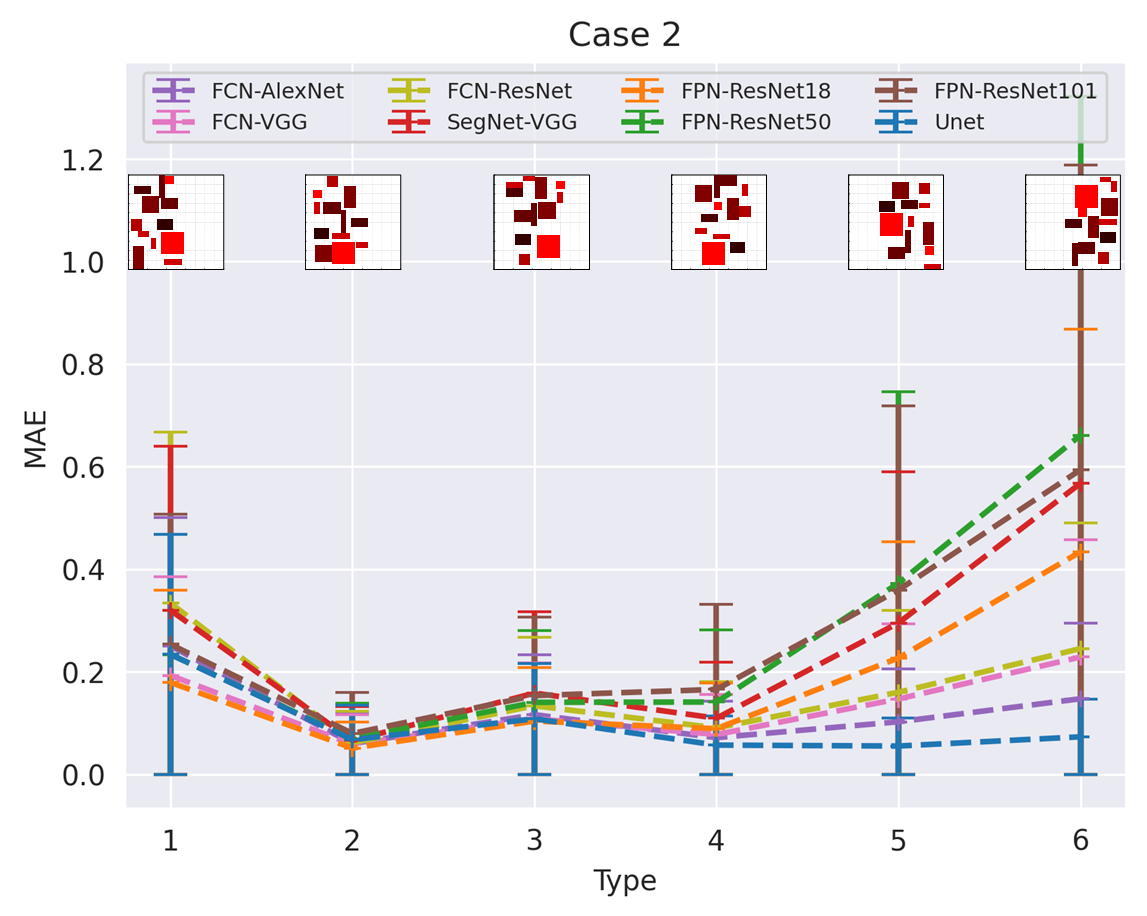}
		\label{fig:maeintest:case2}
	}
	\subfigure[MAE in test set 5 of Case 3]{
		\includegraphics[width=0.31\linewidth]{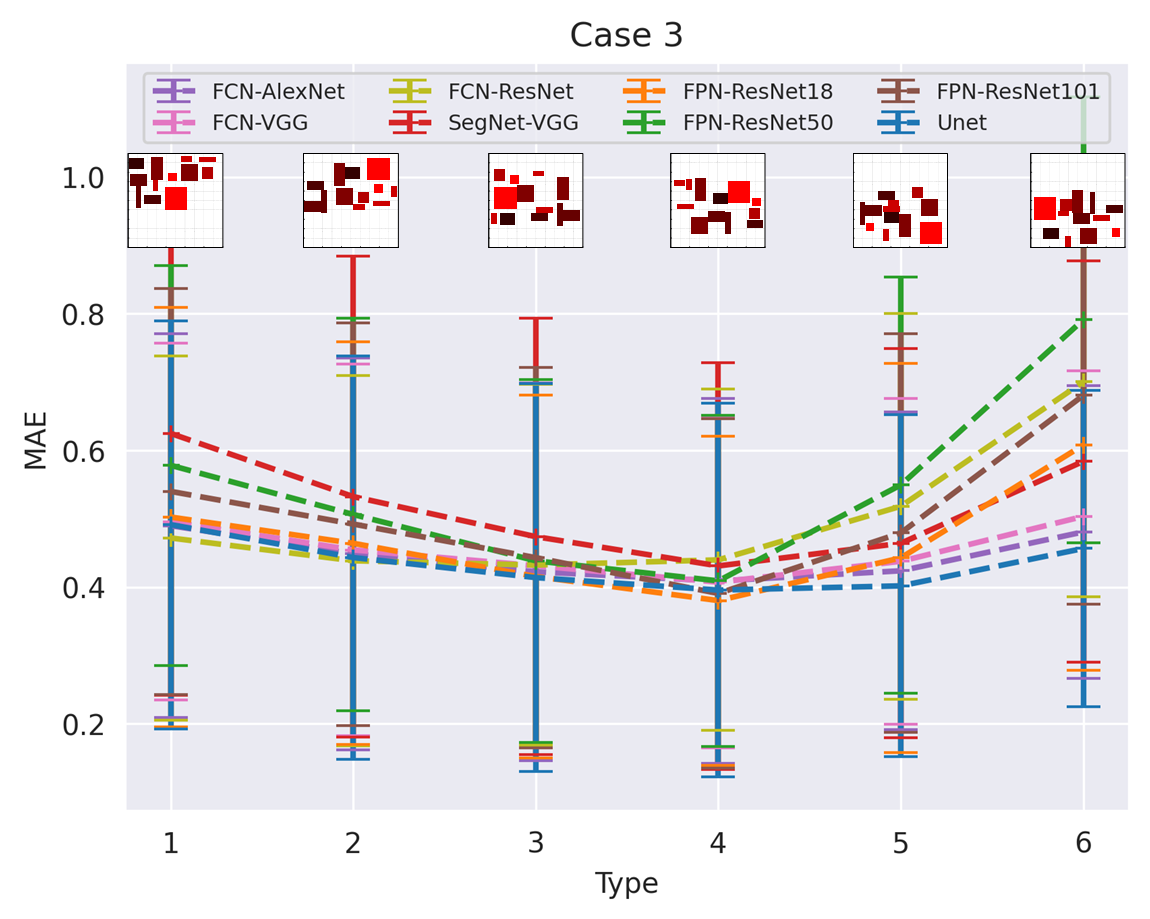}
		\label{fig:maeintest:case3}
	}
	\\
	\subfigure[Max AE in test set 6 of Case 2]{
		\includegraphics[width=0.31\linewidth]{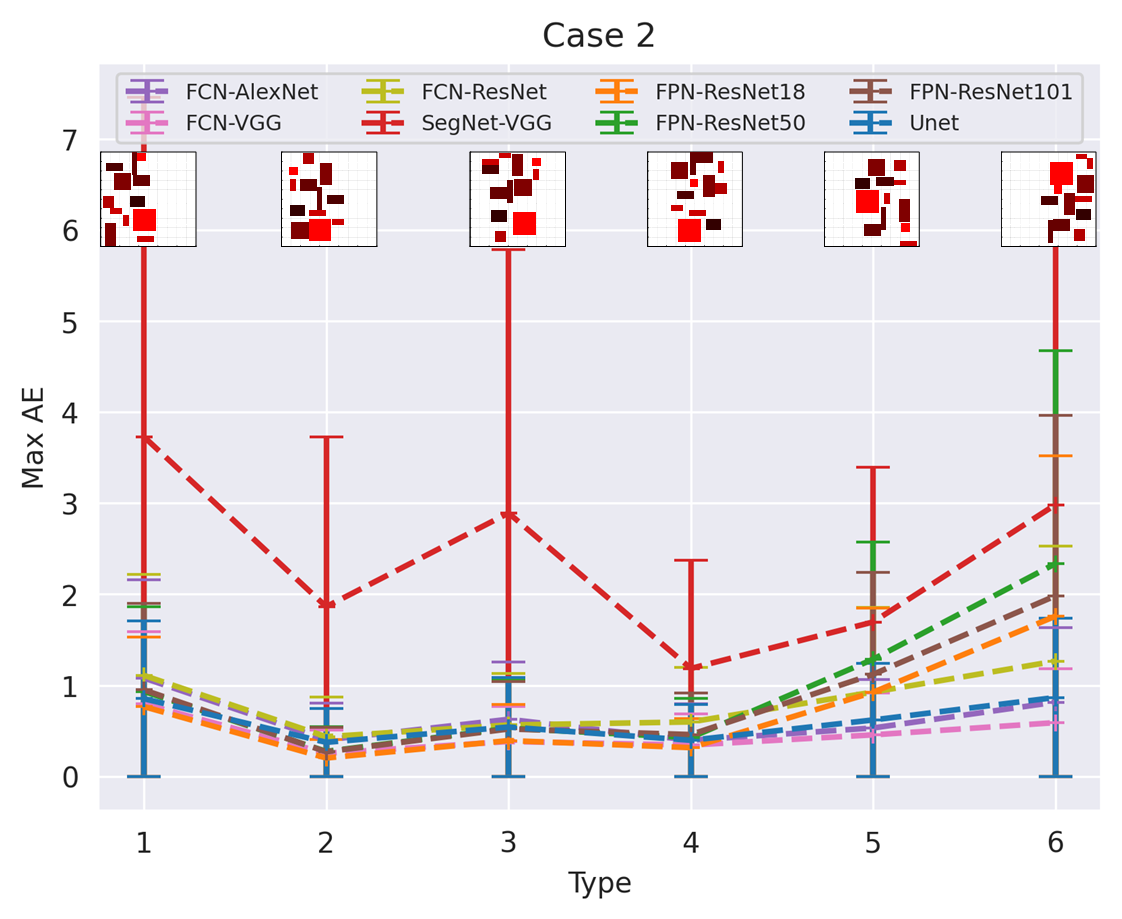}
		\label{fig:maxaeintest:case2}
	}
	\subfigure[Max AE in test set 5 of Case 2]{
		\includegraphics[width=0.31\linewidth]{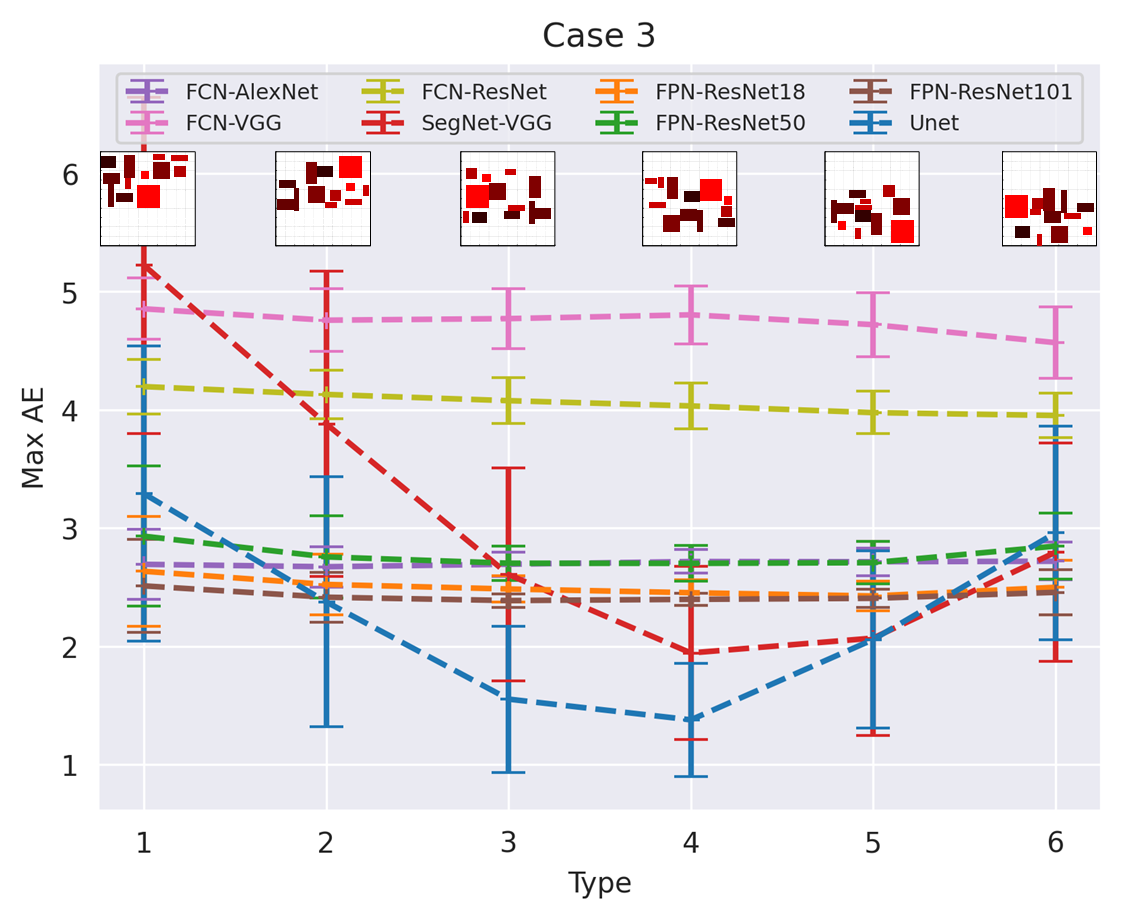}
		\label{fig:maxaeintest:case3}
	}
	\caption{The comparison of the prediction performance on different types of part-space test samples in test set 6 for case 2 and test 5 for case 3.}
	\label{fig:mae-maxae-intest65}
\end{figure*}

\subsubsection{Comparison and discussion of different DNN surrogates in different test sets and evaluation metrics}

Fig.~\ref{fig:mae-maxae}-\ref{fig:rhomae} illustrate the predictive performance of different DNN surrogates on three types of evaluation metrics, aiming to provide a comprehensive assessment on surrogates from different aspects.
The statistical results, including the mean and standard deviation value, in terms of one metric for different surrogates in nine test sets of one case, is displayed in one single image.
The resulting images with similar purposes are presented in one figure for better comparison.
In the following, we will analyze and discuss these results in detail from four aspects of problem settings, test sets, DNN surrogate models, and metrics, trying to present some insights for task data learning or uncover some useful instructions for surrogate modeling.



\smallskip
\noindent
\textbf{Analysis on three different cases.}
For each case, the predictive MAE and Max AE of surrogates in nine test sets are displayed in Fig.~\ref{fig:mae-maxae} to reveal the overall performance of different DNN models.
From the figure, it can be clearly seen that with feeding the same number of training samples, the surrogates in terms of both MAE and Max AE metrics for Case 3 present a more large prediction deviation than those for Case 1 and 2. 
Also, the MAE and Max AE for Case 2 are larger than those for Case 1.
This indicates that Case 3, which is the VP problem, is the most difficult task for surrogates to learn the inherent laws from the provided layout and temperature distribution data, and further Case 2 is harder than Case 1.
The reason that leads to this is that in the VP problem, the temperature distribution varies dramatically over the whole layout domain and the learned temperature data maintains a larger temperature rise and a more complex variation trend under the same layout scheme, significantly raising the complication of the image-to-image regression task.
By comparison, there are relatively more gentle temperature variations in two VB problems, which make the data-fitting task easier to be implemented.

\smallskip
\noindent
\textbf{Analysis on nine test sets.}
For each DNN surrogate, the prediction capability is evaluated in a relatively comprehensive test dataset. 
In case of the fact that the finite number of random sampling cannot prevent the imbalance of the generated data, in our paper, nine types of test sets are created heuristically based on some special considerations.
In spite of some fluctuations, a general trend in terms of test errors reveals the fact that the difficulty of test sets gradually increases from test set 1 to 9.

As shown in Fig.~\ref{fig:mae-maxae}, test errors in test set 1 and 2 are usually smaller than those in other test sets, which illustrates that the test samples generated by random layout sampling methods can be easily predicted by DNN surrogates as the training samples are obtained by one of the random layout sampling method, namely the SeqLS method.
Despite this, test set 2 is slightly more difficult than test set 1 since the Max AE of test set 2 is larger than that of test set 1 in all three cases.
It is also understandable because there is a slight difference in the sample distribution between test set 2 which consists of the GibLS data and the training set which consists of the SeqLS data.

In addition, it can be observed from Fig.~\ref{fig:mae-maxae} that test set 3 composed of corner samples is more challenging in Case 2 and 3.
It is because of the fact that when the component with the maximum heat intensity is placed around the corner which is far away from the heat sink, the temperature field of this layout might experience a higher rise, which has not been seen by the neural networks, resulting in worse prediction performance.
Although there is no much degradation in test set 3 for case 1 in MAE and Max AE metrics, the prediction on the position where the maximum temperature happens has displayed a huge standard deviation than the other test sets by seeing Fig.~\ref{fig:mtpae:case1}.
This phenomenon reminds us that the assessment of the surrogates should be conducted from many sides, as well as the determination of loss function for training the neural networks.
As for test set 4 containing group samples, only worse performance in Case 1 is observed, which illustrates that group samples can make a pretty larger effect on the surrogate modeling task for the VB problem with identical isothermal boundaries.

Test set 5-9 are made up of part-space samples, which simulate the clustering scenarios of components during layout optimization.
In test set 5 and 6, components are restricted in half of the domain and the difference is the position of the predetermined half-space.
From the results in Fig.~\ref{fig:mae:case2}-\ref{fig:lapmae:case2}, it is clearly demonstrated that surrogates make a worse prediction in test set 6 than test set 5 in Case 2.
However, the phenomenon in Case 3 is reversed according to Fig.~\ref{fig:mae:case3}-\ref{fig:lapmae:case3}.
This induced difference is closely related to the heat dissipation process in different problem settings, specifically the location of isothermal boundary in Case 2 or heat sink in Case 3.
To verify this point, we further draw the predicted results on different types of test samples in test set 6 for case 2 and test set 5 for case 3, which is shown in Fig.~\ref{fig:mae-maxae-intest65}.
In Fig.~\ref{fig:maeintest:case2}-\ref{fig:maxaeintest:case3}, we can find that when the components are close to the heat dissipation boundary or far away from this boundary, the prediction errors of surrogates increase significantly.
Notice that the heat dissipation boundary is the left side of the domain as demonstrated in Fig.~\ref{fig:case2}.
The layout scenarios where components are close to the heat dissipation boundary will lead to a pretty smaller rise of temperature distribution while those where components are far away from the boundary will produce a pretty larger rise of temperature distribution, which can be regarded as extreme situations.
Due to the fact that the SeqLS method is less likely to generate such extreme samples included in the training set, the prediction on these test samples can be regarded as out-of-distribution inference of neural networks, thus causing more inaccurate predictions. 
The same rule can be applied to analyze the drop of inference performance in the test set 5 for Case 3.
Note that the tiny heat sink in Case 3 is located in the middle of the downside of the domain.

In terms of test set 7, 8 and 9, components gather more and more closely in the provided test layout samples, which causes a relatively larger temperature variation and thus increases the difficulty of the test sets.
The analysis is consistent with the predictions shown in Fig.~\ref{fig:mae-maxae}-\ref{fig:lapmae}.

In a word, the insight of training an accurate enough surrogate model by investigating such test sets is to prepare a high-quality training set by providing as diversified training samples as possible in not only the layout design space but also the temperature response space.

\smallskip
\noindent
\textbf{Discussion on different DNN surrogate models.}
By observing Fig.~\ref{fig:mae-maxae}, it can be concluded that generally, Unet outperforms most of the other DNN surrogates in terms of MAE and Max AE metrics across three cases.
In some cases, FPN-ResNet18 and FCN-VGG have presented comparable prediction performance with Unet.
On the contrary, the models with the worst prediction performance are SegNet-AlexNet and SegNet-ResNet, which have extremely large Max AE in three cases.
It demonstrates that these two models can hardly deal with the ultra-high dimensional surrogate modeling task even for the simplest problem of Case 1.
Therefore, in order to better visualize the test performance results, these two worst models are removed from the figures below and only eight comparable DNN surrogates are involved.

Another interesting finding is that the prediction performance of surrogates in Case 3 presents a pretty much larger standard deviation in all test sets than those for Case 1 and 2 although the global MAE of Case 3 is only slightly worse by around 0.2K.
This instability of inference capability illustrates not only the problem complexity but also the insufficiency of 2000 training samples for obtaining an accurate enough surrogate for Case 3.

From Fig.~\ref{fig:mae-maxae}, it can be found that there is no significant difference in the testing MAE between different DNN surrogates, which illustrates two things.
One is that different backbone networks for extracting features will lead to similar prediction performance and hence the choice of this backbone network should be determined according to other indicators, such as the number of parameters or inference latency.
This observation differs from the situation in the image classification or segmentation task where ResNet backbone is slightly better than VGG an AlexNet.
The other is that different deep learning architectures including FCN, FPN, SegNet and Unet, cannot bring in huge fundamental improvements in the generalization capability of surrogates, especially between FCN and FPN.
However, it should be worthy of noticing that Unet has a superior prediction performance than SegNet though they share a similar network architecture with encoder and decoder embedded.
The reason why this happens is that features from different levels are combined by skip connections in Unet, which is not included in SegNet, thus being helpful to the regression task and enabling a better inference capability of Unet.
In addition to considering the multi-level or multi-scale feature fusion in networks, in order to build superior neural network-based surrogate models, more efforts should be paid on other aspects simultaneously, such as the design of the training loss function.

\smallskip
\noindent
\textbf{Discussion on different evaluation metrics.}
Apart from MAE and Max AE discussed above, some other metrics should be put more emphasis on in order to investigate more aspects to analyze the characteristic for better surrogate modeling.

One important variable in layout optimization is the maximum temperature in the whole field.
To account for this local prediction capability, the statistics about MT-AE and MT-PAE are plotted in Fig.~\ref{fig:mtae-mtpae}.
From this figure, a similar trend of MT-AE in inference performance with MAE can be observed, which is understandable.
Regardless of this, it should be noticed that SegNet-VGG maintains a rather larger error and standard deviation in MT-AE of Case 1 and 2, indicating an inaccurate prediction on the maximum temperature by SegNet-VGG.
Besides, the huge standard deviation on MT-PAE as shown in Fig.~\ref{fig:mtpae:case1}-\ref{fig:mtpae:case3} reflects the unstable prediction performance for almost all of DNN surrogates, especially for Case 2 and 3.
If it is required to accurately predict the maximum temperature in not only its value but also its position, the inconsistency problem should be paid more attention to.

Another two important metrics about the prediction on the boundary temperature are illustrated in Fig.~\ref{fig:bmaed-bmaen}.
Note that there is no Neumann boundary involved in Case 1, thus displaying zero-value results in Fig.~\ref{fig:bmaen:case1} for better demonstration.
Comparing Fig~\ref{fig:bmaed:case1}-\ref{fig:bmaed:case3} with Fig.~\ref{fig:bmaen:case1}-\ref{fig:bmaen:case3}, it can be obviously seen that BMAE$_N$ is rather larger than BMAE$_D$ in Case 2 and 3, which illustrates the different difficulty of surrogates fitting the boundary temperature data.
This is due to the fact that the temperature on the Dirichlet boundary is of constant value while the laws behind the data involved in the Neumann boundary is the constant first-order normal gradient value.
For neural networks, it is easier to learn the direct constant value rather than the constant first-order gradient implicitly involved in the data.
This can also be used to explain why the first-order and second-order gradient errors as shown in Fig.~\ref{fig:gmae} and \ref{fig:lapmae} are significantly larger than MAE.
In addition, in Fig.~\ref{fig:bmaed-bmaen}(a)-(c), some regular pattern is demonstrated in terms of BMAE$_D$ that Unet and SegNet maintain the smallest error in all cases and FPN surrogates (FPN-ResNet18, FPN-ResNet50 and FPN-ResNet101) always keep nearly the same prediction error in Dirichlet boundary.
The reason is that in Unet and SegNet, the 200$\times$200 temperature matrix is directly learned from the provided labeled data while in FPN and FCN, the temperature matrix is recovered from the feature map of the smaller size.
Specifically, the predicted temperature matrix is directly obtained from the learned 25$\times$25 feature map in FCN and 50$\times$50 in FPN by the bilinear interpolation-based upsampling technique, resulting in such a performance gap in fitting the constant value.

Besides, the CMAE metric, which evaluates the maximum component temperature inference error, is investigated as shown in Fig.~\ref{fig:cmae}.
Even if the general trend in different test sets of CMAE is similar to global MAE, the value of errors is more larger than global MAE.
It indicates that some component temperatures cannot be predicted as accurately as the average level.
Therefore, it would be a serious situation that should be taken care of if certain special component, which might be sensitive to the temperature, is required to be predicted with high accuracy.

Fig.~\ref{fig:gmae} and \ref{fig:lapmae} illustrate the capability of different regression surrogates learning the inherent first-order and second-order gradient laws.
From these two figures, it can be apparently observed that FCN and FPN outperform Unet and SegNet in terms of these two metrics in the absence of perfect fitting. 
Theoretically and intuitively, if the model can perfectly perform the prediction for layout samples, the predicted temperature distribution as well as its gradient field will infinitely approach the labeled one, which is demonstrated by seeing Unet model for Case 1 in Fig.~\ref{fig:gmae:case1}. 
Otherwise, the network architecture will have a stronger impact on its gradient errors.
As stated in some previous paragraphs, the 8 times and 4 times linear interpolation in FCN and FPN lead to the apparent performance advantage compared with Unet and SegNet in terms of Lap-MAE as shown in Fig.~\ref{fig:lapmae}.

The Spearman coefficient $\rho_{MT}$ is presented in Fig.~\ref{fig:rhomae}, which reveals the rank-order capability of surrogates in assessing different layout samples by its maximum temperature.
Unet undoubtedly performs best in Case 1 while in Case 2 and 3, FPN-ResNet18 and FCN-VGG show comparable ability.
The fluctuation in different test sets reflects the unstable prediction performance of different DNN surrogates.
Hence, in order to guarantee the surrogate-assisted layout optimization search towards the right direction, the surrogate should keep a consistent rank-order capability for rightly selecting better layout samples in the entire layout design space. 

\subsubsection{Investigation on the model efficiency}

\begin{table*}[!htbp]
	\centering
	\caption{The results of the number of parameters and the computing latency in different hardware for different DNN surrogates.}
	\label{table:efficiency}
	\newcommand{\tabincell}[2]{\begin{tabular}{@{}#1@{}}#2\end{tabular}}  
	\begin{tabular}{l c c c c}
		\toprule
		Model & $\#$Params & \tabincell{c}{CPU Lat.\\(ms)} &\tabincell{c}{GPU Lat.\\(batch=1, ms)} &\tabincell{c}{GPU Lat.\\(batch=64, ms)} \\
		\hline
		FCN-AlexNet    & 5.3M  & 10.8  & 1.4  & 0.25 \\
		FCN-VGG        & 18.8M & 54.4  & 2.2  & 1.14 \\
		FCN-ResNet     & 15.2M & 19.6  & 3.3  & 0.40 \\
		SegNet-AlexNet & 4.9M  & 21.6  & 2.3  & 0.39 \\
		SegNet-VGG     & 29.5M & 76.2  & 4.5  & 2.26 \\
		SegNet-ResNet  & 20.1M & 141.1 & 5.2  & 2.36 \\
		FPN-ResNet18   & 13.1M & 24.2  & 4.1  & 0.80 \\
		FPN-ResNet50   & 26.1M & 50.2  & 7.7  & 1.37 \\
		FPN-ResNet101  & 45.1M & 76.8  & 13.3 & 1.87 \\
		Unet           & 31.0M   & 134.5 & 6.0    & 2.86 \\
		\bottomrule
	\end{tabular}
\end{table*}

To investigate the model efficiency, the number of parameters of neural networks and the computing latency in different hardware are reported in Table~\ref{table:efficiency}.
Note that the unit (M) in the second column stands for million, which means that there are 31.0 million parameters in Unet.
Run times are important but hard to compare as they can vary a lot in different machines.
We take a practical view and report the mean value of the time it takes after evaluating the test set of 1000 instances, either on a CPU system of Intel Xeon Gold 6242R or on a single GPU of Nvidia GTX3090.
It should be noticed that GPU allows parallel processing by simultaneously inputting a batch of examples.
Thus, we further compute the mean run time for one inference when setting the batch size as 64, as listed in the last column of Table \ref{table:efficiency}.
From this table, it can be clearly observed that the computing latency of DNN surrogates on either CPU or GPU keeps no more than 1s even in FPN-ResNet101 with the maximum number of parameters 45.1 million, which firmly illustrates the advantage in high computing efficiency of neural network-based surrogates.
Unsurprisingly, the inference time on GPU can be significantly reduced compared with that on CPU by at least one order of magnitude.

Following the analysis in the previous subsection, it can be known that Unet, which is with the best prediction performance, has an almost maximum CPU latency of 134.5ms or GPU latency of 2.86ms (batch=64). 
By comparison, another two competitive DNN surrogates, FPN-ResNet18 and FCN-VGG, maintain a smaller number of parameters and shorter computing latency.
When choosing the proper model as the surrogate, the model efficiency and the prediction performance should be balanced with a thorough consideration.

\subsection{Instructions on surrogate modeling}

From the above experimental results, some interesting and meaningful insights are summarized as follows.

\begin{enumerate}
	\item In most cases, GN, namely group normalization, is more suitable for the HSL-TFP task compared with batch normalization, which can accelerate the convergence and improve the training efficacy.
	\item The first choice of model architecture for this image-to-image regression task can be Unet, then FPN and FCN in terms of the prediction performance. However, when taking the model efficiency into consideration, FPN-RestNet18 and FCN-VGG can be tried first to implement the surrogate modeling task.
	\item Experimental results show that Case 3 is more challenging than Case 1 and 2. To build more accurate surrogates, more efforts can be made on preparing data with sufficient amount and diversity. Note that the diversity should be considered not only in the layout design space but also in the temperature response space.
	\item Besides, more efficient and effective loss functions are desired to guide the neural network to converge with better performance. If the local predictive capability on some interesting area is important, one possible approach is to integrate this information into the loss function in the proper manner.
	\item When there is no guarantee that models can be tested over the entire layout space, one useful method is to heuristically construct relatively comprehensive test sets by special layout sampling strategies, which are proven to be effective for an overall evaluation.
	\item The global MAE cannot completely measure the performance of surrogates. Given that, the proposed various metrics provide a multi-view investigation on them. More useful metrics are encouraged to be developed to precisely and comprehensively reveal the inherent characteristic behind models.
	\item The results about the gradient-based metrics remind us that the upsampling based on quadratic interpolation which is embedded in network architecture might be helpful to enhance the model performance in our HSL-TFP task.
\end{enumerate}

The above observations can provide the researchers some meaningful instructions to explore more effective DNN models for this challenging benchmark, promoting the progress in the HSL-TFP task and helping the surrogate-assisted heat source layout optimization mission.

\section{Conclusion and future prospects}
\label{sec:conclusions}

In this paper, to deal with the heat source layout optimization driven by the thermal performance, the HSL-TFP task is first summarized to construct computationally cheap but accurate surrogates instead of using the expensive numerical simulation for evaluating the temperature field. 
To solve this ultra-high dimensional nonlinear regression mission, the DNN surrogate modeling benchmark on three typical heat conduction problems is proposed.
Our previous work \citep{chen2020} has demonstrated the feasibility of using the deep learning surrogate to predict temperature distribution for varying heat source layout schemes.
However, only identical square components were considered.
By comparison, a more complex layout task involving components with various sizes, shapes and intensities is constructed in this work, making it more practical and simultaneously difficult.
To implement the data acquisition for this general problem, two random layout sampling methods and some special layout sampling strategies are proposed, thus producing a relatively comprehensive heat source layout dataset, i.e., HSLD.
By combining FCN, SegNet, FPN and Unet with AlexNet, VGG and ResNet backbone, ten representative DNN models are developed as baseline surrogates.
Three types of evaluation metrics provide a multi-view investigation on the prediction performance or the generalization capability of different DNN surrogates.
Experimental studies illustrate not only some useful instructions in the task characteristic, data preparation and neural architecture selection but also the disadvantages of current DNN surrogates in training efficacy and model construction.
These can serve as the baseline results to advance the state-of-the-art surrogate modeling methods in future research.

Future research based on this benchmark can be conducted from three aspects including data preparation, surrogate modeling methods, and model evaluation.
First, 
it would be interesting to investigate other strategies, such as active learning, to generate other training sets with high diversity and fewer number of training samples in the data preparation process. 
Then, we advise to consider constructing other effective training losses, designing adaptive deep surrogate architectures for surrogate modeling.
Finally, developing other meaningful evaluation metrics which can satisfy the specific requirements in engineering would be another interesting direction. 
All these efforts will contribute to the final heat source layout design method that can be applied in engineering, such as satellite system design.
We hope our benchmark can attract more inter-disciplinary scholars to promote this research based on their valuable talent.


%
\section*{Conflict of interest}

The authors declare that they have no conflict of interest.

\section*{Replication of results}

The codes for reproducing this supervised DNN surrogate modeling benchmark are released at the project Web page: \url{https://github.com/idrl-lab/supervised_layout_benchmark}.


\begin{acknowledgements}

This work was supported in part by National Natural Science Foundation of China under Grant No.11725211, 52005505, and 62001502, and Post-graduate Scientific Research Innovation Project of Hunan Province under Grant No.CX20200023.

\end{acknowledgements}

\bibliographystyle{spbasic}
\bibliography{references}

\begin{thebibliography}{45}
\providecommand{\natexlab}[1]{#1}
\providecommand{\url}[1]{{#1}}
\providecommand{\urlprefix}{URL }
\expandafter\ifx\csname urlstyle\endcsname\relax
  \providecommand{\doi}[1]{DOI~\discretionary{}{}{}#1}\else
  \providecommand{\doi}{DOI~\discretionary{}{}{}\begingroup
  \urlstyle{rm}\Url}\fi
\providecommand{\eprint}[2][]{\url{#2}}

\bibitem[{Arshad et~al.(2020)Arshad, Jabbal, Sardari, Bashir, Faraji, and
  Yan}]{arshad2020}
Arshad A, Jabbal M, Sardari PT, Bashir MA, Faraji H, Yan Y (2020) Transient
  simulation of finned heat sinks embedded with pcm for electronics cooling.
  Thermal Science and Engineering Progress 18:100520

\bibitem[{Aslan et~al.(2018)Aslan, Puskely, and Yarovoy}]{aslan2018}
Aslan Y, Puskely J, Yarovoy A (2018) Heat source layout optimization for
  two-dimensional heat conduction using iterative reweighted l1-norm convex
  minimization. International journal of heat and mass transfer 122:432--441

\bibitem[{Badrinarayanan et~al.(2017)Badrinarayanan, Kendall, and
  Cipolla}]{badrinarayanan2017segnet}
Badrinarayanan V, Kendall A, Cipolla R (2017) Segnet: A deep convolutional
  encoder-decoder architecture for image segmentation. IEEE transactions on
  pattern analysis and machine intelligence 39(12):2481--2495

\bibitem[{Ballester and Araujo(2016)}]{ballester2016}
Ballester P, Araujo RM (2016) On the performance of googlenet and alexnet
  applied to sketches. In: AAAI Conference on Artificial Intelligence, pp
  1124--1128

\bibitem[{Bennell et~al.(2000)Bennell, Dowsland, and Dowsland}]{Bennell2000a}
Bennell JA, Dowsland KA, Dowsland WB (2000) {The irregular cutting-stock
  problem - a new procedure for deriving the no-fit polygon}. Computers and
  Operations Research 28(3):271--287, \doi{10.1016/S0305-0548(00)00021-6}

\bibitem[{Bishop(2006)}]{2006Pattern}
Bishop CM (2006) Pattern Recognition and Machine Learning (Information Science
  and Statistics). Springer-Verlag New York, Inc.

\bibitem[{Chen et~al.(2016{\natexlab{a}})Chen, Wang, and Song}]{chen2016b}
Chen K, Wang S, Song M (2016{\natexlab{a}}) Optimization of heat source
  distribution for two-dimensional heat conduction using bionic method.
  International Journal of Heat and Mass Transfer 93:108--117

\bibitem[{Chen et~al.(2016{\natexlab{b}})Chen, Wang, and Song}]{chen2016a}
Chen K, Wang S, Song M (2016{\natexlab{b}}) Temperature-gradient-aware bionic
  optimization method for heat source distribution in heat conduction.
  International Journal of Heat and Mass Transfer 100:737--746

\bibitem[{Chen et~al.(2018)Chen, Yao, Zhao, Chen, and Zheng}]{chen2018}
Chen X, Yao W, Zhao Y, Chen X, Zheng X (2018) A practical satellite layout
  optimization design approach based on enhanced finite-circle method.
  Structural and Multidisciplinary Optimization 58(6):2635--2653

\bibitem[{Chen et~al.(2020)Chen, Chen, Zhou, Zhang, and Yao}]{chen2020}
Chen X, Chen X, Zhou W, Zhang J, Yao W (2020) {The heat source layout
  optimization using deep learning surrogate modeling}. Structural and
  Multidisciplinary Optimization 62(6):3127--3148,
  \doi{10.1007/s00158-020-02659-4}

\bibitem[{Chen et~al.(2021)Chen, Yao, Zhao, Chen, and Liu}]{Chen2021}
Chen X, Yao W, Zhao Y, Chen X, Liu W (2021) {A novel satellite layout
  optimization design method based on phi-function}. Acta Astronautica
  180:560--574, \doi{10.1016/j.actaastro.2020.12.034}

\bibitem[{Cheng et~al.(2009)Cheng, Xu, and Liang}]{cheng2009}
Cheng K, Xu X, Liang XG (2009) Homogenization of temperature field and
  temperature gradient field. Science in China Series E: Technological Sciences
  52(10):2937--2942

\bibitem[{Chernov et~al.(2010)Chernov, Stoyan, and Romanova}]{Chernov2010}
Chernov N, Stoyan Y, Romanova T (2010) {Mathematical model and efficient
  algorithms for object packing problem}. Computational Geometry
  43(5):535--553, \doi{10.1016/j.comgeo.2009.12.003}

\bibitem[{Ciccazzo et~al.(2014)Ciccazzo, Pillo, and Latorre}]{ciccazzo2014}
Ciccazzo A, Pillo GD, Latorre V (2014) Support vector machines for surrogate
  modeling of electronic circuits. Neural Computing and Applications
  24(1):69--76

\bibitem[{Cuco et~al.(2015)Cuco, D., and Neto}]{cuco2015}
Cuco A, D SFL, Neto AS (2015) A multiobjective methodology for spacecraft
  equipment layouts. Optimization and Engineering 16(1):165--181

\bibitem[{Emam et~al.(2019)Emam, Ookawara, and Ahmed}]{emam2019}
Emam M, Ookawara S, Ahmed M (2019) Thermal management of electronic devices and
  concentrator photovoltaic systems using phase change material heat sinks:
  Experimental investigations. Renewable energy 141:322--339

\bibitem[{Forrester and Keane(2009)}]{forrester2009}
Forrester AIJ, Keane AJ (2009) Recent advances in surrogate-based optimization.
  Progress in Aerospace Sciences 45:50--79

\bibitem[{G\'eczy(2017)}]{geczy2017}
G\'eczy A (2017) Investigating heat transfer coefficient differences on printed
  circuit boards during vapour phase reflow soldering. International Journal of
  Heat and Mass Transfer 109:167--174

\bibitem[{Goel et~al.(2007)Goel, Haftka, Shyy, and Queipo}]{goel2007}
Goel T, Haftka RT, Shyy W, Queipo NV (2007) Ensemble of surrogates. Structural
  and Multidisciplinary Optimization 33(3):199--216

\bibitem[{Gong et~al.(2019)Gong, Zhong, Yu, and Li}]{gong2019}
Gong Z, Zhong P, Yu Y, Li S (2019) A cnn with multiscale convolution and
  diversified metric for hyperspectral image classification. IEEE Transactions
  on Geoscience and Remote Sensing 57(6):3599--3618

\bibitem[{Gu et~al.(2017)Gu, Wang, Chen, Zhang, and He}]{gu2017}
Gu Y, Wang L, Chen W, Zhang C, He X (2017) Application of the meshless
  generalized finite difference method to inverse heat source problems.
  International Journal of Heat and Mass Transfer 108:721--729

\bibitem[{He et~al.(2016)He, Zhang, Ren, and Sun}]{he2016deep}
He K, Zhang X, Ren S, Sun J (2016) Deep residual learning for image
  recognition. In: Proceedings of the IEEE conference on computer vision and
  pattern recognition, pp 770--778

\bibitem[{Huang et~al.(2017)Huang, Liu, Maaten, and Weinberger}]{huang2017}
Huang G, Liu Z, Maaten LVD, Weinberger KQ (2017) Densely connected
  convolutional networks. In: IEEE conference on computer vision and pattern
  recognition, pp 4700--4708

\bibitem[{Hughes(2012)}]{hughes2015}
Hughes TJR (2012) The finite element method: linear static and dynamic finite
  element analysis. Courier Corporation

\bibitem[{Ioffe and Szegedy(2015)}]{ioffe2015batch}
Ioffe S, Szegedy C (2015) Batch normalization: Accelerating deep network
  training by reducing internal covariate shift. arXiv preprint arXiv:150203167

\bibitem[{Kokoska and Zwillinger(2000)}]{kokoska2000crc}
Kokoska S, Zwillinger D (2000) CRC standard probability and statistics tables
  and formulae. Crc Press

\bibitem[{Krizhevsky et~al.(2012)Krizhevsky, Sutskever, and
  Hinton}]{2012ImageNet}
Krizhevsky A, Sutskever I, Hinton G (2012) Imagenet classification with deep
  convolutional neural networks. In: NIPS

\bibitem[{Lackey et~al.(2019)Lackey, P\"urrer, Taracchini, and
  Marsat}]{lackey2019}
Lackey BD, P\"urrer M, Taracchini A, Marsat S (2019) Surrogate model for an
  aligned-spin effective-onebody waveform model of binary neutron star
  inspirals using gaussian process regression. Physical Review D 100(2):024002

\bibitem[{Lin et~al.(2017)Lin, Doll{\'a}r, Girshick, He, Hariharan, and
  Belongie}]{lin2017feature}
Lin TY, Doll{\'a}r P, Girshick R, He K, Hariharan B, Belongie S (2017) Feature
  pyramid networks for object detection. In: Proceedings of the IEEE conference
  on computer vision and pattern recognition, pp 2117--2125

\bibitem[{Long et~al.(2015)Long, Shelhamer, and Darrell}]{2015Fully}
Long J, Shelhamer E, Darrell T (2015) Fully convolutional networks for semantic
  segmentation. IEEE Transactions on Pattern Analysis and Machine Intelligence
  39(4):640--651

\bibitem[{Martinez-Maradiaga et~al.(2019)Martinez-Maradiaga, Damonte, Manzo,
  Haertel, and Engelbrecht}]{martinez2019}
Martinez-Maradiaga D, Damonte A, Manzo A, Haertel JH, Engelbrecht K (2019)
  Design and testing of topology optimized heat sinks for a tablet.
  International Journal of Heat and Mass Transfer 142:118429

\bibitem[{Monier-Vinard et~al.(2017)Monier-Vinard, Bissuel, Rogi\'e, Laraqi,
  and Daniel}]{monier2017}
Monier-Vinard E, Bissuel V, Rogi\'e B, Laraqi N, Daniel O (2017) Investigation
  on steady-state and transient boundary condition scenarios for optimizing the
  creation of multiport surrogate thermal models. IEEE Transactions on
  Components, Packaging and Manufacturing Technology 8(6):1042--1055

\bibitem[{Murcia et~al.(2018)Murcia, R\'ethor\'e, Dimitrov, Natarajan,
  Sorensen, Graf, and Kim}]{murcia2018}
Murcia JP, R\'ethor\'e PE, Dimitrov N, Natarajan A, Sorensen JD, Graf P, Kim T
  (2018) Uncertainty propagation through an aeroelastic wind turbine model
  using polynomial surrogates. Renewable Energy 119:910--922

\bibitem[{Reimer and Cheviakov(2013)}]{Reimer2013}
Reimer AS, Cheviakov AF (2013) {A Matlab-based finite-difference solver for the
  Poisson problem with mixed Dirichlet-Neumann boundary conditions}. Computer
  Physics Communications 184(3):783--798, \doi{10.1016/j.cpc.2012.09.031}

\bibitem[{Ronneberger et~al.(2015)Ronneberger, Fischer, and
  Brox}]{ronneberger2015u}
Ronneberger O, Fischer P, Brox T (2015) U-net: Convolutional networks for
  biomedical image segmentation. In: International Conference on Medical image
  computing and computer-assisted intervention, Springer, pp 234--241

\bibitem[{Simonyan and Zisserman(2014)}]{2014Very}
Simonyan K, Zisserman A (2014) Very deep convolutional networks for large-scale
  image recognition. Computer ence

\bibitem[{Song and Guo(2011)}]{song2011}
Song B, Guo Z (2011) Robustness in the volume-to-point heat conduction
  optimization problem. International journal of heat and mass transfer
  54:4531--4539

\bibitem[{Szegedy et~al.(2015)Szegedy, Liu, Jia, Sermanet, Reed, Anguelov,
  Erhan, Vanhoucke, and Rabinovich}]{szegedy2015}
Szegedy C, Liu W, Jia Y, Sermanet P, Reed S, Anguelov D, Erhan D, Vanhoucke V,
  Rabinovich A (2015) Going deeper with convolutions. In: IEEE conference on
  computer vision and pattern recognition, pp 1--9

\bibitem[{Visin et~al.(2015)Visin, Kastner, Cho, Matteucci, Courville, and
  Bengio}]{visin2015}
Visin F, Kastner K, Cho K, Matteucci M, Courville A, Bengio Y (2015) Renet: A
  recurrent neural network based alternative to convolutional networks. arXiv
  preprint arXiv: 150500393

\bibitem[{White et~al.(2019)White, Arrighi, Kudo, and Watts}]{white2019}
White DA, Arrighi WJ, Kudo J, Watts SE (2019) Multiscale topology optimization
  using neural network surrogate models. Computer Methods in Applied Mechanics
  and Engineering 346:1118--1135

\bibitem[{Wu and He(2018)}]{wu2018group}
Wu Y, He K (2018) Group normalization. In: Proceedings of the European
  conference on computer vision (ECCV), pp 3--19

\bibitem[{Xu et~al.(2019)Xu, Gao, Wang, Tao, and Xu}]{xu2019}
Xu Z, Gao Y, Wang X, Tao X, Xu Q (2019) Surrogate thermal model for power
  electronic modules using artificial neural network. In: Annual Conference of
  the IEEE Industrial Electronics Society, IEEE, pp 3160--3165

\bibitem[{Yang et~al.(2017)Yang, Tan, Feng, Liu, Guo, and Yan}]{yang2017}
Yang W, Tan RT, Feng J, Liu J, Guo Z, Yan S (2017) Deep joint rain detection
  and removal from a single image. In: Proceedings of the IEEE Conference on
  Computer Vision and Pattern Recognition, pp 1357--1366

\bibitem[{Zhang et~al.(2012)Zhang, Chowdhury, and Messac}]{zhang2012}
Zhang J, Chowdhury S, Messac A (2012) An adaptive hybrid surrogate model.
  Structural and Multidisciplinary Optimization 46(2):223--238

\bibitem[{Zhang et~al.(2020)Zhang, Yang, Yu, and Jiang}]{zhang2020}
Zhang X, Yang B, Yu T, Jiang L (2020) Dynamic surrogate model based
  optimization for mppt of centralized thermoelectric generation systems under
  heterogeneous temperature difference. IEEE Transactions on Energy Conversion
  35(2):966--976

\end{thebibliography}

\end{document}